\documentclass[twoside,11pt]{article}

\usepackage{jmlr2e}

\usepackage{amsmath,amsfonts,bm}

\def\1{\bm{1}}

\def\eps{{\epsilon}}

\def\rvx{{\mathbf{x}}}
\def\rvy{{\mathbf{y}}}

\def\vmu{{\boldsymbol{\mu}}}

\def\valpha{{\boldsymbol{\alpha}}}

\def\vSigma{{\boldsymbol{\Sigma}}}

\def\vy{{\mathbf{y}}}

\def\vL{{\mathbf{L}}}

\def\mA{{\mathbf{A}}}
\def\mB{{\mathbf{B}}}

\def\mX{{\mathbf{X}}}

\DeclareMathAlphabet{\mathsfit}{\encodingdefault}{\sfdefault}{m}{sl}
\SetMathAlphabet{\mathsfit}{bold}{\encodingdefault}{\sfdefault}{bx}{n}

\DeclareMathOperator*{\argmax}{arg\,max}

\newcommand{\vpi}{\boldsymbol{\pi}}

\renewcommand{\vy}{\mathbf{y}}

\usepackage{hyperref}

\usepackage{url}
\usepackage{cleveref}

\usepackage{booktabs}       
\usepackage{arydshln}       
\usepackage{graphicx}       
\usepackage{multirow}       
\usepackage{amssymb}        

\usepackage{tikz, pgfplots}
\pgfplotsset{compat=newest}

\usetikzlibrary{arrows,shapes,plotmarks,decorations.pathmorphing}
\usetikzlibrary{backgrounds,calc,positioning,fadings}

\newlength{\figheight}
\newlength{\figwidth}

\tikzset{every picture/.style={node distance=2cm}}
\tikzset{>=stealth'} 
\tikzstyle{edge}  =[draw=white,double=black,thick,-]

\makeatletter
\newcommand{\superimpose}[2]{%
  {\ooalign{$#1\@firstoftwo#2$\cr\hfil$#1\@secondoftwo#2$\hfil\cr}}}
\makeatother
\newcommand{\ostimes}{\circledast}
\renewcommand{\ostimes}{\mathpalette\superimpose{{\otimes}{\ominus}}}

\usepackage{algorithm}
\usepackage{algpseudocode}

\jmlrheading{?}{2022}{?}{?}{?}{?}{Marius Hobbhahn and Philipp Hennig}

\ShortHeadings{Laplace Matching}{Hobbhahn and Hennig}
\firstpageno{1}

\begin{document}

\title{Laplace Matching for fast Approximate Inference\\ in Latent Gaussian Models}

\author{\name Marius Hobbhahn \email marius.hobbhahn@uni-tuebingen.de \\
       \AND
       \name Philipp Hennig \email philipp.hennig@uni-tuebingen.de \\
       University of Tübingen\\
       Tübingen, Germany}


\maketitle

\begin{abstract}
Bayesian inference on non-Gaussian data is often non-analytic and requires computationally expensive approximations such as sampling or variational inference. We propose an approximate inference framework primarily designed to be computationally cheap while still achieving high approximation quality. The concept, which we call \emph{Laplace Matching}, involves closed-form, approximate, bi-directional transformations between the parameter spaces of exponential families. These are constructed from Laplace approximations under custom-designed basis transformations. The mappings can then be leveraged to effectively turn a latent Gaussian distribution into an approximate conjugate prior to a rich class of observable variables. 
This allows us to train latent Gaussian models such as Gaussian Processes on non-Gaussian data at nearly no additional cost. 
The method can be thought of as a pre-processing step which can be implemented in $<$5 lines of code and runs in less than a second. 
Furthermore, Laplace Matching yields a simple way to group similar data points together, e.g.~to produce inducing points for GPs. 
We empirically evaluate the method with experiments for four different exponential distributions, namely the Beta, Gamma, Dirichlet and inverse Wishart, showing approximation quality comparable to state-of-the-art approximate inference techniques at a drastic reduction in computational cost.
\end{abstract}


\section{Introduction}
\label{sec:introduction}
Probabilistic inference often involves latent and observable variables of different types, lying in different domains. Apart from the most basic cases, one cannot expect to find a joint conjugate prior for all latent quantities. If we observe, for example, discrete samples from a latent categorical distribution, then the Dirichlet exponential family offers a conjugate prior, and Bayesian inference is analytic and computationally extremely cheap---it simply involves adding floating-point numbers. However, if we observe several such discrete samples from several latent categoricals that must be assumed to relate to each other somehow, the Dirichlet is the wrong model. A standard idea would then be to construct a latent Gaussian model with a latent multi-output Gaussian process prior connected to the individual categoricals through a softmax link function. However, the softmax of a Gaussian random variable has no useful analytic form, and inference is intractable. Approximate inference methods constructed from Laplace approximations \citep[cf.~\textsection 3 in][]{Rasmussen06gaussianprocesses} or even custom-built Markov Chain Monte Carlo (MCMC) methods \citep{murray2009elliptical} are available, but they have significantly higher computational cost than conjugate prior inference.

\newcommand{\hwplotStandard}{\raisebox{2pt}{\tikz{\draw[black,solid,line width=0.9pt](0,0) -- (5mm,0);}}}
\newcommand{\hwplotLog}{\raisebox{2pt}{\tikz{\draw[red,solid,line width=1.2pt](0,0) -- (5mm,0);}}}
\newcommand{\hwplotSqrt}{\raisebox{2pt}{\tikz{\draw[blue,solid,line width=1.2pt](0,0) -- (5mm,0);}}}

\begin{figure}[!tb]
    \sbox0{\hwplotStandard}\sbox1{\hwplotLog}\sbox2{\hwplotSqrt}%
    \setlength{\figwidth}{0.1\textwidth}
    \setlength{\figheight}{0.2\textheight}
    \centering
    \includegraphics[width=\textwidth]{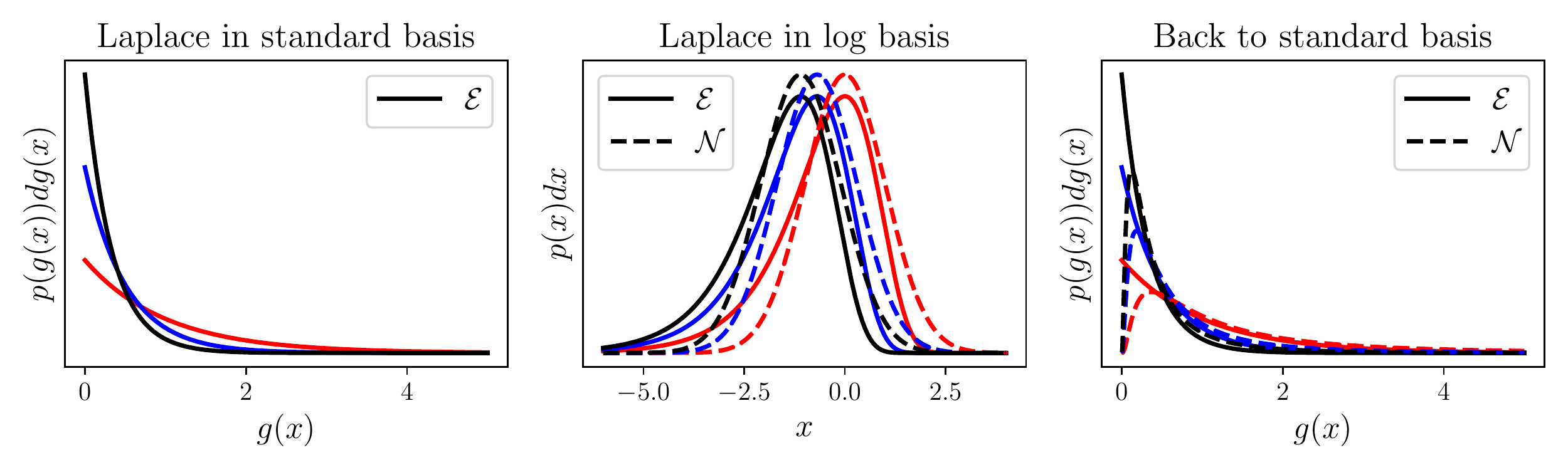} \\
    \includegraphics[width=\textwidth]{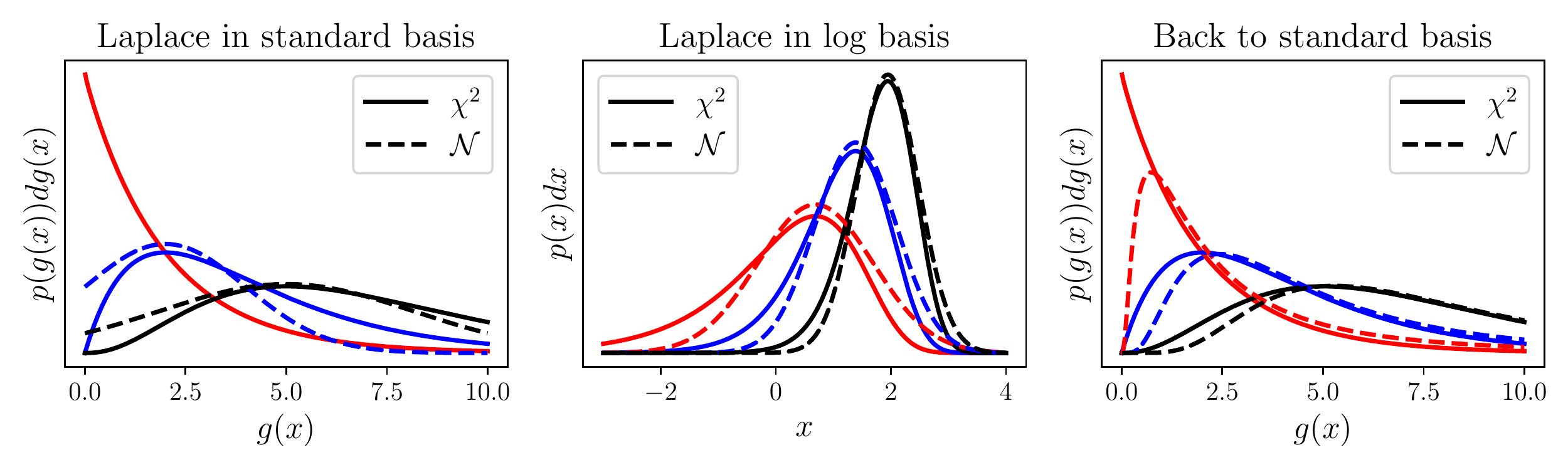} \\
    \includegraphics[width=\textwidth]{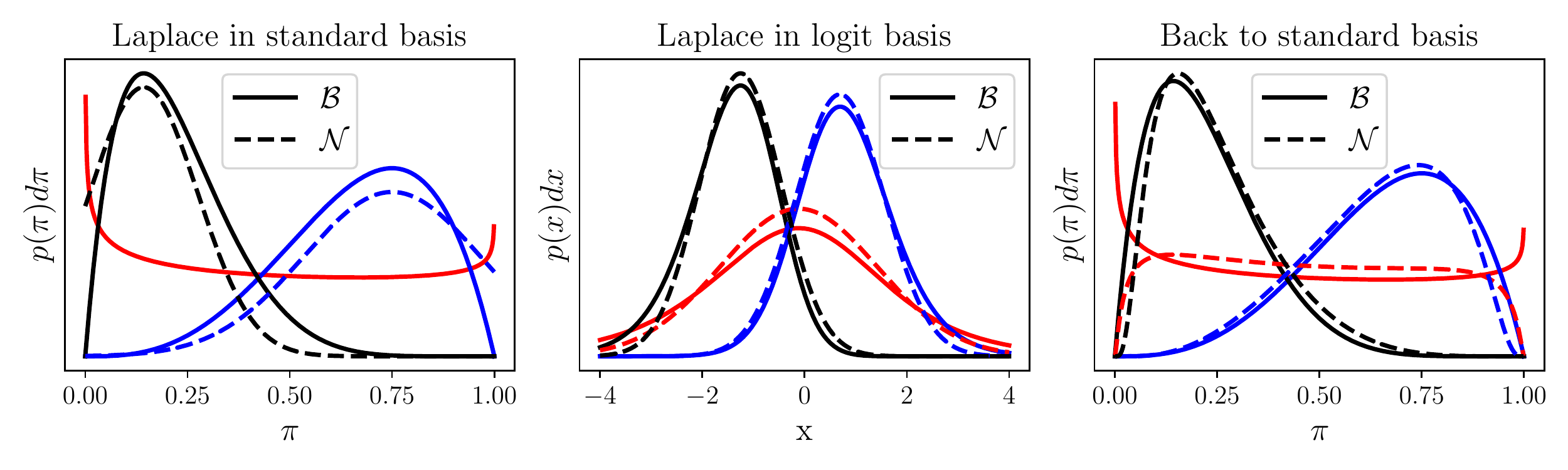}
    \caption{Laplace approximations for the exponential distribution (top), the Chi-squared distribution (middle row), and the Beta distribution (bottom) in the standard basis (left) and a more suitable basis (middle column) for three different sets of parameters (\usebox0, \usebox1 and \usebox2). The Laplace approximations are then transformed back to the standard base for comparison (right). The parameter for the red line does not have a valid Laplace approximation in the standard base, since the Hessian is not positive definite. In the new basis, however, this problem does not occur. Similar figures for all other exponential families can be found in the appendix.}
    \label{fig:LA_different_basis}
\end{figure}

We introduce a principled framework, which we call \emph{Laplace Matching} (LM), to construct a fast, closed-form, bi-directional approximate transformation between the parameters of any exponential family distribution and a Gaussian. Of course, Laplace approximations are well-studied.
The key novel part of our construction is to first transform the variable of the original distribution such that it can be approximated better by a Laplace approximation in the new base (cf.~Figure \ref{fig:LA_different_basis} for a visual sketch). Our approach builds upon an insight by \citet{MacKay1998} which we extend and generalize. 
With LM, we can effectively do Gaussian inference on likelihoods that have exponential family conjugate priors with very little additional cost. In this paper, we focus primarily on Gaussian Process (GP) inference but LM can be used in many more settings to seamlessly translate between exponential families and Gaussians. 

Our contributions include
a) setting up the general framework for Laplace Matching (LM), 
b) providing derivations for most commonly used exponential families, 
c) showcasing how LM leads to a natural and simple way to choose inducing points in GP inference, 
d) showing feasibility by applying LM to GPs on multiple non-Gaussian, commonly used data types (binary, counts, categorical and covariances) with a few lines of code, and 
e) comparing LM to multiple alternatives w.r.t.~approximation quality and speed.

\section{Background}
\label{background}
Necessary components for Laplace Matching are change of variables, Laplace approximations, and exponential families.

\subsection{Variable Transformation for pdfs}
\label{subsec:variable_transform}

Let $X$ have a continuous density $f_X$.\footnote{This definition follows ``Probability Essentials'' by \citet{jacod2004probability}} Let $g: \mathbb{R} \rightarrow \mathbb{R}$ be piecewise strictly monotonic and continuously differentiable, i.e.~there exist intervals $I_1,I_2, ..., I_n$ that partition $\mathbb{R}$ such that $g$ is strictly monotonic and continuously differentiable on the interior of each $I_i$. For each interval $i,\, g:I_i \rightarrow \mathbb{R}$ is invertible on $g(I_i)$; let $g_i^{-1}$ be its inverse function. Let $Y = g(X)$ and $r(y) = \{y \,|\, y=g(x), x \in \mathbb{R}\}$ be the range of $g$. Then, the density function $f_Y$ of $Y$ exists and is given by 
\begin{equation}
\label{eq:1D_variable_transform}
f_Y(y) = \sum_{i=1}^{n} f_X(g_i^{-1}(y)) \left\vert\frac{\partial g_i^{-1}(y)}{\partial y} \right\vert \mathbf{1}_{r(y)}\,.
\end{equation}
This is also true for the multi-dimensional case
\begin{equation}
f_Y(\mathbf{y}) = \sum_{i=1}^{n} f_\rvx\Big(g^{-1}(\mathbf{y})\Big)\left\vert \det\left[\frac{dg^{-1}(\mathbf{x})}{d\mathbf{x}}\Bigg \vert_{\mathbf{x}=\mathbf{y}}\right]\right \vert \mathbf{1}_{r(y)}\,.
\end{equation}
where the function in each interval is multiplied with the determinant of the Jacobian of $g^{-1}$ evaluated at $\rvy$.
\subsection{Laplace Approximation}
\label{subsec:LA}

The Laplace approximation fits a normal distribution to a given function, in our case a probability density function (pdf). If $\hat{\theta}$ denotes the mode of a pdf $h(\theta)$, then it is also the mode of the log-pdf $q(\theta) = \log h(\theta)$ since the logarithm is a monotonic transformation. The 2nd order Taylor expansion of $q(\theta)$ is
\begin{align}
	q(\theta) &\approx q(\hat{\theta}) + \nabla q(\hat{\theta})(\theta - \hat{\theta}) + \frac{1}{2}(\theta- \hat{\theta})\nabla\nabla^\top q(\hat{\theta}) (\theta - \hat{\theta})\\
	&= 	q(\hat{\theta}) + 0 +  \frac{1}{2}(\theta- \hat{\theta})\nabla\nabla^\top q(\hat{\theta}) (\theta - \hat{\theta}) \qquad \text{[since } \nabla q(\theta) = 0]\\
	&= c - \frac{1}{2} (\theta - \mu)^\top \Sigma^{-1} (\theta - \mu)\,,
\end{align}
where $c$ is a constant, $\mu = \hat{\theta}$ and $\Sigma = \{-\nabla \nabla^\top q(\hat{\theta})\}^{-1}$. Since the Gaussian is fit at the mode, the gradient term is zero. The right-hand side of the last line matches the log-pdf of a Gaussian. Therefore, the pdf $h(\theta)$ is approximated by the pdf of the normal distribution $\mathcal{N}(\mu, \Sigma)$ where $\mu = \hat{\theta}$ and $\Sigma = \{-\nabla\nabla^\top q(\hat{\theta})\}^{-1}$. For a visual interpretation consider the first two columns of Figure \ref{fig:LA_different_basis}.

\subsection{Exponential Families}
\label{subsec:EF}

A pdf that can be written in the form 
\begin{align}\label{EF}
p(x) &= h(x)\cdot \exp\left(w^\top \phi(x) - \log Z(w) \right)
\end{align}
where
\begin{align}
Z(w)&:= \int_{\mathbf{X}} h(x)\exp\left(w^\top \phi(x)\right) \,dx
\end{align}
is called an exponential family. $\phi(x): \mathbb{X} \rightarrow \mathbb{R}^d$ are the sufficient statistics, $w \in \mathcal{D} \subseteq \mathbb{R}^d$ the natural parameters with domain $\mathcal{D}$, $\log Z(w): \mathbb{R}^d \rightarrow \mathbb{R}$ is the (log) partition function (normalization constant), and $h(x): \mathbb{X} \rightarrow \mathbb{R}_{+}$ the base measure. 

The product of two pdfs of different instances of the same exponential family is proportional to another instance of this exponential family.
Assume we have two instances of the same exponential family
\begin{equation}
    p(x; w_i) = h(x) \cdot \exp(\phi(x)^\top w_i - \log Z(w_i))
\end{equation}
where $i \in \{1,2\}$ and $w_i \in \mathcal{D}$. Then, the product of their pdfs is
\begin{equation}
    p(x; w_1) \cdot p(x; w_2) \propto p(x; w_1 + w_2)
\end{equation}
if $w_1+w_2 \in \mathcal{D}$. 
Thus, the product of two pdfs can be computed by adding their parameters. 

Another important property of exponential families is that their expected value is given by the differential of the normalizing constant
\begin{equation}\label{eq:exp_fam_expected}
    \mathbb{E}_{p_w}(\phi(x)) = \nabla_w \log Z(w)\,.
\end{equation}
This makes the computation of moments easier since it replaces integration with differentiation. 
\section{Laplace Matching}
\label{sec:laplace_matching}
\begin{algorithm}
\caption{Laplace Matching}\label{alg:LM}
\begin{algorithmic}
\Require Exponential family pdf $p(x\vert \theta)$ with parameters $\theta$, transformation $g(x)$
\State $p_y(y; \theta) = p_x(g^{-1}(y); \theta)$   \Comment{Apply transformation of variable to pdf}
\State $\mu =  \argmax_y p_y(y; \theta) \Leftrightarrow \frac{\partial p_y(y; \theta)}{\partial \theta} = 0$  \Comment{Compute mode}
\State $\Sigma =  {-\left(\frac{\partial^2 p_y(y; \mu)}{\partial^2 y}\right)}^{-1}$  \Comment{Compute Hessian and invert}
\State results in $f: \theta \rightarrow (\mu, \Sigma)$
\State $f^{-1}: (\mu, \Sigma) \rightarrow \theta$   \Comment{Invert mapping} \\
\Return $f: \theta \leftrightarrow (\mu, \Sigma)$   \Comment{Return mapping between parameters}
\end{algorithmic}
\end{algorithm}
The goal of Laplace Matching is to fix the approximation of the likelihood as a Laplace approximation and then to reduce the KL-divergence between the true distribution and the Gaussian approximation by representing the likelihood in a new basis.
Most other approximation schemes \citep[e.g.][]{GLMtoolboxNickisch2012,MinkaEP2001,murray2009elliptical, Neal93probabilisticinference,VISeeger2009,WilliamsBarberGPClassification1998,wilson2010generalised} are based on an internal step of numerical optimisation. They define a loss for the approximation quality, then define an approximation scheme and iteratively adapt the approximation to improve the quality. Each iteration has some cost.
In contrast, LM starts by proposing an approximation of fixed cost, e.g.~the Laplace approximation, and then changes the representation of the data to minimize the loss. The transformation is done in closed form, i.e. there are no iterations, and the approximation error is therefore fixed. The cost of finding a good approximation is externalized because it is chosen beforehand and the primary advantage of LM is its speed. Put differently, the aforementioned methods expend computational resources in order to find a particularly good approximation -- while Laplace matching adopts the more simplistic approach of proposing one particular approximation of known, very low cost.

While we thus specifically do not aim to numerically minimize some loss, we instead \emph{motivate} the choice of a particular analytic approximation by the KL-divergence between an exponential family distribution and its Gaussian approximation, under a variable transformation.
This works as follows. Let $p(x)$ be an exponential family distribution in the `standard basis' $x$:
\begin{equation}
    p_x(x) = h(x)\exp\left(w^T \phi(x)  - \log Z(w)\right)\,.
\end{equation}
Then, the random variable $x$ can be changed to another basis with transformation $g(x)$ via Equation \ref{eq:1D_variable_transform} (see Section \ref{subsec:variable_transform}). Let $x(y) = g^{-1}(x)$ be the inverse transform of $g(x)$. The resulting distribution $p_y(x(y))$ in basis $y$ is again an exponential family
\begin{align}
    p_y(x(y)) &= h(x(y)) \exp\left(w^T \phi(x(y)) - \log Z(w)\right) \cdot \left\vert \frac{\partial x(y)}{\partial y}\right\vert \\
    &= \exp\left(w^T \phi(x(y)) + \log(h(x(y))) + \log \left\vert \frac{\partial x(y)}{\partial y}\right\vert - \log Z(w)\right)\,.
\end{align}
%
In the new basis $y$ a Laplace approximation is performed to yield a Gaussian $q(y) = \mathcal{N}(y; \mu, \Sigma)$, which can be written as
\begin{equation}\label{eq:q(y)_explicit}
    q(y) = \exp\left( \underbrace{(\Sigma_g^{-1}\mu_g)^\top y - \frac{1}{2}\operatorname{Tr}\left(\Sigma_g^{-1}yy^\top\right)}_{w_q^T\phi_q(y)} - \underbrace{\frac{k}{2}\log(2\pi)}_{\log h_q(y)} + \underbrace{\frac{1}{2} \mu_g^\top\Sigma_g^{-1} \mu_g - \frac{1}{2} \log \vert \Sigma_g \vert}_{-\log Z_q(w_q)} \right),
\end{equation}
where the mean and covariance
\begin{align}\label{eq:mu_g}
    \mu_g &= \argmax_y p(x(y)) \Leftrightarrow \frac{\partial p_y(x(y))}{\partial y} = 0\\
    \Sigma_g &= -\left\{\frac{\partial^2 p(x(y); \mu)}{\partial^2 y}\right\}^{-1}
\end{align}
depend on the choice of parameters $w$ of $p(x(y))$, which in turn depends on $g(x)$.
We then define Laplace Matching as the procedure of finding a transformation $g^{-1}(x) = x(y)$ with accompanying basis $y$ for distribution $p(x)$ such that $\text{KL}\left(p(x(y))\vert \vert q(y)\right)$ is small. 
Thus, LM describes a map from the parameters $\theta$ of $p(x)$ to the parameters $(\mu, \Sigma)$ of the Gaussian approximation $q(y)$. The aim is to invert this map to also have a mapping from $(\mu, \Sigma)$ to $\theta$. However, since not all variable transformations are bijective (e.g. the softmax), the inverse map has to be approximated in some cases. 
Therefore, Laplace Matching ultimately yields a bi-directional, closed-form mapping between the parameters $\theta$ and $(\mu, \Sigma)$.
One way to define an \emph{optimal} transformation $g$ would be that which minimizes $\text{KL}\left(p(x(y))\vert \vert q(y)\right)$. 
\begin{align}
    &\text{KL}(p_y(x(y)) \vert\vert q(y)) = \int p(y) \log\left(\frac{p(y)}{q(y)}\right) dy \\
    &= \int p(y) \log(p(y)) dy - \int p(y) \log(q(y)) dy \\
    &= \int p(y) \left(\log h_p(y) + w_p^\top \phi_p(y) - \log Z(w_p) \right) dy \\ \nonumber
    &-\int p(y) \left(\log h_q(y) + w_q^\top \phi_q(y) - \log Z(w_q) \right) dy \\
    &= \mathbb{E}_p \left[\log h_p(y) \right] + w_p^\top \mathbb{E}_p \left[ \phi_p(y) \right] - \log Z(w_p) - \mathbb{E}_p \left[\log h_q(y) \right] - w_q^T \mathbb{E}_p \left[ \phi_q(y) \right] + \log Z(w_q) \label{eq:KL_div_lastline}
\end{align}
The objective $\text{KL}\left(p(x(y))\vert \vert q(y)\right)$ is in general not differentiable w.r.t.~$g(x)$, due to the $\argmax$-function in the mode (see Equation \ref{eq:mu_g}). So even numerical optimization is challenging, let alone finding a global analytic extremum. Thus, other strategies have to be used to find an optimal transformation $g(x)$.

It is generally known (see,~e.g.~\citep[cf.~\textsection 10.7 in][]{BishopPRML}) that the KL divergence between a distribution $p(y)$ and its Gaussian approximation $q(y)$ is minimized when the first two \emph{moments match}. This comes from optimizing $\text{KL}\left(p_y(x(y)) \vert\vert q(y)\right)$~w.r.t.~$w_q$. Writing Equation \ref{eq:KL_div_lastline} as a function of $w_q$ yields
\begin{equation}
    \text{const.} - w_q^T \mathbb{E}_p \left[ \phi_q(y) \right] + \log Z(w_q)
\end{equation}
which, if differentiated w.r.t $w_q$, yields
\begin{align}
    - \mathbb{E}_p \left[ \phi_q(y) \right] &+ \nabla_{w_q}\log Z(w_q) \overset{!}{=} 0 \\
    \Leftrightarrow \mathbb{E}_p \left[ \phi_q(y) \right] &= \nabla_{w_q}\log Z(w_q) \\
    \Leftrightarrow \mathbb{E}_p \left[ \phi_q(y) \right] &= \mathbb{E}_q \left[ \phi_q(y) \right] \qquad
\end{align}
due to the properties of exponential families described in Subsection \ref{subsec:EF}. 
Therefore, the optimal choice for $g(x)$ is that which matches both sufficient statistics of the Gaussian $[x, xx^\top]^\top$. 
However, it is not possible to match both moments jointly with one transformation $g$ (otherwise our distribution would be the Gaussian). Therefore, we instead propose two different approaches: Either we choose $g$ such that the mode $\mu$ of $p_y(x(y))$ matches the mean, $\Bar{\mu} = \argmax_w \log p(x; w) \overset{!}{=} g^{-1}\left(\mathbb{E}_p (x; w) \right)$. Or we choose $g$ such that the second moment matches,~i.e. $\mathbb{E}(xx^\top) = -\left[\nabla\nabla^\top \log p(x; \Bar{\mu})\right]^{-1}$.

Note that neither of these two approaches necessarily provides a minimum of the KL divergence. But both are analytically tractable! In absence of a clear formal motivation to prefer one option over the other, Section \ref{subsec:distance_measures} provides an empirical comparison.

Furthermore, the \emph{domain} of $p_y(x(y))$ should be $\mathbb{R}^{d}$ since this is the domain of the $d$-dimensional Gaussian $q(y)$, which is given by the Laplace approximation. If this is not the case, the support of $p_y(x(y))$ is smaller than the support of $q(y)$ and therefore $\text{KL}\left(p_y(x(y)) \vert\vert q(y)\right)$ is undefined. Matching the first or second moment both provides the correct domain which is an additional reason to only match one moment. However, there are multiple further ways to choose the distribution that could be investigated in future work. 

\renewcommand{\arraystretch}{1.1}
\begin{table}[tb]
    \centering
    \resizebox{\textwidth}{!}{
    \begin{tabular}{lllcccc}
        \toprule
        \textbf{Distribution}    &\textbf{$g(x)$} &\textbf{$g^{-1}(x)$} & standard $\phi(x)$ & new $\phi(y)$ & Standard domain & New domain    \\
        \midrule
        \multirow{2}{*}{Exponential}  & log & exp & $x$ & $(\mathbf{y},\exp(y))$ & $\mathbb{R}_{+}$ & $\mathbb{R}$\\
          & sqrt & sqr & $x$ & $(\log(y), \mathbf{y}^2)$ & $\mathbb{R}_{+}$ & $\mathbb{R}$\\[0.1cm]
        \multirow{2}{*}{Gamma}       & log & exp & $(\log(x), x)$ & $(\mathbf{y},\exp(y))$ & $\mathbb{R}_{+}$ & $\mathbb{R}$\\
        & sqrt & sqr & $(\log(x), x)$ & $(\log(y),\mathbf{y}^2)$ & $\mathbb{R}_{+}$ & $\mathbb{R}$ \\[0.1cm]
        \multirow{2}{*}{Inv. Gamma} & log & exp& $(\log(x), x)$ & $(\mathbf{y},\exp(y))$ & $\mathbb{R}_{+}$ & $\mathbb{R}$\\
        & sqrt & sqr& $(\log(x), x)$ & $(\log(y),\mathbf{y}^2)$ & $\mathbb{R}_{+}$ & $\mathbb{R}$\\[0.1cm]
        \multirow{2}{*}{Chi-squared}     & log & exp &  $(\log(x), x)$ & $(\mathbf{y},\exp(y))$ & $\mathbb{R}_{+}$ & $\mathbb{R}$\\
        & sqrt & sqr &  $(\log(x), x)$ & $(\log(y),\mathbf{y}^2)$ & $\mathbb{R}_{+}$ & $\mathbb{R}$\\[0.1cm]
        Beta        & logit & logistic&  $(\log(x),\log(1-x))$ & $(\log(\sigma(\mathbf{y}))), (1-\log(\sigma(y)))$  & $\mathbb{P}$ & $\mathbb{R}$\\[0.1cm]
        Dirichlet     & - & softmax &  $(\log(x_i))$ & $\log(\pi_i(\mathbf{y}))$ & $\mathbb{P}^d$ & $\mathbb{R}^d$\\[0.1cm]
        \multirow{2}{*}{Wishart}        & logm & expm & $(\operatorname{logm}(X), X)$ & $(\mathbf{Y}, \operatorname{expm}(Y))$ &  $\mathbb{R}^{d\times d}_{++}$ & $\mathbb{R}_{S}^{d\times d}$\\
        & sqrtm & sqrm & $(\operatorname{logm}(X), X)$ & $(\operatorname{logm}(Y), \mathbf{Y}^2)$ & $\mathbb{R}^{d\times d}_{++}$ & $\mathbb{R}_{S}^{d\times d}$\\[0.1cm]
        \multirow{2}{*}{Inv. Wishart} & logm & expm & $(\operatorname{logm}(X), X)$ & $(\mathbf{Y}, \operatorname{expm}(Y))$ & $\mathbb{R}^{d\times d}_{++}$ & $\mathbb{R}_{S}^{d\times d}$\\
        & sqrtm & sqrm & $(\operatorname{logm}(X), X)$ & $(\operatorname{logm}(Y), \mathbf{Y}^2)$ & $\mathbb{R}^{d\times d}_{++}$ & $\mathbb{R}_{S}^{d\times d}$\\
        \bottomrule
    \end{tabular}
    } 
    \caption{Overview of the transformations and the sufficient statistics and domains of the exponential families in the standard and new basis.
    Bold notation for the new sufficient statistics denotes which moment has been matched by the transformation.
    Beta and Dirichlet only have one transformation each, as they have only one kind of sufficient statistic, namely $\log(x_i)$. $\mathbb{R}^{d\times d}_{++}$ and $\mathbb{R}_{S}^{d\times d}$ describe the spaces of positive definite and symmetric matrices respectively.}
    \label{table:basis_overview}
\end{table}
\renewcommand{\arraystretch}{1.0}

LM, as presented in this paper, applies to all forms of latent Gaussian models. This includes GP regression, Kalman filters, stochastic differential equations, or the last layer of Neural Networks \citep{hobbhahn2021fast}.
In more simple terms, all operations that can be applied to Gaussians can also be applied to any other random variable with an exponential family conjugate prior through LM at the cost of a fixed approximation loss.

\subsection{Bases and Transformations}
\label{subsec:bases}

The basis transformations presented in the following have been chosen through the different approaches outlined above, i.e.~they always match the first or second sufficient statistic. An overview of the distributions, their transformations, their standard, and new sufficient statistics, and their domain can be found in Table \ref{table:basis_overview}. It is complemented by Table \ref{table:transformations}, which provides the explicit transformations between the parameters of different exponential families and a Gaussian for a given basis. For a visual interpretation, we refer the reader to Figure \ref{fig:LA_different_basis} where three different distributions with their respective Laplace approximations are shown.
Even though the figure does not contain all distributions listed in Table \ref{table:basis_overview} the Beta distribution can be seen as a representative for the Dirichlet, as it is its 1D special case and the Gamma distribution can be seen as a representative for Chi-square, the {(inverse-)~Gamma} and {(inverse-)~Wishart} since the Chi-square and the Gamma are 1D special cases of the {(inverse-)~Wishart}.

\renewcommand{\arraystretch}{1.3}
\begin{table}[tb]
    \centering
    \resizebox{\textwidth}{!}{  
    \begin{tabular}{l  lll}
        \toprule
        Distribution & Basis & $\theta \rightarrow \mathcal{N}$& $\mathcal{N} \rightarrow \theta$ \\
        \midrule 
        \multirow{4}{*}{Exponential} & \multirow{2}{*}{log} & $\mu =  \log(\frac{1}{\lambda})$ & $\lambda = \frac{1}{\exp(\mu)}$ \\
        & & $\sigma^2 = 1$ &  \\
        & \multirow{2}{*}{sqrt} & $\mu = \sqrt{\frac{1}{2\lambda}} $ & $\lambda = \frac{1}{2\mu^2}$ \\
        & & $\sigma^2 = \frac{1}{4\lambda}$ &  \\
        \midrule 
        \multirow{4}{*}{Gamma} & \multirow{2}{*}{log} & $\mu = \log\left(\frac{\alpha}{\lambda}\right)$ & $\alpha = \frac{1}{\sigma^2}$ \\
        && $\sigma^2 = \frac{1}{\alpha}$ &    $\lambda =  \frac{1}{\exp(\mu)\sigma^2}$ \\
        & \multirow{2}{*}{sqrt} & $\mu = \sqrt{\frac{\alpha-0.5}{\lambda}}$ & $\alpha = \frac{\mu^2}{4\sigma^2}-0.5$ \\
        & & $\sigma^2 = \frac{1}{4\lambda}$ & $\lambda = \frac{4}{\sigma^2}$ \\
        \midrule 
        \multirow{4}{*}{Inv. Gamma} & \multirow{2}{*}{log} & $\mu = \log\left(\frac{\lambda}{\alpha}\right)$ & $\alpha = \frac{1}{\sigma^2}$ \\
        && $\sigma^2 = \frac{1}{\alpha}$ &    $\lambda =  \frac{\exp(\mu)}{\sigma^2}$ \\
         & \multirow{2}{*}{sqrt} & $\mu = \sqrt{\frac{\lambda}{\alpha}}$ & $\alpha = \frac{\mu^2}{4\sigma^2} - 0.5$ \\
        && $\sigma^2 = \frac{\lambda}{4\alpha^2}$ &    $\lambda = \frac{\mu^4}{4\sigma^2} $ \\
        \midrule 
        \multirow{4}{*}{Chi-squared} & \multirow{2}{*}{log} & $\mu = \log(k)$ & $k = \exp(\mu)$ \\
        && $\sigma^2 = 2/k $ &    \\
         & \multirow{2}{*}{sqrt} & $\mu = \sqrt{k}$ & $k = \mu^2$ \\
        && $\sigma^2 = 1/2$ & \\
        \midrule 
        \multirow{2}{*}{Beta} & \multirow{2}{*}{logit} & $\mu = \log(\frac{\alpha}{\beta})$ & $\alpha = \frac{\exp(\mu) + 1}{\sigma^2}$ \\
        && $\sigma^2 = \frac{\alpha + \beta}{\alpha\beta}$ &    $\beta = \frac{\exp(-\mu) + 1}{\sigma^2} $ \\
        \midrule 
        \multirow{2}{*}{Dirichlet*} & \multirow{2}{*}{softmax$^{-1}$} & $\mu_k = \log \alpha_k  - \frac{1}{K} \sum_{l=1}^{K} \log \alpha_l$ & $\alpha_k = \frac{1}{\Sigma_{kk}}\left(1 - \frac{2}{K} + \frac{e^{\mu_k}}{K^2}\sum_{l=1}^K e^{-\mu_l} \right)$ \\
        && $\Sigma_{k\ell} = \delta_{k\ell}\frac{1}{\alpha_k} - \frac{1}{K}\left[\frac{1}{\alpha_k} + \frac{1}{\alpha_\ell} - \frac{1}{K}\sum_{u=1} ^K \frac{1}{\alpha_u} \right]$ & \\
        \midrule 
        \multirow{4}{*}{Wishart} & \multirow{2}{*}{logm} & $\mu = \operatorname{logm}((n-p+1)V) $ & $V = \frac{\operatorname{expm(\mu)}}{(n-p+1)}$ \\
        && $\Sigma = \frac{2}{(n-p+1)} (I_p \ostimes I_p)^{-1}$ & $n=\frac{2p^2}{\operatorname{tr}(\Sigma)} + p - 1$\\
         & \multirow{2}{*}{sqrtm} & $\mu = \operatorname{sqrtm}\left((n-p)V\right)$ & $V =**$ \\
        && $\Sigma = \left(V^{\frac{1}{2}}\ostimes V^{\frac{1}{2}}\right) \left(I_{p^2} + V^{\frac{1}{2}} \ostimes V^{-\frac{1}{2}}\right)^{-1}$ & $n=**$\\
        \midrule 
        \multirow{4}{*}{Inv. Wishart} & \multirow{2}{*}{logm} & $\mu = \operatorname{logm}\left(\frac{1}{(\nu+p-1)\Psi}\right)$ & $\Psi = (\nu + p - 1)\operatorname{expm}(\mu)$ \\
        && $\Sigma = 2(\nu + p - 1)(I_p \ostimes I_p)^{-1}$ & $\nu=\frac{\operatorname{tr}(\Sigma)}{2p^2} - p + 1$\\
         & \multirow{2}{*}{sqrtm} & $\mu = \operatorname{sqrtm}(\frac{1}{(\nu + p)}\Psi^{-1})$ & $\Psi =**$ \\
        && $\Sigma = \frac{1}{(\nu + p)^2} \left(I_p \ostimes \Psi\right)\left(\Psi^{\frac{1}{2}} \ostimes \Psi^{\frac{1}{2}} + I_p\right)^{-1}$ & $\nu=**$\\
        \bottomrule
    \end{tabular}
    } 
    \caption{Overview of all closed-form transformations for Laplace Matching. * indicates an approximate inversion since the transformation is not bijective. ** means that the solution has to be looked up in Appendix \ref{sec:appendix_B}, which also contains practical tips for the (inverse-)~Wishart.}
    \label{table:transformations}
\end{table}
\renewcommand{\arraystretch}{1.0}
%
\section{Laplace Matching with Gaussian Processes}
\label{sec:LM+GP}
\begin{algorithm}[htb!]
\caption{Laplace Matching + Gaussian Process (LM+GP) V1}\label{alg:LMGP1}
\begin{algorithmic}
\Require non-Gaussian data distribution $d$ which is conjugate to an EF, Transformation
\State $\mathcal{EF}(x(f);\eps_\theta + \theta_d)$   \Comment{Use EF to define pseudo-likelihoods in the data domain}
\State $\mathcal{GP}(f;\mu_p,\Sigma_p) = LM^{-1}(\mathcal{EF})$  \Comment{Use LM to map EF to Gaussian}
\State fit GP   \Comment{Apply standard GP toolbox on resulting Gaussian}
\State $X' = x(y), y\sim\mathcal{GP}$   \Comment{Draw samples from latent GP and apply transformation} \\
\Return $\mathcal{GP}(f;\mu_p,\Sigma_p), X'$ \Comment{Return GP and transformed samples} 
\end{algorithmic}
\end{algorithm}
\begin{algorithm}[htb!]
\caption{Laplace Matching + Gaussian Process (LM+GP) V2}\label{alg:LMGP2}
\begin{algorithmic}
\Require non-Gaussian data distribution $d$ which is conjugate to an EF, Transformation
\State $\mathcal{GP}(f;\mu_0,\Sigma_0)$   \Comment{Define prior GP in Gaussian latent space}
\State $\mathcal{EF}(x(f);\theta_0) = LM(GP)$   \Comment{Transform GP with LM}
\State $\mathcal{EF}(x(f);\theta_0 + \theta_d)$  \Comment{Update Exp. Fam. with data}
\State $\mathcal{GP}(f;\mu_p,\Sigma_p) = LM^{-1}(\mathcal{EF})$  \Comment{Transform back to latent Gaussian process}
\State $X' = x(y), y\sim\mathcal{GP}(f;\mu_p,\Sigma_p)$   \Comment{Draw samples from latent GP and apply transformation}
\Return $\mathcal{GP}(f;\mu_p,\Sigma_p), X'$ \Comment{Return GP and transformed samples}
\end{algorithmic}
\end{algorithm}
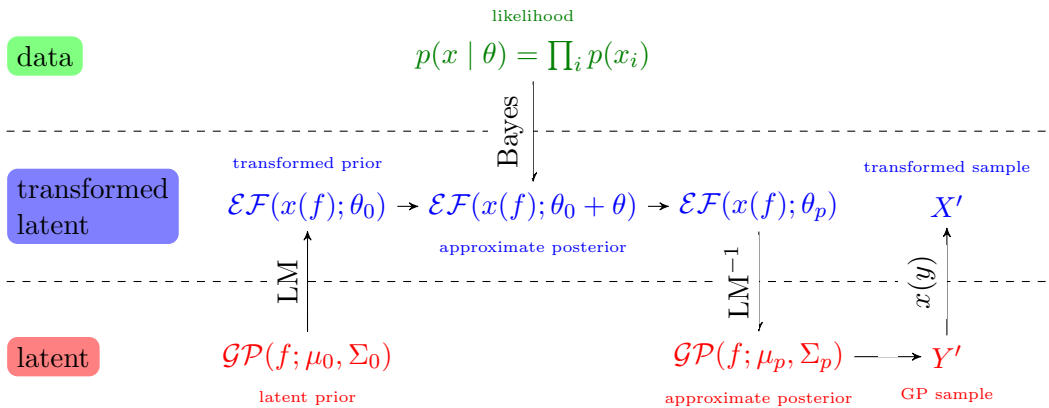
\begin{figure}[!tb]
    \centering
    \begin{tikzpicture}
    \node[rounded corners, fill=red!50!white, anchor=west] at (0,0) {latent};
    \node[rounded corners, fill=blue!50!white, anchor=west, align=left] at (0,2) {transformed \\latent};
    \node[rounded corners, fill=green!50!white, anchor=west] at (0,4) {data};
    
    \draw[dashed] (0,1) -- (14,1);
    \draw[dashed] (0,3) -- (14,3);
    
    \node[text=red] (prior) at (4,0) {$\mathcal{GP}(f;\mu_0,\Sigma_0)$};
    \node[text=blue] (tprior) at (4,2) {$\mathcal{EF}(x(f);\theta_0)$};
    \draw (prior) edge[->] node[midway,above,rotate=90,fill=white] {LM} (tprior);
    
    \node[text=blue] (post) at (7,2) {$\mathcal{EF}(x(f);\theta_0 + \theta)$} edge[<-] (tprior);
     
    \node[text=green!50!black] (obs) at (7,4) {$p(x \mid \theta) = \prod_i p(x_i)$} edge[->] node[midway,above,rotate=90,fill=white] {Bayes} (post);
    
    \node[text=blue] (cav) at (10,2) {$\mathcal{EF}(x(f); \theta_p)$} edge[<-] (post);
    
    \node[text=red] (tpost) at (10,0) {$\mathcal{GP}(f;\mu_p,\Sigma_p)$};
    
    \draw (cav) edge[->] node[midway,above,rotate=90,fill=white] {LM$^{-1}$} (tpost);
    
    \node[text=red] (s_post) at (12.5,0) {$Y'$};
    
    \draw (tpost) edge[->] node[midway, above, rotate=0, fill=white] {} (s_post);
    
    \node[text=blue] (s_post_t) at (12.5,2) {$X'$};
    
    \draw (s_post) edge[->] node[midway, above, rotate=90, fill=white] {$x(y)$} (s_post_t);

    \node[anchor=north,text=red] at (prior.south) {\tiny latent prior};
    \node[anchor=south,text=blue] at (tprior.north) {\tiny transformed prior};
    \node[anchor=north,text=blue] at (post.south) {\tiny approximate posterior};
    \node[anchor=south,text=green!50!black] at (obs.north) {\tiny likelihood};
    \node[anchor=south,text=blue] at (cav.north) {};
    \node[anchor=north,text=red] at (tpost.south) {\tiny approximate posterior};
    \node[anchor=north,text=red] at (s_post.south) {\tiny GP sample};
    \node[anchor=south,text=blue] at (s_post_t.north) {\tiny transformed sample};
    
    \end{tikzpicture}
    \caption{Visual description of Bayesian Inference with LM+GP. A latent prior GP is transformed to the latent exponential family of choice via Laplace Matching. It is then updated by the likelihood which is conjugate to the latent exponential family. This posterior distribution is then transformed back via LM and used to update the latent GP. From this approximate posterior GP we can draw samples and apply the original transformation $x(y) = g^{-1}(x)$ yielding a transformed sample in the desired space. Rather than using the full pipeline, we can skip the latent prior and replace it with a pseudo-likelihood (see Algorithm \ref{alg:LMGP1}).}
    \label{fig:LM_GP}
\end{figure}
%


There are various approaches to model latent Gaussian inference on non-Gaussian data (see Section \ref{sec:related_work} for comparison). 
Many of these approaches such as VI, EP and MCMC require multiple iterations of optimization or sampling. This is costly. Most of the fast one-step approaches based on approximate inference are specialized for specific kinds of data, e.g \citep{DirichletGPC2018} for classification or \citep{Jia2021CountGPs} for count data.

LM+GP is general in the sense that it can be applied to any likelihood that is a conjugate prior to an exponential family while still being as fast as the specialized one-step approaches. Furthermore, LM can be seen as a fast pre-processing step that makes it possible to apply conventional Gaussian inference tools to non-Gaussian data. 

There are two possible ways to combine LM with GPs. First, we can model the likelihood with an exponential family, then apply LM to transform the likelihood to a Gaussian and then use standard GP inference (see Algorithm \ref{alg:LMGP1}). In this case, the exponential family is fitted by creating pseudo-observations for each data point or multiple data points together. We choose the exponential family distributions such that their respective likelihoods are conjugate to them. Since LM is very fast, the transformation cost is effectively negligible and therefore allows to fit GPs on non-Gaussian data with nearly zero additional cost (see Section \ref{subsec:timing} for details). 

Alternatively, we could define a latent GP, transform the GP prediction at the training datapoints via LM, update the resulting exponential family via conjugacy and transform the updated EF back via the inverse LM map (see Algorithm \ref{alg:LMGP2} for details). This more elaborate form may be preferable in cases where the prior GP is highly structured due to concrete prior knowledge.

For this paper, we use Version 1 of the algorithm because it is simpler and faster. We call it ``LM+GP'' for brevity.

In both versions, the GP posterior in the latent domain can then be transformed into a posterior in the data domain either by applying $x(y)$ to samples from the latent Gaussian or by drawing latent sample functions and transforming them individually through LM$^{-1}$. Figure \ref{fig:LM_GP} shows a high-level summary of the concept for the entire inference pipeline. 

Per default, we model the data points with pseudo-likelihoods, e.g. in binary classification, a data point with label 1 would be modeled as $\mathcal{B}(1+\eps_a, \eps_a)$. 
One of the advantages of choosing exponential families for the pseudo-likelihoods is that we can easily group data points together via conjugacy. If, for example, we have many data points close in time or that share the same cluster after k-means clustering, we can summarize them in the same instance of the respective exponential family. This yields a natural choice for grouping many data points into a few inducing points for GP inference. 

Since LM yields both an estimate for the mean and the covariance per inducing point, we can use a heteroskedastic noise model for our GP, i.e. we can use the measured covariance from LM as an approximation of the noise variable of the GP.  
\section{Related Work}
\label{sec:related_work}
The original idea to use basis changes to increase the quality of Laplace approximations goes back to \citet{MacKay1998}, who specifically studied Laplace approximations of Dirichlet distributions. This was extended to a bi-directional map by \citet{Hennig2010}. This work generalizes this idea to most of the popular exponential family distributions, provides explicit parameter mappings, and analyzes their costs and approximation quality. 

There is a host of other approximate inference schemes for latent Gaussian models, including a Laplace approximation on the joint posterior \citep{WilliamsBarberGPClassification1998}, Expectation Propagation \citep{MinkaEP2001}, Markov Chain Monte Carlo (MCMC) \citep{Neal93probabilisticinference, murray2009elliptical}, variational approximations \citep{VISeeger2009, challis2013gaussian}, and Integrated Nested Laplace Approximations (INLA) \citep{INLA2009, INLA2019_integrated}.

All approximate inference schemes ultimately strike a trade-off between approximation quality and speed. MCMC methods are asymptotically correct in the limit of infinitely many samples. Variational approximations converge to an approximation of finite error in finite time. Laplace matching arguably lies at the extreme end of this spectrum, providing a closed-form  approximation of fixed quality. In the experiments below we argue that the resulting approximation error is nevertheless often so small that the very low computational cost makes LM a feasible alternative to the previously mentioned methods. 
Since LM is so cheap (see Section \ref{subsec:timing}), it offers itself as a natural first choice, e.g. for testing purposes, and for low-level implementations where computational efficiency is paramount.

%
\section{Experiments}
\label{sec:experiments}
We first provide a number of experiments that explain the setup and provide quantitative results on four different non-Gaussian exponential families, namely the Beta (\Cref{subsec:Beta_experiment}), Gamma (\Cref{subsec:Gamma_experiment}), Dirichlet (\Cref{subsec:Dirichlet_experiment}) and inverse Wishart (\Cref{subsec:Wishart_experiment}).
Then we measure the time it took for LM to pre-process the data in all experiments in (\cref{subsec:timing}) to showcase its speed.
Finally, we investigate the quality of the proposed LM approximations by measuring the local KL divergence (\Cref{subsec:distance_measures}). 
Code for all experiments is available.\footnote{\url{https://github.com/mariushobbhahn/Laplace_Matching_for_GLMs}}
%

%
\subsection{Binary Classification (Beta)}
\label{subsec:Beta_experiment}
\begin{figure}
    \centering
    \includegraphics[width=\textwidth]{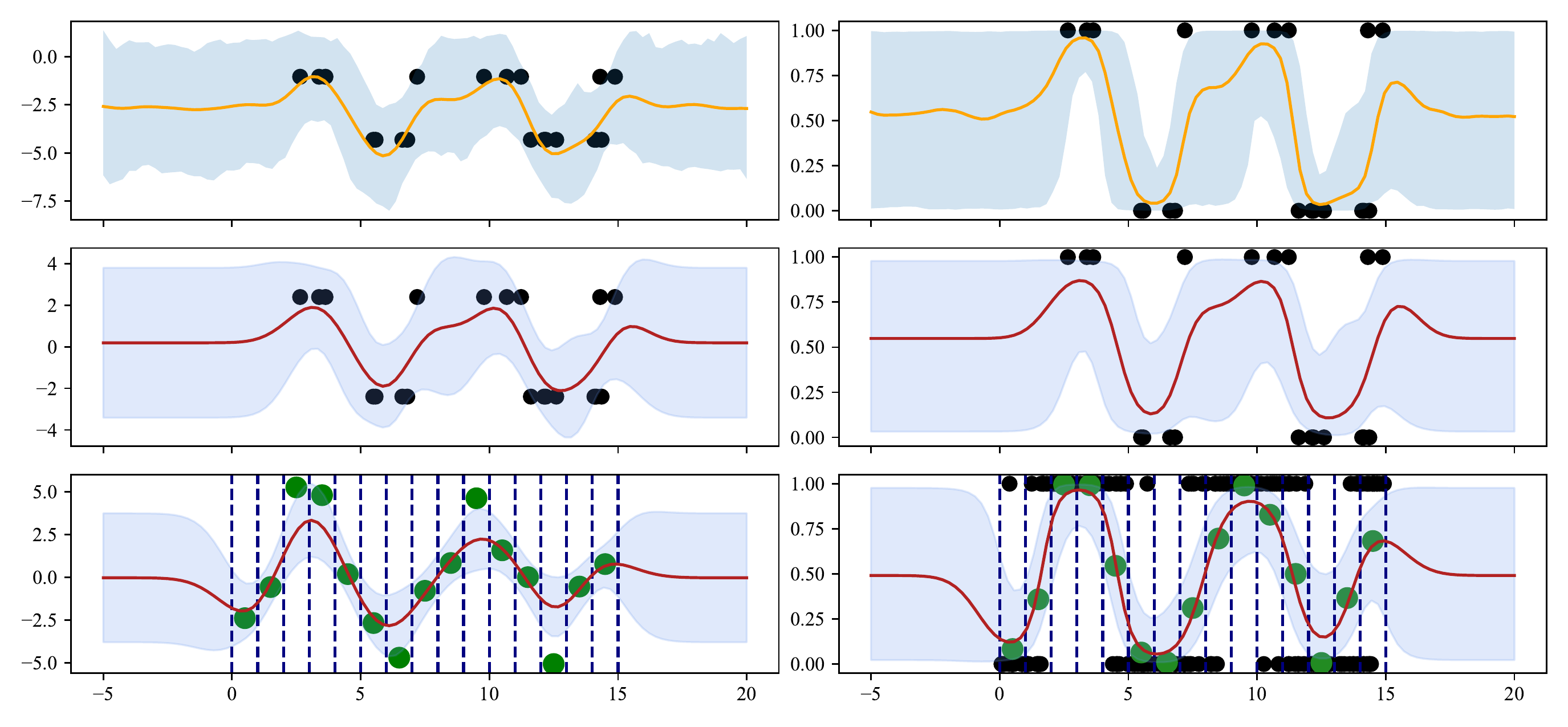}
    \caption{Conceptual figure for DirichletGPC and LM(Beta)+GP; \textbf{Left:} latent GP in $\mathbb{R}$; \textbf{Right:} transformed GP on the 1-simplex; \textbf{Top:} DirichletGPC; \textbf{Middle:} LM(Beta)+GP; \textbf{Bottom:} LM(Beta)+GP using the conjugacy of the Beta to group datapoints that are close in time into one pseudo-likelihood}
    \label{fig:beta_conceptual}
\end{figure}

First, we use binary classification to showcase LM+GP in a simple setting. The predictions for binary classification are probabilities. Thus, training a vanilla GP will yield predictions in the wrong domain. However, we can translate binary labels into Beta pseudo-likelihoods by adding a small value $\eps_a$ to both parameters such that a data point with label 1 would be modeled as $\mathcal{B}(1+\eps_a, \eps_a)$. We can use LM to map these pseudo-likelihoods to Gaussians and train a vanilla GP on the resulting means and variances (see Figure \ref{fig:beta_conceptual}). 
\begin{equation}
    \mu = \log\left(\frac{\alpha}{\beta}\right) \qquad\qquad
    \sigma^2 = \frac{\alpha + \beta}{\alpha\beta}
\end{equation}
Since LM yields estimates for mean and variance, we can use the $\sigma^2$ to specify a heteroskedastic noise model. By utilizing the conjugacy of the Beta distribution, we can group multiple classes together, e.g. when they are close in time (see bottom panel in Figure \ref{fig:beta_conceptual}). By grouping together multiple datapoints we can use conjugacy to create one ``representative'' for each cluster and use them as inducing points. 

For predictions, we can either draw samples and transform them using the logistic function or translate the GP back via LM.
We use the toy example from \citet{DirichletGPC2018} to conceptually compare LM(Beta)+GP with DirichletGPC which is another fast way of doing GP classification. In Figure \ref{fig:beta_conceptual} we can see that LM(Beta)+GP produces estimates of similar quality to DirichletGPC but additionally allows for a simple way of aggregating inducing points. 

\subsubsection{quantitative experiments}
\begin{figure}
    \centering
    \includegraphics[width=\textwidth]{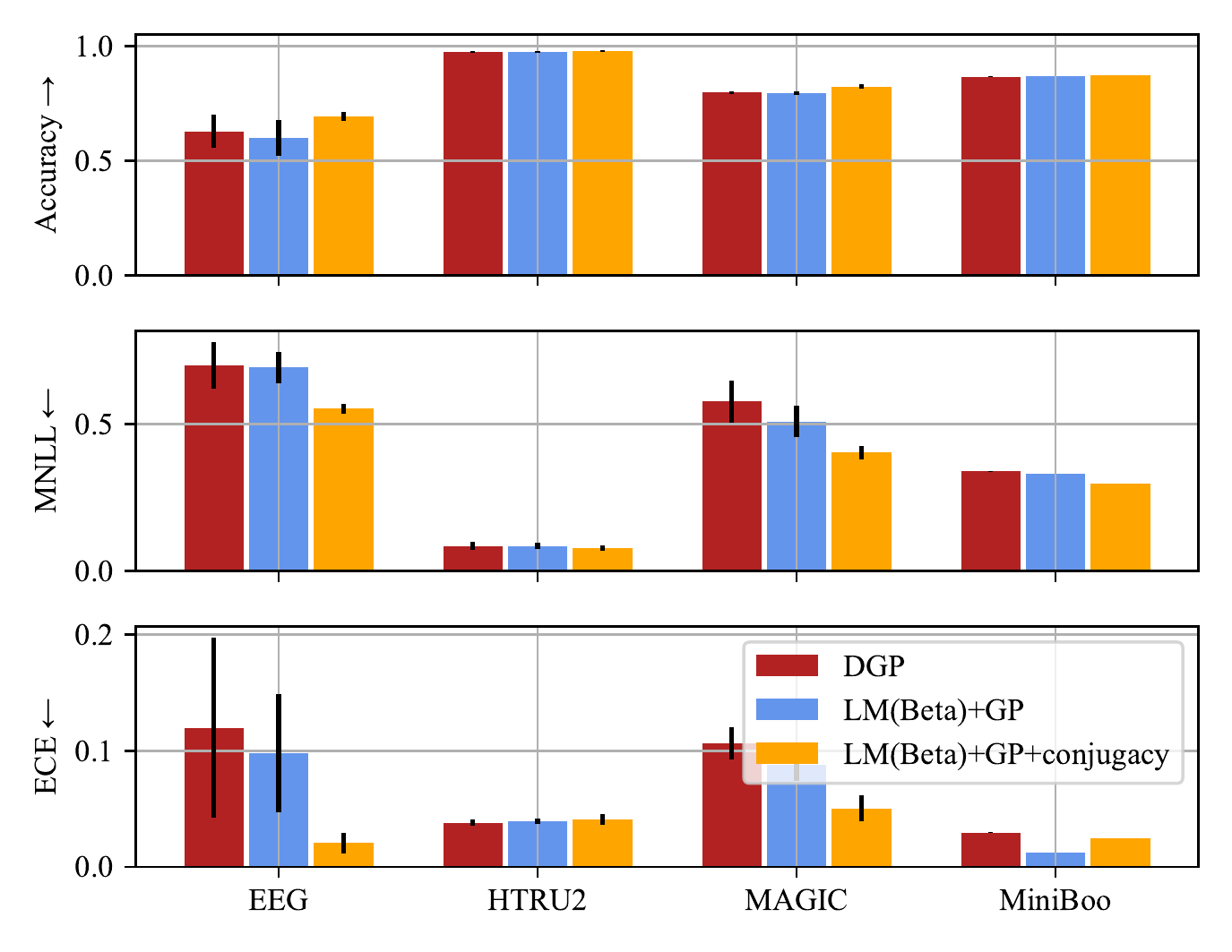}
    \caption{Results for binary classification experiment: LM(Beta)+GP is often competitive with DGPC, and outperforms both alternatives when the conjugacy of the Beta is leveraged to create inducing points.}
    \label{fig:bin_class_results}
\end{figure}

Furthermore, we compare DirichletGPC with LM(Beta)+GP on four benchmarks from the UCI ML repository \citep{UCI_ML} that were already used in \citep{DirichletGPC2018} (see Table \ref{tab:classification_overview}). For all experiments classification experiments, we use an RBF kernel.  
\begin{table}[]
    \centering
    \begin{tabular}{l  r  r  r  r  r}
         \toprule
         Dataset & Classes & Training instances & Test instances & Dimensionality & Inducing points\\
         \midrule
         EEG & 2 & 10980 & 4000 & 14 & 200 \\
         HTRU2 & 2 & 12898 & 5000 & 8 & 200 \\
         MAGIC & 2 & 14020 & 5000 & 10 & 200 \\
         MINIBOO & 2 & 120064 & 5000 & 50 & 400 \\
         LETTER & 26 & 15000 & 2000 & 16 & 200 \\
         DRIVE & 11 & 48509 & 2000 & 48 & 500 \\
         MOCAP & 5 & 68095 & 2000 & 37 & 500 \\
         \bottomrule
    \end{tabular}
    \caption{\textbf{Dataset specifications for classification experiments:} Adapted from \citep{DirichletGPC2018}. The experiments with binary classification can be found in the Beta section and multi-class classification can be found in the Dirichlet section.}
    \label{tab:classification_overview}
\end{table}
All inducing points are the centers of k-means clustering applied to the training data set. We compare the results using the three common metrics of accuracy, mean negative log-likelihood (MNLL) and expected-calibration error (ECE). The results can be found in Figure \ref{fig:bin_class_results}. We find that LM(Beta)+DP is competitive with DirichletGPC and that LM(Beta)+GP using the conjugacy of the Beta to create inducing points often outperforms DirichletGPC. 
Since \citet{DirichletGPC2018} have extensively compared DirichletGPC with many other methods for binary GP classification, we argue that LM(Beta)+DP is also competitive with their comparisons. 
\subsection{Count data (Gamma)}
\label{subsec:Gamma_experiment}
\begin{figure}[!tb]
    \setlength{\figwidth}{0.49\textwidth}
    \setlength{\figheight}{0.2\textheight}
    \centering
    \scriptsize
    \includegraphics[width=0.49\textwidth]{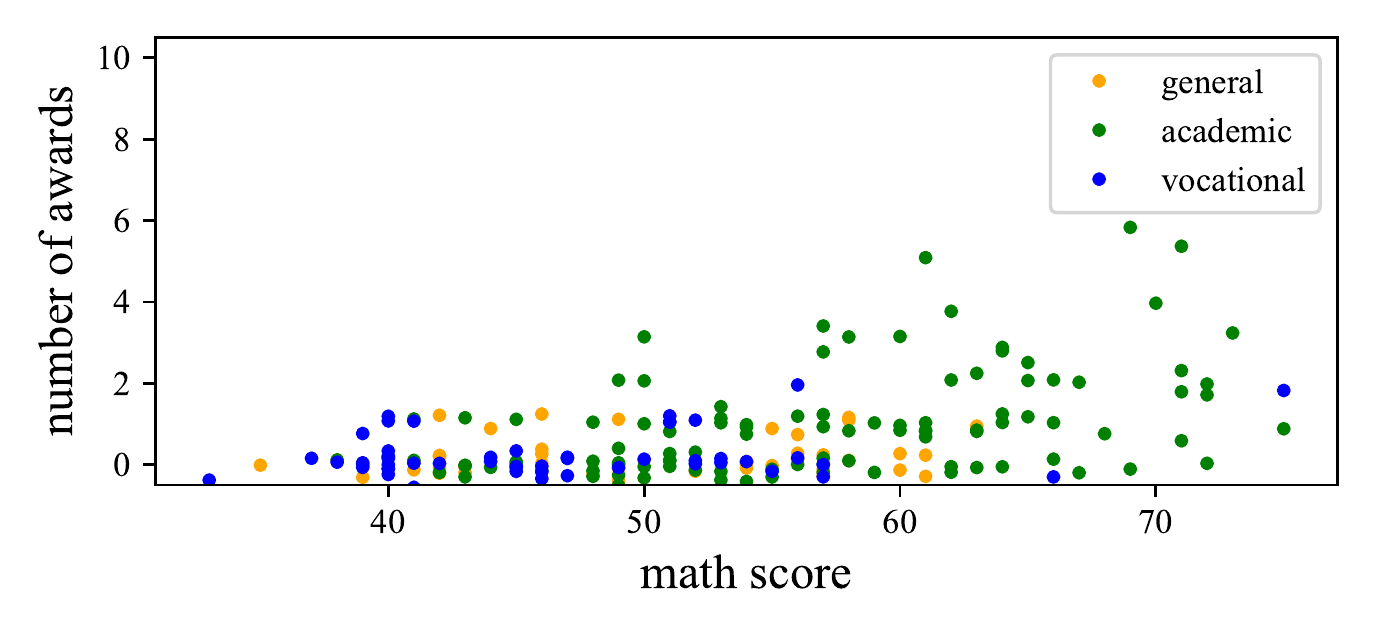}
    \includegraphics[width=0.49\textwidth]{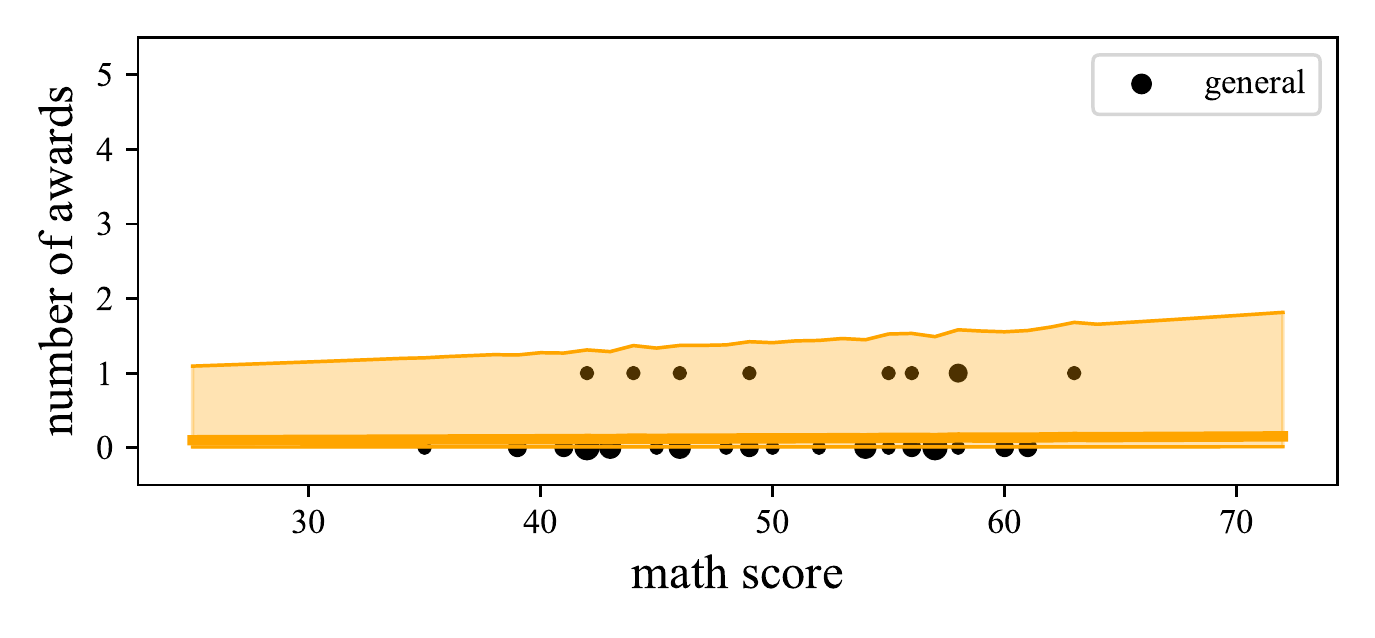}
    
    \includegraphics[width=0.49\textwidth]{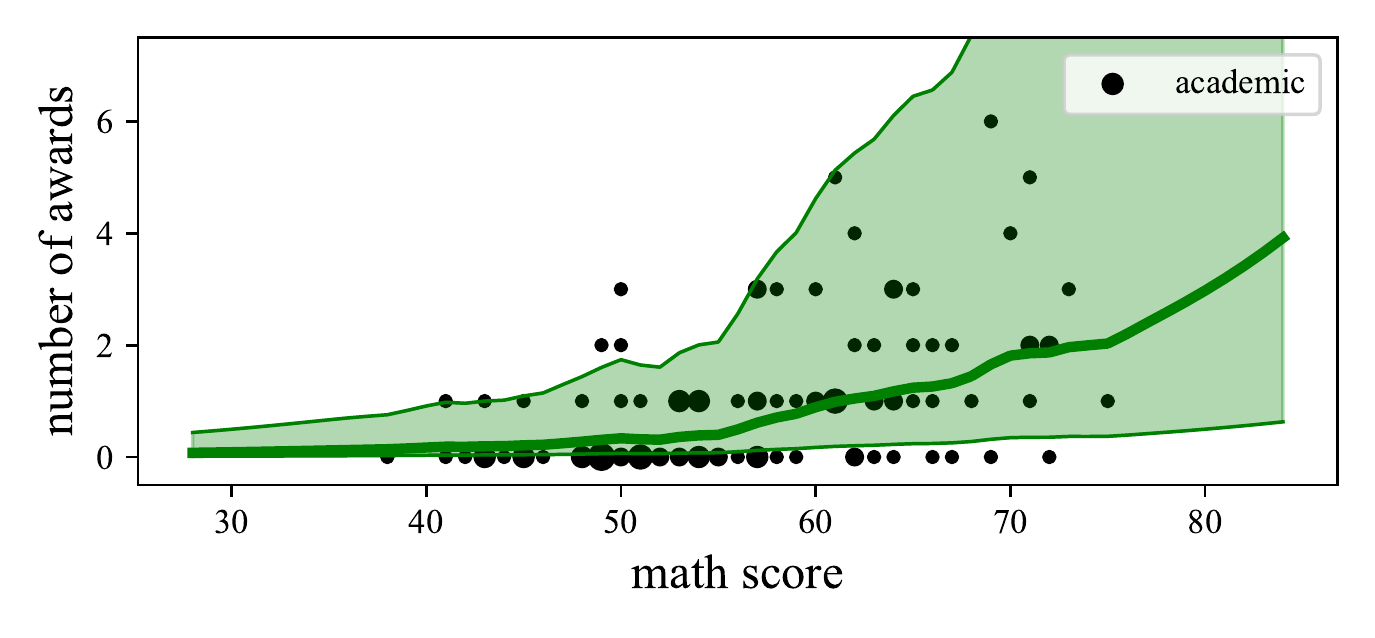}
    \includegraphics[width=0.49\textwidth]{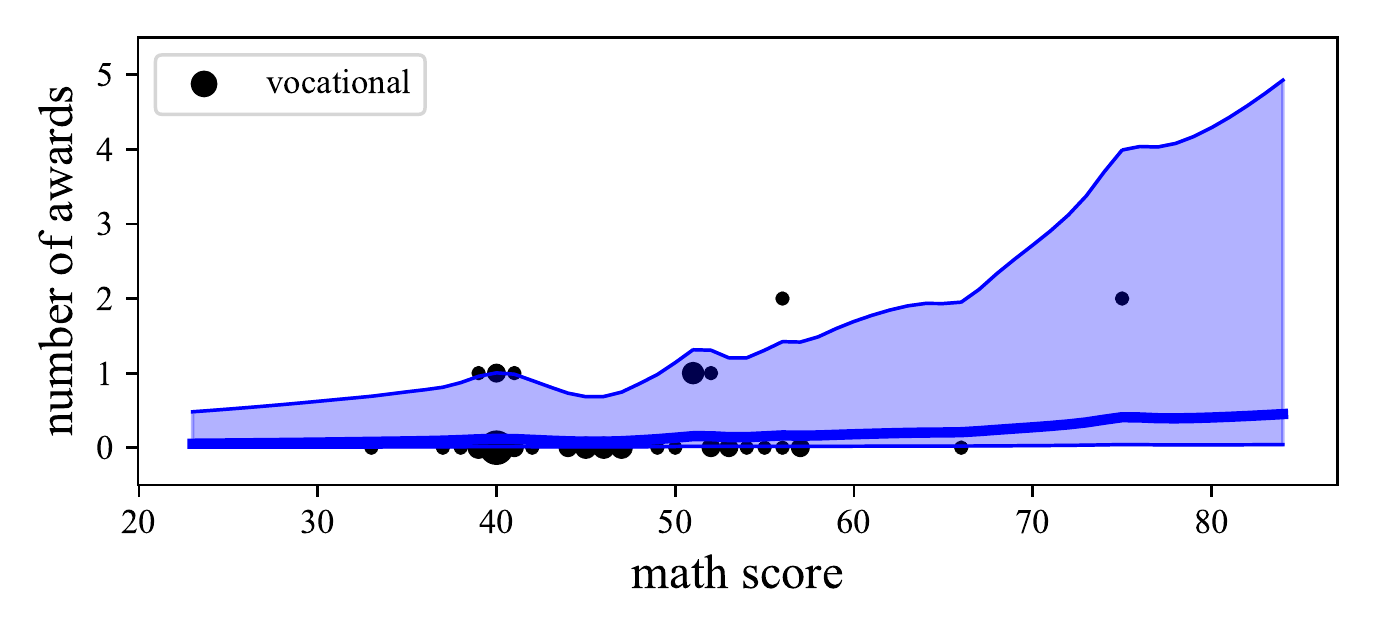}

    \caption{Probability of award as a function of math score. Top left: ground truth data (integers) with small noise added for visibility. The three other figures show LM+GP with count data and Gamma conjugate prior for the three classes respectively. The shadings in the figure show the \emph{transformed} Gaussian measure on the domain $x = \exp(y)$ (hence the constraint to positive values).}
    \label{fig:Poisson_GP}
\end{figure}

In this section, we first show how to apply LM(Gamma)+GP on a toy dataset and then compare it on a set of benchmarks to a host of alternatives. 

\subsubsection{Toy dataset}
\label{subsec:toy_dataset_counts}

We use a dataset from \citet{PoissonUCLA} containing integer counts reflecting the number of awards and math scores of students. The students belong to three groups: general, academic, and vocational. Since the number of awards are integers and reflect rare events, we assume a Poisson likelihood. The Gamma distribution is the conjugate prior to Poisson count data and its parameters can therefore be updated through simple addition. Thus we can create a marginal Gamma pseudo-likelihood distribution $p(x(y))$ at each math score for all three classes. Through LM in the log-base
\begin{equation}
\mu = \log\left(\frac{\alpha}{\lambda}\right) \qquad\qquad
\sigma^2 = \frac{1}{\alpha}
\end{equation}
we transform this Gamma distribution into a latent space $p(y)$ where a GP $q(y)$ is constructed. As a kernel function, we choose a sum of a rational quadratic and a linear kernel. Through the inverse transformation we are able to represent the GP posterior back in the original space,~i.e.~$\mathbb{R}_+$, by transforming the mean function and two standard deviations with the exponential function. 
The result is a GP $q(x(y))$ defined only on the positive domain and fitted to Poisson likelihoods.
The advantage of this simple yet fast procedure is that we can use the benefits of GPs on non-Gaussian data with nearly no additional cost. The resulting GPs can be found in Figure \ref{fig:Poisson_GP}.

\subsubsection{Quantitative experiments}
\label{subsec:count_data_quantitative}

We compare LM(Gamma)+GP on a set of 5 different datasets from \citep{AER_R_data} where the target variable is a non-negative integer. LM(Gamma)+GP is a good fit for these variables since the Gamma distribution is defined on $\mathbb{R}^+$. We compare LM+GP with a vanilla GP(ignoring the data domain), a GP trained on log-transformed data and SVIGP on log-transformed data. To evaluate our findings, we use three metrics: RMSE, mean NLL (using Poisson likelihoods) and the number of test-points within two standard deviations of the mean posterior prediction (in2std). The results can be found in Table \ref{tab:experiments_count_data}. We find that LM(Gamma)+GP is comparable with or better than alternative methods in many cases. 
\begin{table}[ht!]
    \scriptsize
    \fontsize{7}{8}\selectfont
    \setlength{\tabcolsep}{1.5pt}
    \centering
    \caption{Quantitative results on count data. We find that GP+LM is either the best or comparable with alternative methods in many cases. 
    }
    \begin{tabular}{l | rrr |rrr |rrr | rrr}
    \toprule
    & \multicolumn{3}{c}{\textbf{vanilla GP}} & \multicolumn{3}{c}{\textbf{log GP}} & \multicolumn{3}{c}{\textbf{LM(Gamma)+GP}} &\multicolumn{3}{c}{\textbf{SVIGP}} \\
    \textbf{Dataset} & \textbf{RMSE} $\downarrow$ & \textbf{MNLL} $\downarrow$& \textbf{in2std} & \textbf{RMSE} $\downarrow$ & \textbf{MNLL} $\downarrow$& \textbf{in2std} & \textbf{RMSE} $\downarrow$ & \textbf{MNLL} $\downarrow$& \textbf{in2std} & \textbf{RMSE} $\downarrow$ & \textbf{MNLL} $\downarrow$& \textbf{in2std} \\
    \midrule
    CreditCard &   0.963 &   1.265 &  0.991 &   1.521 &   11.818 &  0.837 &   1.301 &   1.490 &  0.344 &   1.304 &   1.576 &  0.275 \\
    Doctor &   0.851 &   1.247 &  0.982 &   0.910 &    4.618 &  0.265 &   0.780 &   0.704 &  0.169 &   0.811 &   0.724 &  0.008 \\
    Citations &  57.996 &  46.393 &  0.256 &  66.491 &  104.156 &  0.047 &  41.556 &  15.038 &  0.972 &  51.950 &  26.337 &  0.888 \\
    GSS7402 &   1.255 &   1.628 &  0.949 &   2.249 &    2.921 &  0.253 &   1.262 &   1.633 &  0.428 &   1.312 &   1.656 &  0.207 \\
    Medicaid &   3.662 &   2.665 &  0.948 &   4.120 &   22.337 &  0.748 &   3.590 &   2.580 &  0.489 &   3.831 &   3.367 &  0.132 \\
    \bottomrule
    \end{tabular}
    \label{tab:experiments_count_data}
\end{table}
\subsection{Multi-class classification (Dirichlet)}
\label{subsec:Dirichlet_experiment}
\begin{figure}[!tb]
    \centering
    
     
    
    \includegraphics[width=0.49\textwidth]{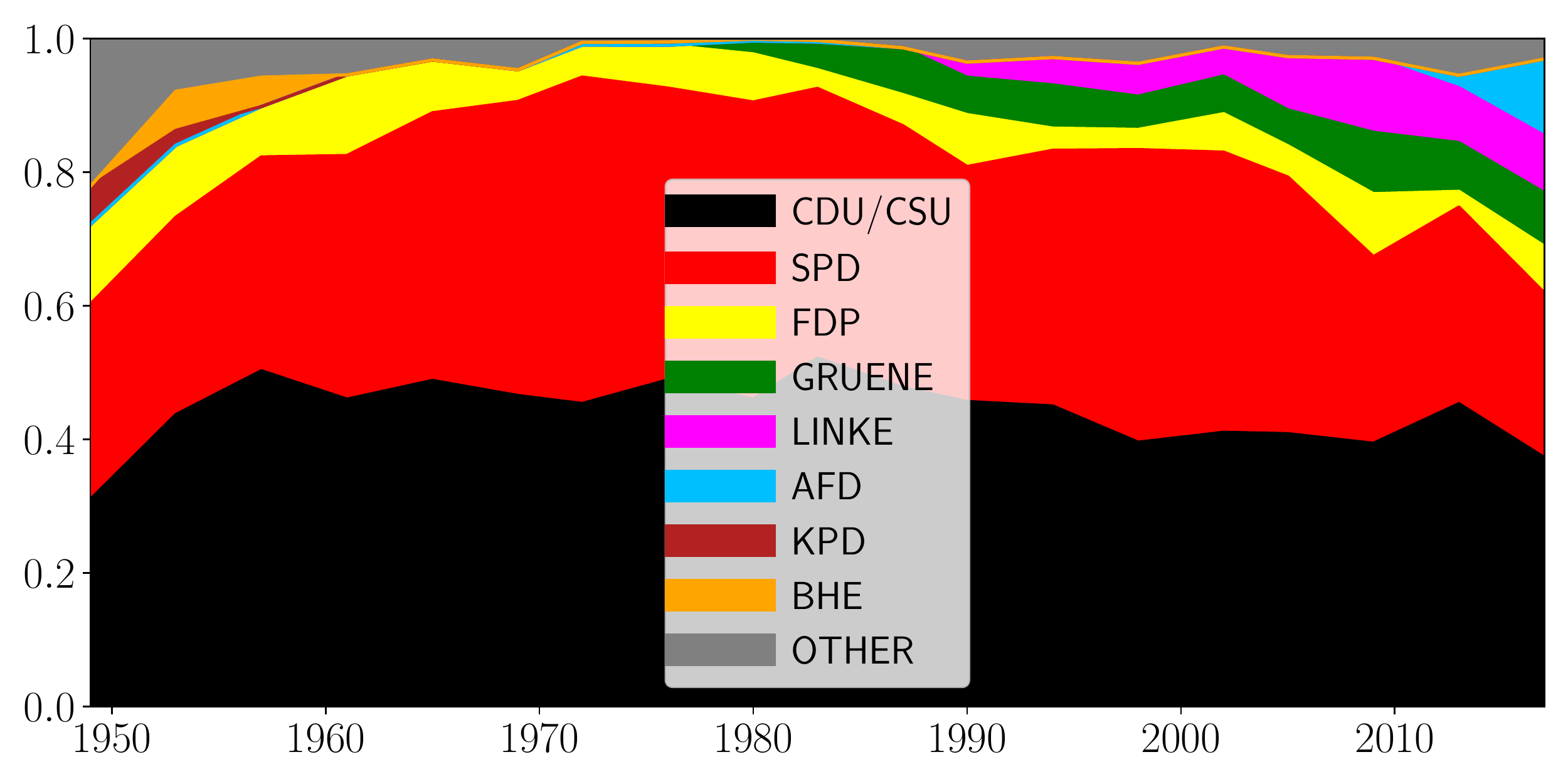} \hfill
    \includegraphics[width=0.49\textwidth]{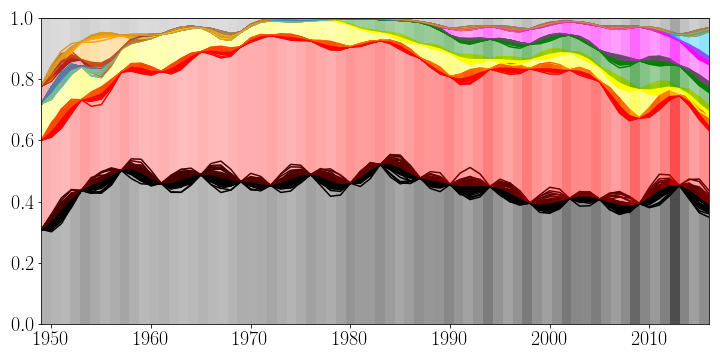}
    
    \includegraphics[width=0.49\textwidth]{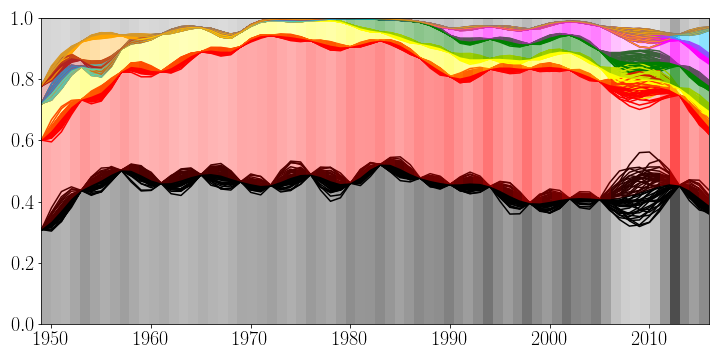}
    \includegraphics[width=0.49\textwidth]{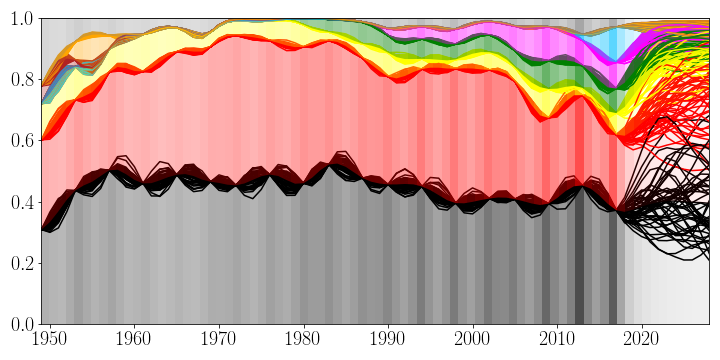}

    \caption{German elections from 1949 to 2017. Top left: Ground truth data points; Top right: LM+GP cumulative predictions for all parties with attached uncertainty (dark means low uncertainty); Bottom left: 2009 election has been left out of the training data; Bottom right: predictions for the next 3 elections (2021, 2025, 2029). The uncertainties are meaningful, e.g.~the existence of data implies low uncertainty and vice versa.}
    \label{fig:german_elections_GP}
\end{figure}
To showcase the application of LM+GP to categorical data we use the results from the German federal elections for the entirety of Germany between 1949 and 2017. The data contain $T$ different dates for the respective elections, $P$ different parties, and $C$ different election counties. Every data point is a tuple $(\mathsf{t}, \mathsf{p}, \mathsf{c})$ with an associated vote count $\mathsf{v}$ for party $\mathsf{p}$ in election year $\mathsf{t}$ and county $\mathsf{c}$. For every year $\mathsf{t}$ and county $\mathsf{c}$ we can create a Dirichlet distribution given the $\mathsf{p}$ parties through $p(x(y)) = \mathcal{D}_{t,c}(v_p)$ since the votes form a categorical distribution for which the Dirichlet is a conjugate prior. To create a latent Gaussian distribution $q(y)$ from the Dirichlet and vice versa we use the LM equations
\begin{subequations}
\begin{align}
    \mu_k &= \log \alpha_k  - \frac{1}{K} \sum_{l=1}^{K} \log \alpha_l \, , \label{eq:mubridge_text}\\
    \Sigma_{k\ell} &= \delta_{k\ell}\frac{1}{\alpha_k} - \frac{1}{K}\left[\frac{1}{\alpha_k} + \frac{1}{\alpha_\ell} - \frac{1}{K}\sum_{u=1} ^K \frac{1}{\alpha_u} \right] \label{eq:Sigmabridge_text}\\
    \alpha_k &= \frac{1}{\Sigma_{kk}}\left(1 - \frac{2}{K} + \frac{e^{\mu_k}}{K^2}\sum_{l=1}^K e^{-\mu_l} \right) \,.
\end{align}
\end{subequations}
to yield $\mu_{\mathsf{t},\mathsf{c}}$ and $\Sigma_{\mathsf{t},\mathsf{c}}$. Since some parties received 0 votes in some elections we set the prior of each Dirichlet to $\alpha=1$.
The likelihood for all election data under the latent Gaussian is then given by
\begin{equation}
    q(y) = LM^{-1}(p(v|\pi_d)) = \prod_d LM^{-1}(\mathcal{D}(v_d))) = \prod_d \mathcal{N}(f; \mu_d, \Sigma_d),
\end{equation}
where $f$ describes the latent Gaussian.
Integrating all of the above into the general equation for GP regression \citep{Rasmussen06gaussianprocesses}
\begin{equation}
    q(f|v) = \mathcal{GP}\left(f; m_k + k_{xX}(k_{XX} + \Sigma_{X})^{-1}(\mu_X - m_d), k_{xx} - k_{xX}(k_{XX} + \Sigma_{X})^{-1}k_{Xx} \right)
\end{equation}
we can construct a $N=T \cdot P \cdot C$ dimensional GP over all combinations of elections, parties and counties. The kernel function $k$ is defined as
\begin{equation}
    k[(\mathsf{t},\mathsf{p},\mathsf{c}), (\mathsf{t}',\mathsf{p}',\mathsf{c}')] = k_T(\mathsf{t}, \mathsf{t}') \cdot k_P(\mathsf{p},\mathsf{p}') \cdot k_C(\mathsf{c}, \mathsf{c}')
    \label{eq:kernel_Dirichlet}
\end{equation}
where the kernels $k_T, k_P$ and $k_C$ can be hand-crafted to their specific domain. The party kernel $k_P$ could, for example, display the distance of parties on the political spectrum or the county kernel $k_C$ could incorporate information about the relative differences in average income, education, crime, etc.~as long as they correlate with voting behavior. The kernel matrix $k_{XX} \in \mathbb{R}^{N\times N}$ is created through Equation \ref{eq:kernel_Dirichlet}. The noise matrix $\Sigma_{X} \in \mathbb{R}^{N\times N}$ is a block diagonal matrix with $T\cdot C$ blocks of size $P\times P$ containing the respective $\Sigma_{\mathsf{t},\mathsf{c}}$ from the Laplace Matching.

In our experiment, we restrict ourselves to the $T=19$ elections held between 1949 and 2017, to the $P=9$ political parties that gained more than $5\%$ of the vote at least once in these elections, and to the $9$ federal states that continuously existed across this time-frame -- plus Germany as a whole, such that $C=10$.

After building the GP we can draw sample functions at different points in time and transform them back to probability space through the softmax function $\sigma(y) = \frac{\exp(y_i)}{\sum_j \exp(y_j)}$ to get our GP predictions.
The resulting election landscape can be found in Figure \ref{fig:german_elections_GP}.

\subsubsection{Timing}
To evaluate the speedup of LM+GP over alternatives, we compare LM+GP with elliptical slice sampling \citep{murray2009elliptical} (ESS). ESS is an efficient sampling method specifically designed for the combination of non-Gaussian likelihoods and GP priors. We calculate the log-likelihood for ESS with
\begin{align}
    l(\mathsf{t},\mathsf{c},\mathsf{p}) &= \log\left(\prod_\mathsf{t} \prod_\mathsf{c} \prod_\mathsf{p} \sigma(f(\mathsf{t},\mathsf{c}))_p^{y_{\mathsf{t},\mathsf{c},\mathsf{p}}}  \right) = \sum_\mathsf{t} \sum_\mathsf{c} \sum_\mathsf{p} y_{\mathsf{t},\mathsf{c},\mathsf{p}} \log(\sigma(f(\mathsf{t},\mathsf{c}))
\end{align}
The predictions with ESS are made by projecting the samples $\hat{f}$ to the desired space with
\begin{align}
    q(f_x | y, \hat{f}) &=  \sum_i \mathcal{N}\left(f_x; m_x + k_{xX}k_{XX}^{-1}(\hat{f}_i - m_x), k_{xx} - k_{xX}k_{XX}^{-1}k_{Xx}\right)
\end{align}

\setlength{\figwidth}{0.49\textwidth}
\setlength{\figheight}{0.22\textheight}
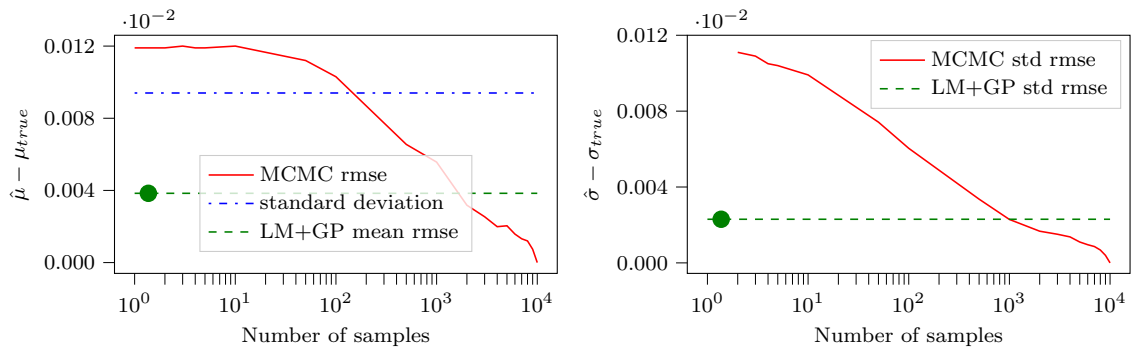
\begin{figure}[!tb]
    \centering

    \scriptsize
\begin{tikzpicture}

\begin{axis}[
height=\figheight,
legend cell align={left},
legend style={
  fill opacity=0.8,
  draw opacity=1,
  text opacity=1,
  at={(0.5,0.09)},
  anchor=south,
  draw=white!80!black
},
log basis x={10},
tick align=outside,
tick pos=left,
width=\figwidth,
x grid style={white!69!black},
xlabel={Number of samples},
xmin=0.631, xmax=1.58e+04,
xmode=log,
xtick style={color=black},
xtick={1,2,3,4,5,6,7,8,9,10,20,30,40,50,60,70,80,90,100,200,300,400,500,600,700,800,900,1e+03,2e+03,3e+03,4e+03,5e+03,6e+03,7e+03,8e+03,9e+03,1e+04},
xticklabels={
  \(\displaystyle 10^{0}\),
  ,
  ,
  ,
  ,
  ,
  ,
  ,
  ,
  \(\displaystyle 10^{1}\),
  ,
  ,
  ,
  ,
  ,
  ,
  ,
  ,
  \(\displaystyle 10^{2}\),
  ,
  ,
  ,
  ,
  ,
  ,
  ,
  ,
  \(\displaystyle 10^{3}\),
  ,
  ,
  ,
  ,
  ,
  ,
  ,
  ,
  \(\displaystyle 10^{4}\)
},
y grid style={white!69!black},
ylabel={\(\displaystyle \hat{\mu} - \mu_{true}\)},
ymin=-0.000599, ymax=0.0126,
ytick style={color=black},
ytick={0,0.004,0.008,0.012},
yticklabels={
  \(\displaystyle 0.000\),
  \(\displaystyle 0.004\),
  \(\displaystyle 0.008\),
  \(\displaystyle 0.012\)
}
]
\addplot [semithick, red]
table {%
1 0.0119
2 0.0119
3 0.012
4 0.0119
5 0.0119
10 0.012
50 0.0112
100 0.0103
500 0.00655
1e+03 0.00557
2e+03 0.00318
3e+03 0.00254
4e+03 0.00199
5e+03 0.00204
6e+03 0.00158
7e+03 0.00132
8e+03 0.0012
9e+03 0.000739
1e+04 0
};
\addlegendentry{MCMC rmse}
\addplot [semithick, blue, dash pattern=on 1pt off 3pt on 3pt off 3pt]
table {%
1 0.0094
2 0.0094
3 0.0094
4 0.0094
5 0.0094
10 0.0094
50 0.0094
100 0.0094
500 0.0094
1e+03 0.0094
2e+03 0.0094
3e+03 0.0094
4e+03 0.0094
5e+03 0.0094
6e+03 0.0094
7e+03 0.0094
8e+03 0.0094
9e+03 0.0094
1e+04 0.0094
};
\addlegendentry{standard deviation}
\addplot [semithick, green!50.2!black, dashed]
table {%
1 0.00384
2 0.00384
3 0.00384
4 0.00384
5 0.00384
10 0.00384
50 0.00384
100 0.00384
500 0.00384
1e+03 0.00384
2e+03 0.00384
3e+03 0.00384
4e+03 0.00384
5e+03 0.00384
6e+03 0.00384
7e+03 0.00384
8e+03 0.00384
9e+03 0.00384
1e+04 0.00384
};
\addlegendentry{LM+GP mean rmse}
\addplot [semithick, green!50!black, mark=*, mark size=3, mark options={solid}, only marks, forget plot]
table {%
1.37 0.00384
};
\end{axis}

\end{tikzpicture}
\begin{tikzpicture}

\begin{axis}[
height=\figheight,
legend cell align={left},
legend style={fill opacity=0.8, draw opacity=1, text opacity=1, draw=white!80!black},
log basis x={10},
tick align=outside,
tick pos=left,
width=\figwidth,
x grid style={white!69!black},
xlabel={Number of samples},
xmin=0.631, xmax=1.58e+04,
xmode=log,
xtick style={color=black},
xtick={1,2,3,4,5,6,7,8,9,10,20,30,40,50,60,70,80,90,100,200,300,400,500,600,700,800,900,1e+03,2e+03,3e+03,4e+03,5e+03,6e+03,7e+03,8e+03,9e+03,1e+04},
xticklabels={
  \(\displaystyle 10^{0}\),
  ,
  ,
  ,
  ,
  ,
  ,
  ,
  ,
  \(\displaystyle 10^{1}\),
  ,
  ,
  ,
  ,
  ,
  ,
  ,
  ,
  \(\displaystyle 10^{2}\),
  ,
  ,
  ,
  ,
  ,
  ,
  ,
  ,
  \(\displaystyle 10^{3}\),
  ,
  ,
  ,
  ,
  ,
  ,
  ,
  ,
  \(\displaystyle 10^{4}\)
},
y grid style={white!69!black},
ylabel={\(\displaystyle \hat{\sigma} - \sigma_{true}\)},
ymin=-0.000554, ymax=0.012,
ytick style={color=black},
ytick={0,0.004,0.008,0.012},
yticklabels={
  \(\displaystyle 0.000\),
  \(\displaystyle 0.004\),
  \(\displaystyle 0.008\),
  \(\displaystyle 0.012\)
}
]
\addplot [semithick, red]
table {%
2 0.0111
3 0.0109
4 0.0105
5 0.0104
10 0.00991
50 0.00742
100 0.00604
500 0.00336
1e+03 0.00229
2e+03 0.00167
3e+03 0.00151
4e+03 0.00137
5e+03 0.0011
6e+03 0.000958
7e+03 0.000861
8e+03 0.000679
9e+03 0.000398
1e+04 0
};
\addlegendentry{MCMC std rmse}
\addplot [semithick, green!50.2!black, dashed]
table {%
1 0.0023
2 0.0023
3 0.0023
4 0.0023
5 0.0023
10 0.0023
50 0.0023
100 0.0023
500 0.0023
1e+03 0.0023
2e+03 0.0023
3e+03 0.0023
4e+03 0.0023
5e+03 0.0023
6e+03 0.0023
7e+03 0.0023
8e+03 0.0023
9e+03 0.0023
1e+04 0.0023
};
\addlegendentry{LM+GP std rmse}
\addplot [semithick, green!50!black, mark=*, mark size=3, mark options={solid}, only marks, forget plot]
table {%
1.37 0.0023
};
\end{axis}

\end{tikzpicture}
    
    \caption{Comparison of approximation quality of LM vs MCMC-samples dependent on the number of samples. In both cases, the approximation quality of the samples meets that of LM after ca.~1000 samples. The green dot indicates the ratio of drawing one sample vs. applying LM. We see that LM costs slightly more than drawing one ESS sample to compute the latent Gaussian.}
    \label{fig:distances_elections_LM}
\end{figure}
To evaluate the timing trade-off between a fast but fixed approximation strategy such as LM and a more costly, but asymptotically exact iterative strategy such as ESS we compare the quality of the approximation on a significantly smaller data-subset, namely the local results (ca.~1000 votes per election) of one voting precinct (the ``Wanne'' neighborhood in Tübingen) from 2002 to 2017. The true distribution is approximated by drawing 10k samples from ESS. The results can be seen in Figure \ref{fig:distances_elections_LM}. We find that it takes around 1000 samples until the MCMC-samples yield the same quality of mean and variance compared to the static LM results. Additionally, we see that the difference of the mean from LM is smaller than one standard deviation from the true distribution. Since the standard deviation is a measure of aleatoric uncertainty of our posterior, being within one standard deviation from the true mean is a strong argument for the quality of LM.

\subsubsection{Quality of the Marginals}
\begin{figure}[!tb]
    \centering
    \input{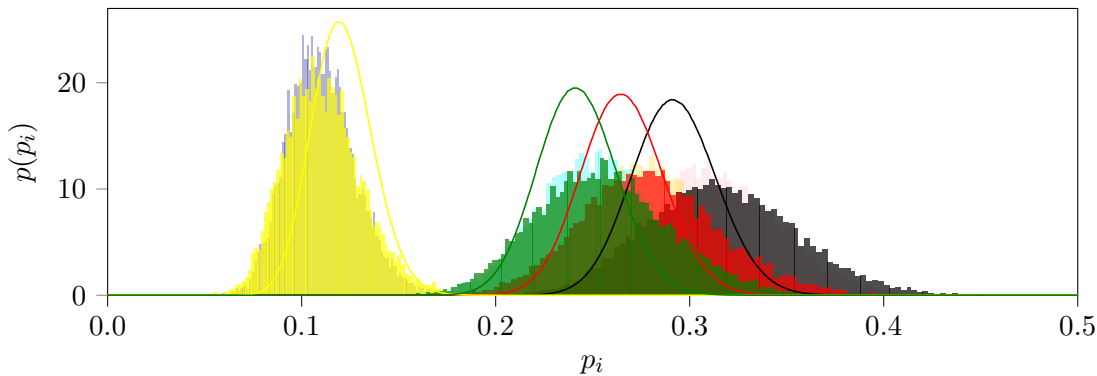}
    \caption{Marginal plot of the four largest parties for the prediction of a local election in 2004 (i.e.~when no actual election took place). We compare predictions of ESS samples (dark histograms) with transformed samples from LM+GP (light histograms) and Beta marginals from the LM-Dirichlet transformation (solid lines). The reasons for their differences are discussed in the text.}
    \label{fig:Tübingen_elections_marginals}
\end{figure}

To evaluate the quality of the prediction given by LM+GP compared to sampling-based methods we use their marginal predictions on the smaller dataset. We compare three different predictions: samples given by ESS, samples from LM+GP that are transformed via the softmax, and Beta marginals from the Dirichlet generated directly by the inverse LM transformation from the GP since variable $x_i$ from a Dirichlet is distributed according to Beta$(\alpha_i, \sum_{j\neq i} \alpha_j)$. A plot of the resulting marginals can be found in Figure \ref{fig:Tübingen_elections_marginals}. We find a) that the transformed samples from LM+GP are slightly different than the Beta marginals from the direct transformation. This is because the inverse Transformation for the LM-Dirichlet is only an approximation and not a bijective transformation. We find b) that the samples from ESS and the transformed samples from LM+GP are very similar but still show slight differences. This is likely due to the fact that we use a small uniform prior for LM+GP but none for ESS. This shifts smaller probabilities to the right and larger probabilities to the left, exactly as seen in Figure \ref{fig:Tübingen_elections_marginals}. While both approximations to ESS are not perfectly exact, they are traded off through computational speed. The LM-transformation costs $\mathcal{O}(n)$, sampling from LM+GP takes $\mathcal{O}(n^2)$ and sampling for ESS costs $\mathcal{O}(n^3)$.

\setlength{\figwidth}{0.95\textwidth}
\setlength{\figheight}{0.25\textheight}

\subsubsection{quantitative experiments}

\begin{figure}
    \centering
    \includegraphics[width=\textwidth]{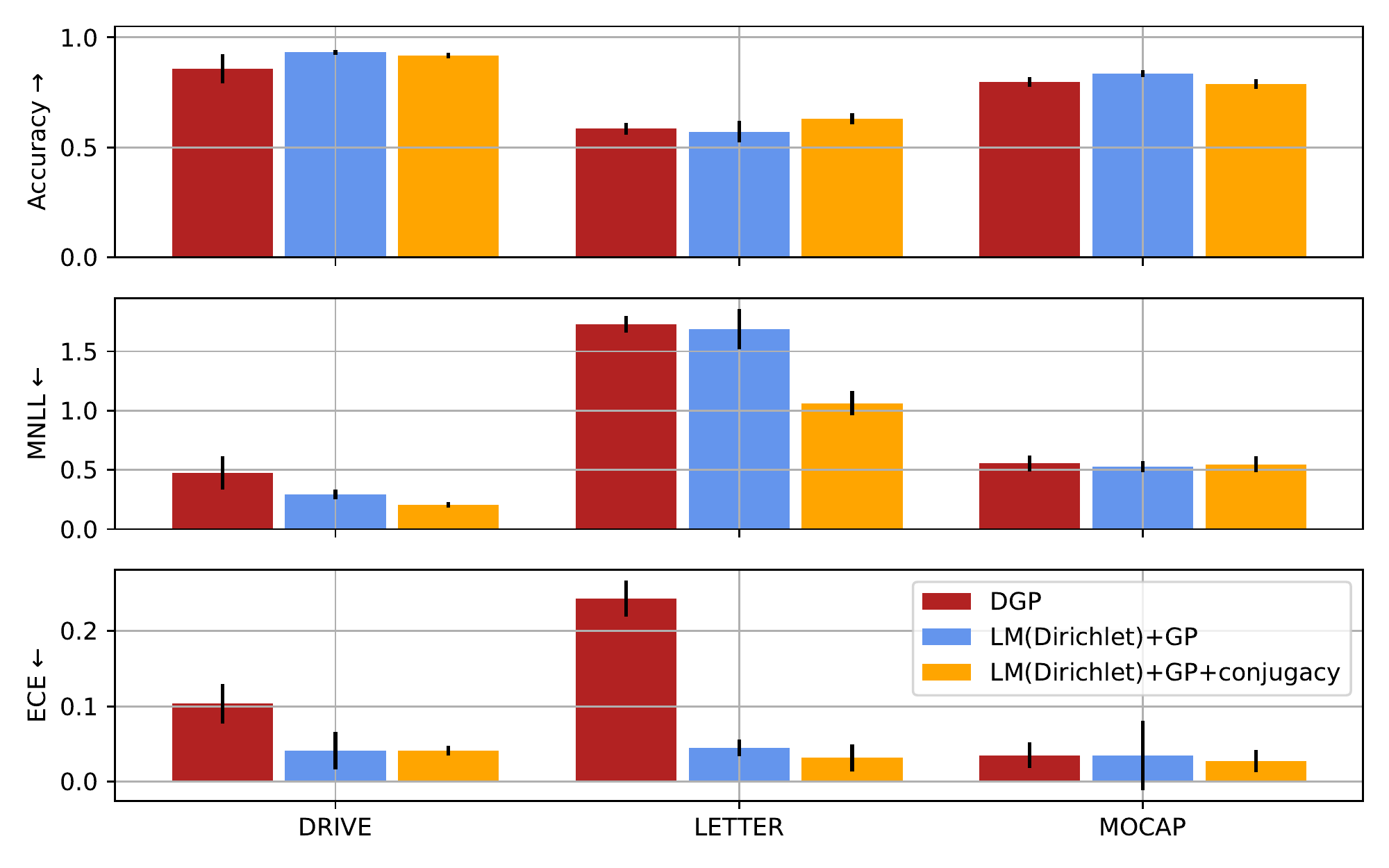}
    \caption{Results of the multiclass classification. We find that LM(Dirichlet)+GP is competitive or better than DirichletGPC.}
    \label{fig:mult_classification}
\end{figure}

In addition to the conceptual and timing experiments, we compare LM(Dirichlet)+GP with DirichletGPC \citep{DirichletGPC2018} on three of the benchmarks provided in their paper. The setup is similar to the Beta experiments detailed in Subsection \ref{subsec:Beta_experiment} and the details of the datasets can be found in Table \ref{tab:classification_overview}. The results of the experiments can be found in Figure \ref{fig:mult_classification}. We find that LM(Dirichlet)+GP is competitive or better than DirichletGPC.

\subsection{Currencies \& Covariances (Inverse Wishart)}
\label{subsec:Wishart_experiment}
\begin{figure}[!tb]
    \centering
    \vspace{-0.75em}
    \setlength{\figwidth}{\textwidth}
    \setlength{\figheight}{0.22\textheight}
    \includegraphics[width=\textwidth]{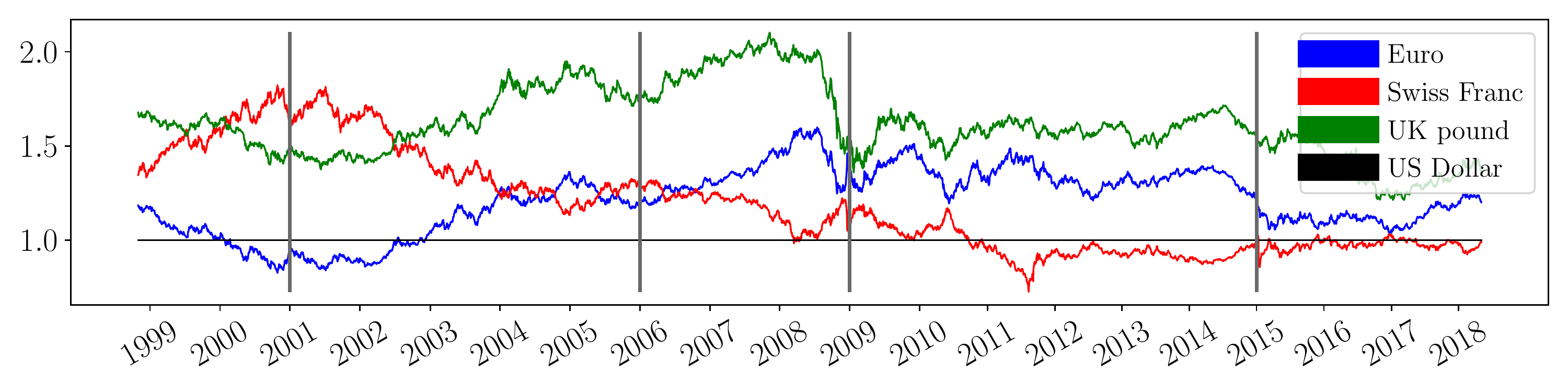} \\
    \vspace{-0.25em}
    \setlength{\figwidth}{\textwidth}
    \setlength{\figheight}{0.12\textheight}
    \includegraphics[width=\textwidth]{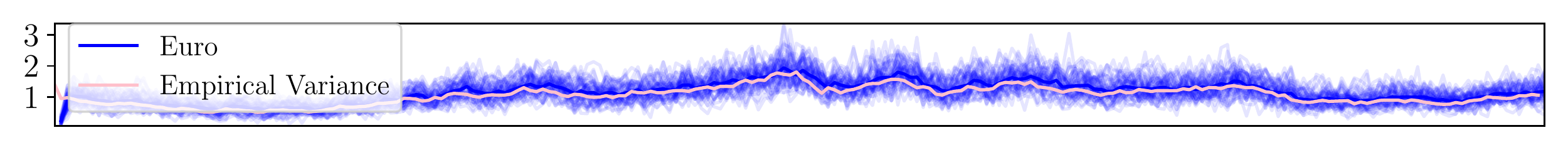} \\
    \vspace{-0.5em}
    \includegraphics[width=\textwidth]{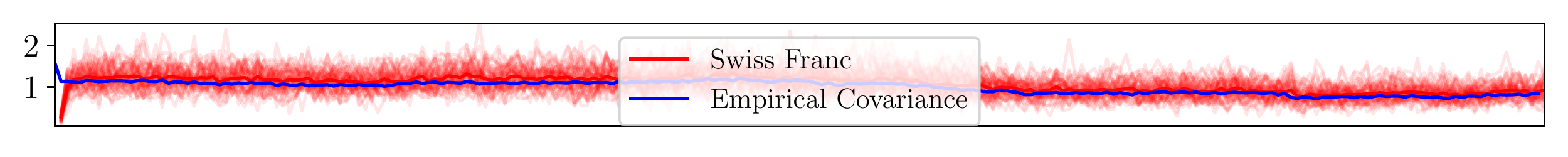} \\
    \vspace{-0.5em}
    \includegraphics[width=\textwidth]{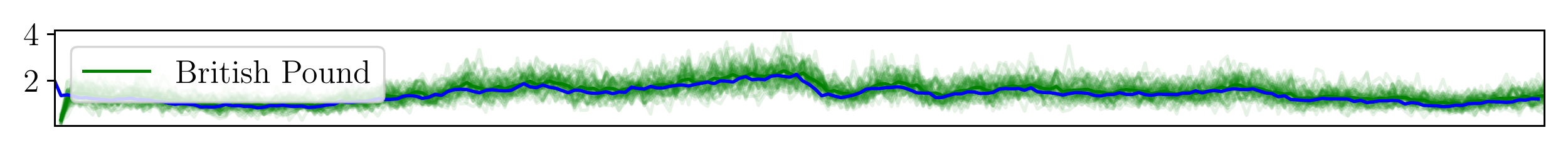} \\
    \vspace{-0.5em}
    \includegraphics[width=\textwidth]{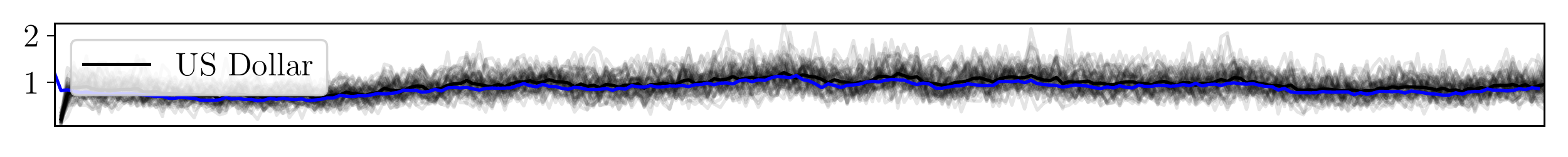} \\
    \vspace{-0.25em}
    \includegraphics[width=\textwidth]{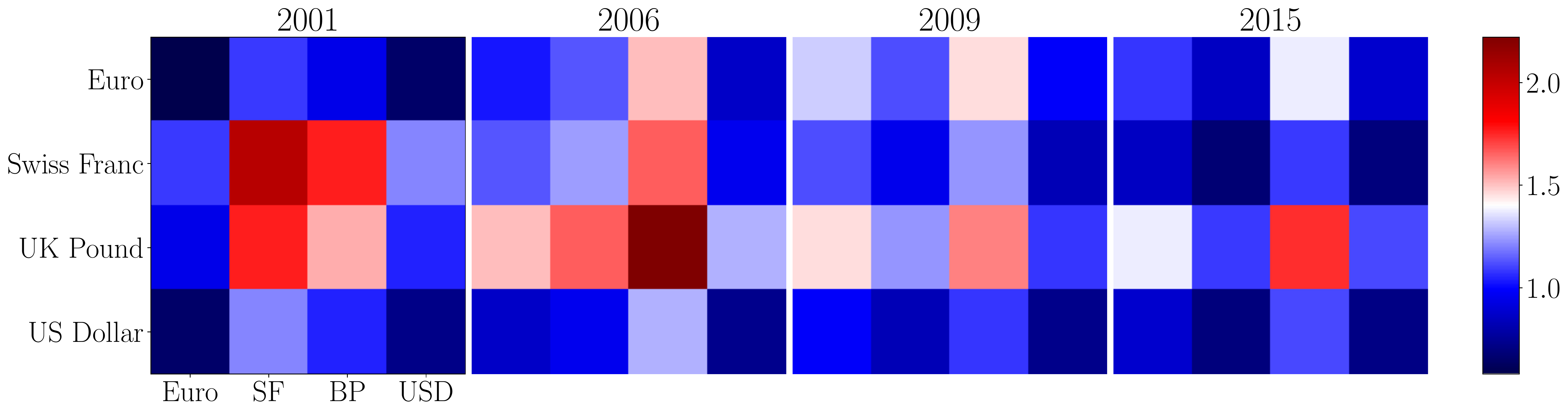} \\
    \includegraphics[width=0.24\textwidth]{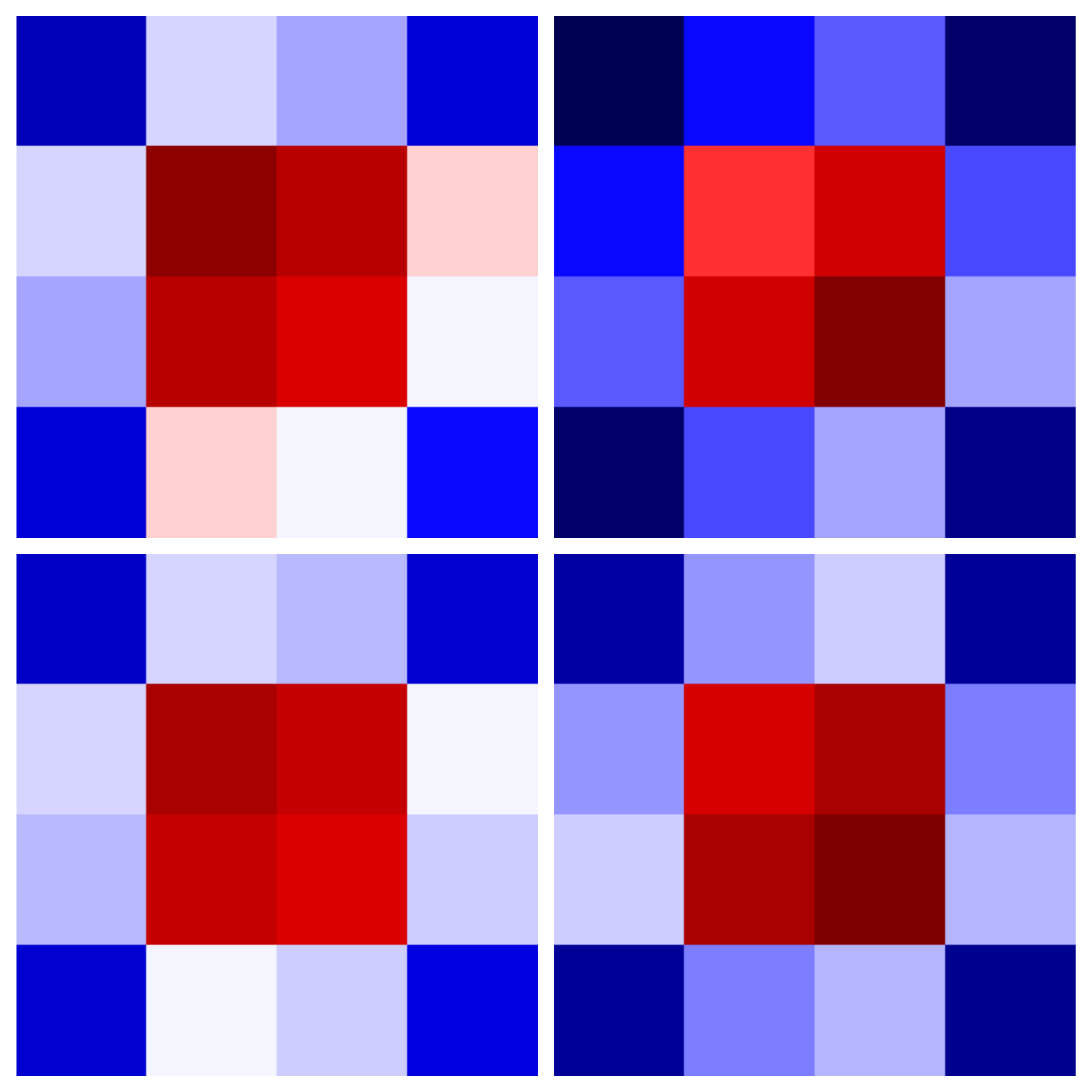}
    \includegraphics[width=0.24\textwidth]{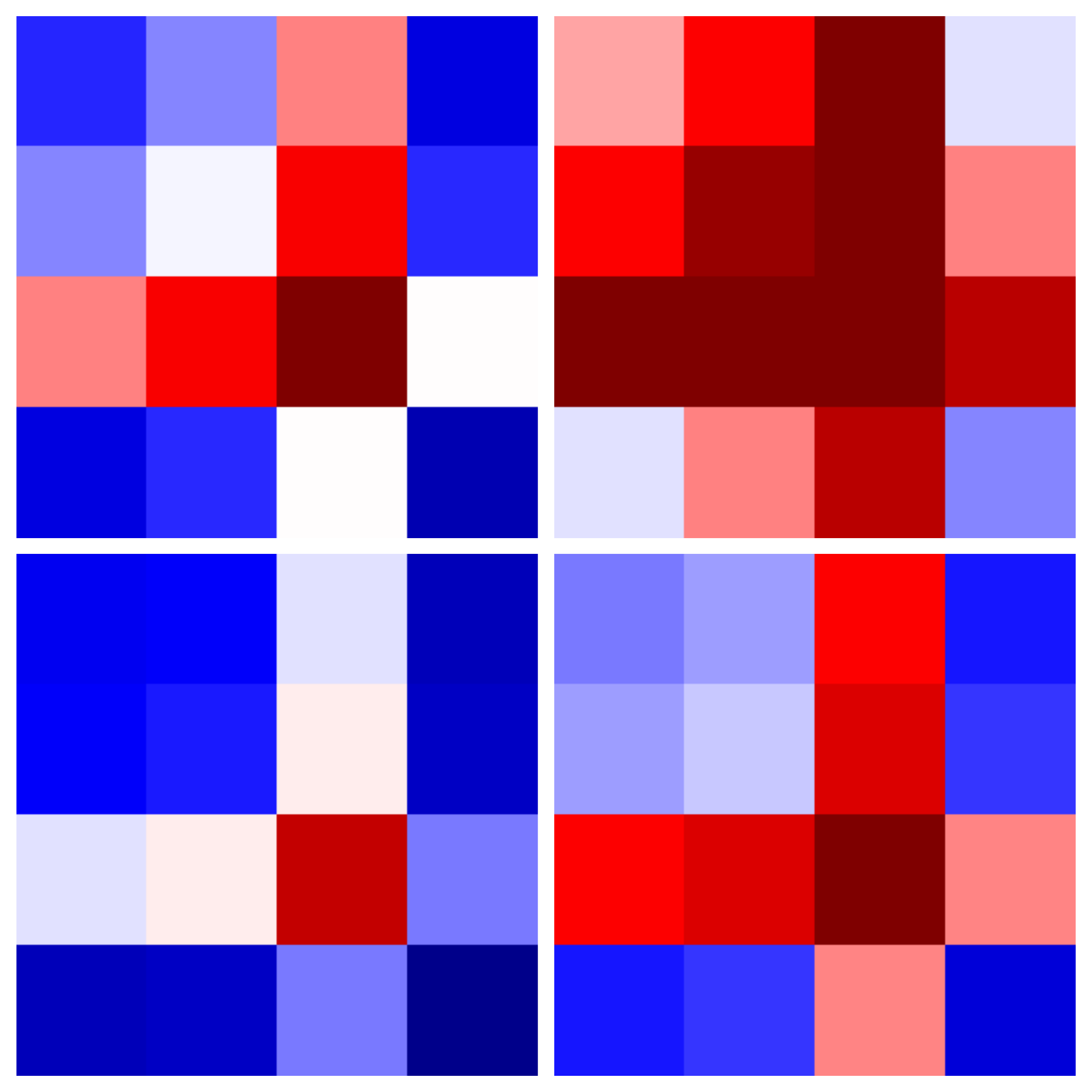}
    \includegraphics[width=0.24\textwidth]{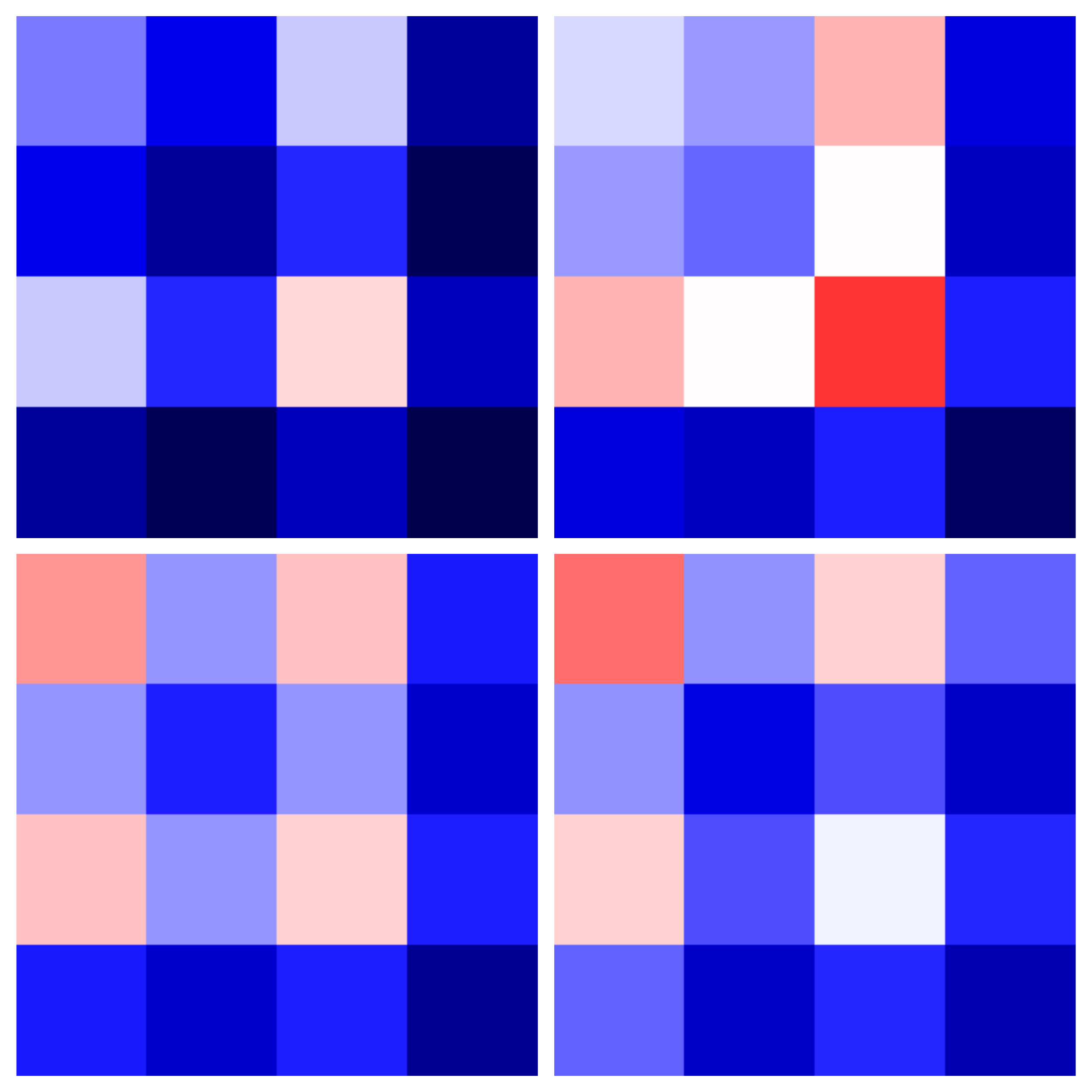}
    \includegraphics[width=0.24\textwidth]{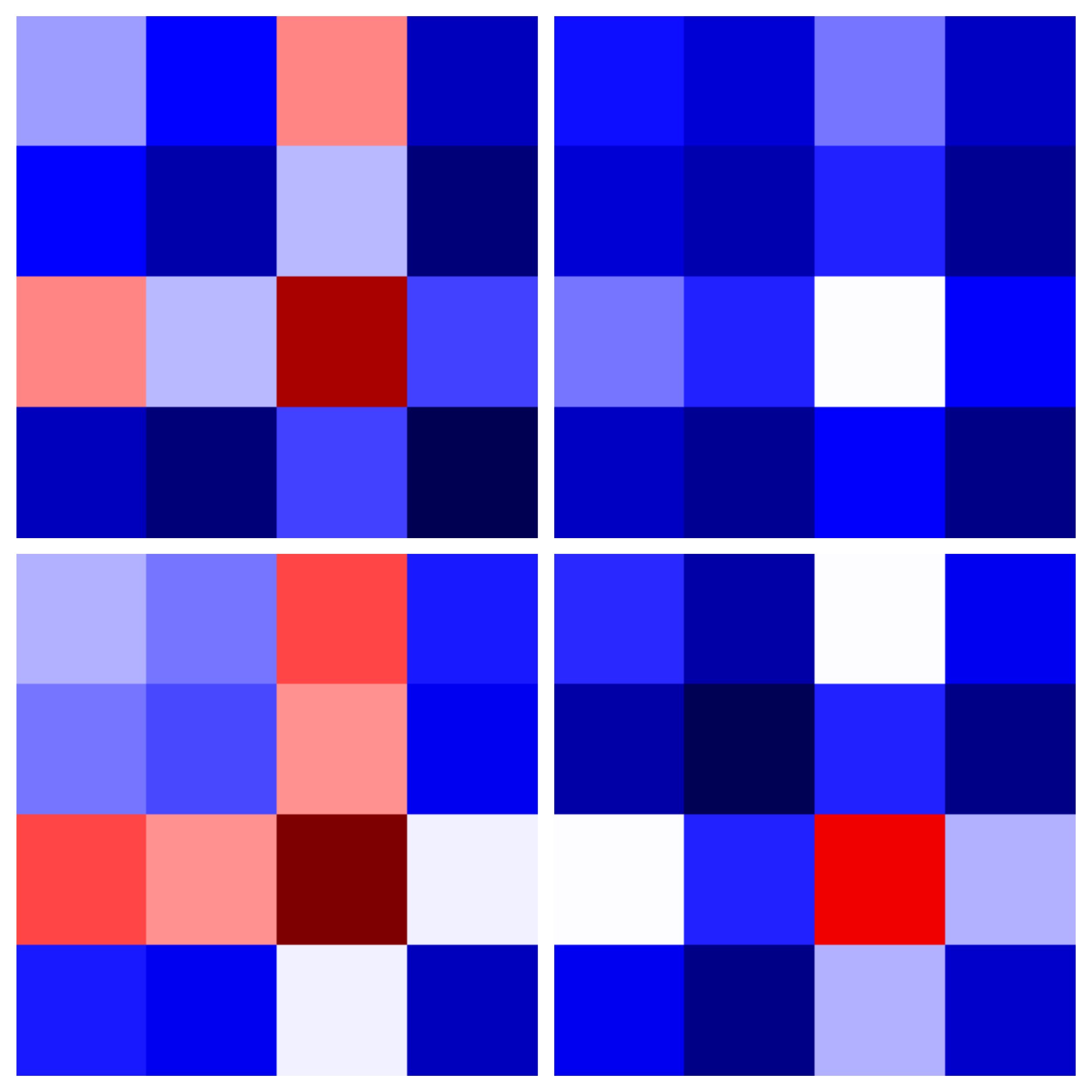}
    \caption{LM+GP applied to the covariances of 4 currencies (Euro, Swiss Franc, UK Pound and US Dollar). \textbf{Top:} the values of the currencies; \textbf{Middle:} sample covariances from the posterior GP; \textbf{Bottom:} selected covariance matrices from the Posterior - the grey lines (top) show the date of the samples.}
    \label{fig:currencies_GP}
    \vspace{-0.75em}
\end{figure}

In this experiment, we estimate the covariance of currency exchange rates over time. The conjugate prior to covariance matrices is the inverse Wishart distribution but it is not straightforward to model temporal variations within a Wishart framework (more below). 
LM+GP allows transforming the cavity distribution at every datapoint $XX^\top$ of the inverse Wishart to a Gaussian, applying LM, and creating a GP similar to the previous sections. We use
\begin{subequations}
\begin{align}
    \mu &= \operatorname{sqrtm}\left(\frac{1}{(\nu + p)}\Psi^{-1}\right) \\
    \Sigma &= \frac{1}{(\nu + p)^2} \left(I_p \otimes \Psi\right)\left(\Psi^{\frac{1}{2}} \otimes \Psi^{\frac{1}{2}} + I_p\right)^{-1}
\end{align}
\end{subequations}
to compute the mean and covariance function for the inverse Wishart.
This time we have $T=235$ months between 1998 and 2018 with $C=4$ currencies (Euro, Swiss Franc, UK Pound, US Dollar). 

Custom-designing a kernel allows us to include information about real-world events (e.g.~a financial crisis) or other statistical correlations (e.g.~the correlation between GDP and currency value) into our model.
To show how we could use the additional uncertainty from LM+GP we plot the covariance of different sample functions drawn from the GP posterior and different samples of covariance matrices drawn at four different points in time in Figure \ref{fig:currencies_GP}. We find that the quality of the resulting posterior is high, especially given the simplicity of LM+GP.

We also compare LM(inv.~Wishart)+GP with the results of a vanilla GP fit on the covariances in Table \ref{tab:experiments_currencies}. We find that the quality of the approximation is comparable on the currency dataset but the LM+GP variant has the advantage of actually reflecting covariance matrices, i.e.~p.s.d matrices that correspond to the covariances of an underlying process rather than matrices that just roughly track values that happen to be covariances. 
\begin{table}[ht!]
    \centering
    \caption{Quantitative results on currency data. We find that both methods have comparable accuracy but LM+GP is slightly better calibrated (0.95 is optimal for in2std). 
    }
    \begin{tabular}{l | rr |rr}
    \toprule
    & \multicolumn{2}{c}{\textbf{vanilla GP}} & \multicolumn{2}{c}{\textbf{LM(inv.~Wishart)+GP}} \\
    \textbf{Dataset} & \textbf{RMSE} $\downarrow$ & \textbf{in2std} & \textbf{RMSE} $\downarrow$ & \textbf{in2std} \\
    \midrule
    Currencies & 0.074 & 1.0 & 0.123 & 0.973\\
    \bottomrule
    \end{tabular}
    \label{tab:experiments_currencies}
\end{table}

In general, modeling time-dependent covariance matrices is non-trivial. \citep{wilson2010generalised} developed the Generalized Wishart Process (GWP) which uses a product out Gaussian processes in combination with MCMC to model time-dependent covariance matrices. However, we observed empirically that the associated MCMC method mixes very slowly, to the point where it did not yield usable results for us, even after 24h of wall-clock time on the currency dataset. In contrast, LM(inv.~Wishart)+GP trained in a few seconds and provided good results. Table \ref{tab:timing_invwishart} provides a comparison of the timing and complexity.

\begin{table}[ht!]
    \centering
    \begin{tabular}{lcc || c}
        \toprule
        & Inference & $\mu_X, \Sigma_X$ (LM) & GWP    \\
        \midrule
        Complexity & $\mathcal{O}(D^2 n_x n_X^2 + D^2 n_X^3)$ & $\mathcal{O}(n_X {D^{2\cdot 2}})$/$\mathcal{O}(n_X D^2)$ & $\mathcal{O}(n_s(D n_X^3 + \nu D^2))$ \\
        Time in $s$& 1.59 & 0.09 & 82530  \\
        \bottomrule
    \end{tabular}
    \caption{Timing and Complexity for Bayesian Inference on p.s.d.~matrices: The LM+GP step is split into an inference and a transformation step. The complexity of the transformation step depends on the basis transformation. For our experiment we chose the sqrtm-transformation as it yielded better results. Since $D << n_X$ LM+GP is much faster than GWP. For both methods $n_X=235$ and $D=4$. For GWP we draw $n_s=5000$ samples of which we only used 100 since they were very correlated.}
    \label{tab:timing_invwishart}
\end{table}
\subsection{Timing}
\label{subsec:timing}
\begin{table}[htb!]
    \centering
    \begin{tabular}{l  r  r  r  r  r}
         \toprule
         Dataset & Outputs & Training instances & Dimensionality & Time in $s$ & Complexity \\
         \midrule 
         EEG & 2 & 10980 & 14 & 0.0024 & $\mathcal{O}(N)$\\
         HTRU2 & 2 & 12898 & 8 & 0.0067 & $\mathcal{O}(N)$\\
         MAGIC & 2 & 14020 & 10 & 0.0021 & $\mathcal{O}(N)$\\
         MINIBOO & 2 & 120064 & 50 & 0.0037 & $\mathcal{O}(N)$\\
         \midrule 
         CREDIDCARD & 1 & 833 & 12 & 0.0018 & $\mathcal{O}(N)$\\
         DOCTOR & 1 & 3477 & 11 & 0.0017 & $\mathcal{O}(N)$\\
         CITATIONS & 1 & 434 & 12 & 0.0005 & $\mathcal{O}(N)$\\
         GSS7402 & 1 & 2219 & 9 & 0.0004 & $\mathcal{O}(N)$\\
         MEDICAID & 1 & 667 & 13 & 0.0004 & $\mathcal{O}(N)$\\
         \midrule 
         LETTER & 26 & 15000 & 16 & 0.0149 & $\mathcal{O}(N O)$\\
         DRIVE & 11 & 48509 & 48 & 0.0077 & $\mathcal{O}(N O)$\\
         MOCAP & 5 & 68095 & 37 & 0.0082 & $\mathcal{O}(N O)$\\
         \midrule 
         CURRENCY & 16 & 235 & 4 & 0.0653 & $\mathcal{O}(N {O^{2}})$ \\
         \bottomrule
    \end{tabular}
    \caption{\textbf{Timings and complexity for LM in all experiments:} In all cases, LM costs less than 0.1 seconds to apply and scales linearly in the number of datapoints.}
    \label{tab:timing_overview}
\end{table}
Table \ref{tab:timing_overview} provides time overhead for the application of LM across the datasets and settings presented in the experiments. We note that this number never exceeded 0.1s, corroborating that LM can indeed be thought of as a fast pre-processing step for standard Gaussian inference.
\subsection{Distance Measures}
\label{subsec:distance_measures}
\begin{figure}[!tb]
    \setlength{\figwidth}{\textwidth}
    \setlength{\figheight}{0.2\textheight}
    \centering
    
    \includegraphics[width=\textwidth]{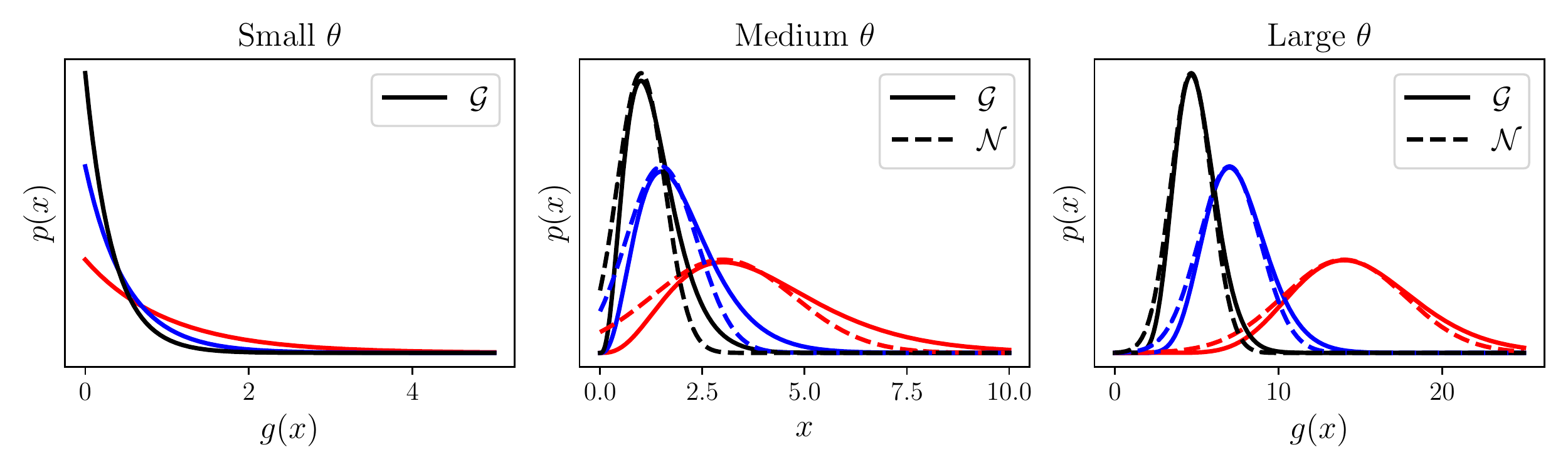}
    \caption{Visualization of small ($\alpha=1,\lambda=1$), medium ($\alpha=4,\lambda=2$) and large ($\alpha=15, \lambda=3$) $\theta$ for the Gamma distribution.}
    \label{fig:sml_Gamma}
\end{figure}

Figure \ref{fig:LA_different_basis} suggests that the distributions are more Gaussian in other bases. Here we attempt to quantify this similarity. To this end, we compare every approximation to the true distribution in every base for ten sets of parameters $\theta$ using two metrics. The ten sets of parameters increase in a linear fashion and represent a cut through the entire space of possible parameters. They are always chosen such that the first two sets of parameters yield no valid Laplace approximation in the standard base but do yield one in the transformed basis (e.g.~$\alpha < 1$ for the Gamma distribution) to emphasize this feature of LM.
For large parameters, many exponential family distributions are approximately Gaussian (e.g. Gamma with $\alpha > 10$). The parameters are always increased in such a way that the last set reflects this fact. To showcase these scenarios consider Figure \ref{fig:sml_Gamma} where we show a Gamma distribution with $\alpha=1,4,15$ and $\lambda=1,2,3$ for a small, medium and large scenario respectively.
The first metric to compare the distributions is  KL-divergence and the second is \textit{maximum mean discrepancy} (MMD) \citep{MMDScholkopf2012}. We calculate the KL-divergence by drawing samples from the distribution we want to approximate, i.e.~$x \sim p(x)$, and then compute $\frac{1}{N}\sum_{i=1}^{N} \log\left(\frac{p(x)}{q(x)}\right)$ where $q(x)$ is the Gaussian given by the Laplace approximation. A table of the exact parameters chosen can be found in Appendix \ref{sec:appendix_C}. The results can be found in Table \ref{table:distances}.

\begin{table}[tb]
    \setlength{\figwidth}{0.5\textwidth}
    \setlength{\figheight}{0.12\textheight}
    \centering
    
    \sbox0{\hwplotStandard}\sbox1{\hwplotLog}\sbox2{\hwplotSqrt}%
    \resizebox{\textwidth}{!}{
    \begin{tabular}{l cc}
        \toprule
        \textbf{Distribution} &  KL-Divergence $\downarrow$ & MMD $\downarrow$ \\
        \midrule
        \multirow[c]{1}{*}[1.5em]{Exponential}  & 
        \tiny 
\begin{tikzpicture}

\definecolor{color0}{rgb}{0.698,0.133,0.133}

\begin{axis}[
height=\figheight,
tick align=outside,
tick pos=left,
width=\figwidth,
x grid style={white!69!black},
xmin=-0.45, xmax=9.45,
xtick style={color=black},
xtick={-2,0,2,4,6,8,10},
xticklabels={
  \(\displaystyle -2\),
  \(\displaystyle 0\),
  \(\displaystyle 2\),
  \(\displaystyle 4\),
  \(\displaystyle 6\),
  \(\displaystyle 8\),
  \(\displaystyle 10\)
},
y grid style={white!69!black},
ymin=0, ymax=0.383,
ytick style={color=black},
ytick={0,0.2,0.4},
yticklabels={\(\displaystyle 0.0\),\(\displaystyle 0.2\),\(\displaystyle 0.4\)}
]
\addplot [semithick, color0]
table {%
0 0.332
1 0.327
2 0.325
3 0.321
4 0.331
5 0.328
6 0.327
7 0.333
8 0.326
9 0.329
};
\addplot [semithick, blue]
table {%
0 0.123
1 0.122
2 0.125
3 0.125
4 0.123
5 0.122
6 0.123
7 0.125
8 0.123
9 0.125
};
\end{axis}

\end{tikzpicture} & \tiny
\begin{tikzpicture}

\definecolor{color0}{rgb}{0.698,0.133,0.133}

\begin{axis}[
height=\figheight,
tick align=outside,
tick pos=left,
width=\figwidth,
x grid style={white!69!black},
xmin=-0.45, xmax=9.45,
xtick style={color=black},
xtick={-2,0,2,4,6,8,10},
xticklabels={
  \(\displaystyle -2\),
  \(\displaystyle 0\),
  \(\displaystyle 2\),
  \(\displaystyle 4\),
  \(\displaystyle 6\),
  \(\displaystyle 8\),
  \(\displaystyle 10\)
},
y grid style={white!69!black},
ymin=0, ymax=0.0287,
ytick style={color=black},
ytick={0,0.02,0.04},
yticklabels={\(\displaystyle 0.00\),\(\displaystyle 0.02\),\(\displaystyle 0.04\)}
]
\addplot [semithick, color0]
table {%
0 0.026
1 0.0287
2 0.0267
3 0.0257
4 0.0254
5 0.0237
6 0.026
7 0.0264
8 0.027
9 0.0222
};
\addplot [semithick, blue]
table {%
0 0.0148
1 0.0103
2 0.00764
3 0.00593
4 0.00497
5 0.00487
6 0.00352
7 0.00357
8 0.00299
9 0.00318
};
\end{axis}

\end{tikzpicture} \\
        \multirow[c]{1}{*}[1.5em]{Gamma}        & 
        \tiny 
\begin{tikzpicture}

\definecolor{color0}{rgb}{0.698,0.133,0.133}

\begin{axis}[
height=\figheight,
tick align=outside,
tick pos=left,
width=\figwidth,
x grid style={white!69!black},
xmin=0.05, xmax=9.95,
xtick style={color=black},
xtick={0,2,4,6,8,10},
xticklabels={
  \(\displaystyle 0\),
  \(\displaystyle 2\),
  \(\displaystyle 4\),
  \(\displaystyle 6\),
  \(\displaystyle 8\),
  \(\displaystyle 10\)
},
y grid style={white!69!black},
ymin=0, ymax=1.72,
ytick style={color=black},
ytick={0,1,2},
yticklabels={\(\displaystyle 0\),\(\displaystyle 1\),\(\displaystyle 2\)}
]
\addplot [semithick, black]
table {%
0.5 nan
1.5 1.72
2.5 0.553
3.5 0.336
4.5 0.236
5.5 0.188
6.5 0.149
7.5 0.127
8.5 0.112
9.5 0.099
};
\addplot [semithick, color0]
table {%
0.5 0.846
1.5 0.202
2.5 0.102
3.5 0.0683
4.5 0.0523
5.5 0.0423
6.5 0.0335
7.5 0.0297
8.5 0.027
9.5 0.0228
};
\addplot [semithick, blue]
table {%
0.5 0.844
1.5 0.062
2.5 0.0308
3.5 0.018
4.5 0.0138
5.5 0.0114
6.5 0.0088
7.5 0.00794
8.5 0.00696
9.5 0.00543
};
\end{axis}

\end{tikzpicture} & \tiny
\begin{tikzpicture}

\definecolor{color0}{rgb}{0.698,0.133,0.133}

\begin{axis}[
height=\figheight,
tick align=outside,
tick pos=left,
width=\figwidth,
x grid style={white!69!black},
xmin=0.05, xmax=9.95,
xtick style={color=black},
xtick={0,2,4,6,8,10},
xticklabels={
  \(\displaystyle 0\),
  \(\displaystyle 2\),
  \(\displaystyle 4\),
  \(\displaystyle 6\),
  \(\displaystyle 8\),
  \(\displaystyle 10\)
},
y grid style={white!69!black},
ymin=0, ymax=0.213,
ytick style={color=black},
ytick={0,0.2,0.4},
yticklabels={\(\displaystyle 0.0\),\(\displaystyle 0.2\),\(\displaystyle 0.4\)}
]
\addplot [semithick, black]
table {%
0.5 nan
1.5 0.138
2.5 0.0603
3.5 0.0377
4.5 0.029
5.5 0.0205
6.5 0.0181
7.5 0.0153
8.5 0.0136
9.5 0.0124
};
\addplot [semithick, color0]
table {%
0.5 0.04
1.5 0.0201
2.5 0.0105
3.5 0.00833
4.5 0.00492
5.5 0.00435
6.5 0.00312
7.5 0.00268
8.5 0.002
9.5 0.00171
};
\addplot [semithick, blue]
table {%
0.5 0.213
1.5 0.00719
2.5 0.00324
3.5 0.00162
4.5 0.00145
5.5 0.000694
6.5 0.000604
7.5 0.000339
8.5 0.000342
9.5 0.000218
};
\end{axis}

\end{tikzpicture} \\
        \multirow[c]{1}{*}[1.5em]{Inv. Gamma} & 
        \tiny 
\begin{tikzpicture}

\definecolor{color0}{rgb}{0.698,0.133,0.133}

\begin{axis}[
height=\figheight,
tick align=outside,
tick pos=left,
width=\figwidth,
x grid style={white!69!black},
xmin=-0.45, xmax=9.45,
xtick style={color=black},
xtick={-2,0,2,4,6,8,10},
xticklabels={
  \(\displaystyle -2\),
  \(\displaystyle 0\),
  \(\displaystyle 2\),
  \(\displaystyle 4\),
  \(\displaystyle 6\),
  \(\displaystyle 8\),
  \(\displaystyle 10\)
},
y grid style={white!69!black},
ymin=0, ymax=5.77,
ytick style={color=black},
ytick={0,5,10},
yticklabels={\(\displaystyle 0\),\(\displaystyle 5\),\(\displaystyle 10\)}
]
\addplot [semithick, black]
table {%
0 5.77
1 3.34
2 2.2
3 1.61
4 1.22
5 0.987
6 0.811
7 0.674
8 0.593
9 0.513
};
\addplot [semithick, color0]
table {%
0 0.323
1 0.139
2 0.0851
3 0.0601
4 0.0462
5 0.0378
6 0.0317
7 0.028
8 0.0251
9 0.0216
};
\addplot [semithick, blue]
table {%
0 2.79
1 1.27
2 0.77
3 0.538
4 0.395
5 0.307
6 0.259
7 0.223
8 0.185
9 0.17
};
\end{axis}

\end{tikzpicture} & \tiny
\begin{tikzpicture}

\definecolor{color0}{rgb}{0.698,0.133,0.133}

\begin{axis}[
height=\figheight,
tick align=outside,
tick pos=left,
width=\figwidth,
x grid style={white!69!black},
xmin=-0.45, xmax=9.45,
xtick style={color=black},
xtick={-2,0,2,4,6,8,10},
xticklabels={
  \(\displaystyle -2\),
  \(\displaystyle 0\),
  \(\displaystyle 2\),
  \(\displaystyle 4\),
  \(\displaystyle 6\),
  \(\displaystyle 8\),
  \(\displaystyle 10\)
},
y grid style={white!69!black},
ymin=0, ymax=0.186,
ytick style={color=black},
ytick={0,0.1,0.2},
yticklabels={\(\displaystyle 0.0\),\(\displaystyle 0.1\),\(\displaystyle 0.2\)}
]
\addplot [semithick, black]
table {%
0 0.186
1 0.1
2 0.0603
3 0.0411
4 0.0299
5 0.0207
6 0.0162
7 0.0129
8 0.0104
9 0.00871
};
\addplot [semithick, color0]
table {%
0 0.027
1 0.0148
2 0.00837
3 0.00677
4 0.00512
5 0.00326
6 0.00284
7 0.00233
8 0.00191
9 0.00168
};
\addplot [semithick, blue]
table {%
0 0.0952
1 0.0363
2 0.0187
3 0.0111
4 0.00778
5 0.00483
6 0.00373
7 0.00297
8 0.00229
9 0.00185
};
\end{axis}

\end{tikzpicture} \\
        \multirow[c]{1}{*}[1.5em]{Chi-squared}        & 
        \tiny 
\begin{tikzpicture}

\definecolor{color0}{rgb}{0.698,0.133,0.133}

\begin{axis}[
height=\figheight,
tick align=outside,
tick pos=left,
width=\figwidth,
x grid style={white!69!black},
xmin=0.55, xmax=10.4,
xtick style={color=black},
xtick={0,2,4,6,8,10,12},
xticklabels={
  \(\displaystyle 0\),
  \(\displaystyle 2\),
  \(\displaystyle 4\),
  \(\displaystyle 6\),
  \(\displaystyle 8\),
  \(\displaystyle 10\),
  \(\displaystyle 12\)
},
y grid style={white!69!black},
ymin=0, ymax=1.66,
ytick style={color=black},
ytick={0,1,2},
yticklabels={\(\displaystyle 0\),\(\displaystyle 1\),\(\displaystyle 2\)}
]
\addplot [semithick, black]
table {%
1 nan
2 nan
3 1.66
4 0.807
5 0.543
6 0.4
7 0.301
8 0.27
9 0.231
10 0.201
};
\addplot [semithick, color0]
table {%
1 0.85
2 0.336
3 0.204
4 0.138
5 0.103
6 0.0833
7 0.07
8 0.0598
9 0.0535
10 0.0476
};
\addplot [semithick, blue]
table {%
1 0.847
2 0.12
3 0.063
4 0.0406
5 0.0299
6 0.0234
7 0.0193
8 0.0164
9 0.0139
10 0.0123
};
\end{axis}

\end{tikzpicture} & \tiny
\begin{tikzpicture}

\definecolor{color0}{rgb}{0.698,0.133,0.133}

\begin{axis}[
height=\figheight,
tick align=outside,
tick pos=left,
width=\figwidth,
x grid style={white!69!black},
xmin=0.55, xmax=10.4,
xtick style={color=black},
xtick={0,2,4,6,8,10,12},
xticklabels={
  \(\displaystyle 0\),
  \(\displaystyle 2\),
  \(\displaystyle 4\),
  \(\displaystyle 6\),
  \(\displaystyle 8\),
  \(\displaystyle 10\),
  \(\displaystyle 12\)
},
y grid style={white!69!black},
ymin=0, ymax=0.213,
ytick style={color=black},
ytick={0,0.2,0.4},
yticklabels={\(\displaystyle 0.0\),\(\displaystyle 0.2\),\(\displaystyle 0.4\)}
]
\addplot [semithick, black]
table {%
1 nan
2 nan
3 0.105
4 0.0521
5 0.0326
6 0.0213
7 0.0161
8 0.0134
9 0.0106
10 0.00874
};
\addplot [semithick, color0]
table {%
1 0.04
2 0.025
3 0.0195
4 0.0141
5 0.00952
6 0.00861
7 0.00651
8 0.00688
9 0.00442
10 0.00502
};
\addplot [semithick, blue]
table {%
1 0.213
2 0.0187
3 0.00865
4 0.00522
5 0.00366
6 0.00259
7 0.00308
8 0.00155
9 0.00203
10 0.0019
};
\end{axis}

\end{tikzpicture} \\
        \multirow[c]{1}{*}[1.5em]{Beta}        & 
        \tiny
\begin{tikzpicture}

\definecolor{color0}{rgb}{0.698,0.133,0.133}

\begin{axis}[
height=\figheight,
tick align=outside,
tick pos=left,
width=\figwidth,
x grid style={white!69!black},
xmin=-0.45, xmax=9.45,
xtick style={color=black},
xtick={-2,0,2,4,6,8,10},
xticklabels={
  \(\displaystyle -2\),
  \(\displaystyle 0\),
  \(\displaystyle 2\),
  \(\displaystyle 4\),
  \(\displaystyle 6\),
  \(\displaystyle 8\),
  \(\displaystyle 10\)
},
y grid style={white!69!black},
ymin=0, ymax=0.822,
ytick style={color=black},
ytick={0,0.5,1},
yticklabels={\(\displaystyle 0.0\),\(\displaystyle 0.5\),\(\displaystyle 1.0\)}
]
\addplot [semithick, black]
table {%
0 nan
1 0.822
2 0.397
3 0.256
4 0.187
5 0.145
6 0.119
7 0.101
8 0.0867
9 0.0747
};
\addplot [semithick, color0]
table {%
0 0.157
1 0.0724
2 0.0491
3 0.0382
4 0.0305
5 0.0256
6 0.0219
7 0.0187
8 0.0158
9 0.016
};
\end{axis}

\end{tikzpicture} & \tiny
\begin{tikzpicture}

\definecolor{color0}{rgb}{0.698,0.133,0.133}

\begin{axis}[
height=\figheight,
tick align=outside,
tick pos=left,
width=\figwidth,
x grid style={white!69!black},
xmin=-0.45, xmax=9.45,
xtick style={color=black},
xtick={-5,0,5,10},
xticklabels={
  \(\displaystyle -5\),
  \(\displaystyle 0\),
  \(\displaystyle 5\),
  \(\displaystyle 10\)
},
y grid style={white!69!black},
ymin=0, ymax=0.0954,
ytick style={color=black},
ytick={0,0.05,0.1},
yticklabels={\(\displaystyle 0.00\),\(\displaystyle 0.05\),\(\displaystyle 0.10\)}
]
\addplot [semithick, black]
table {%
0 nan
1 0.0954
2 0.0196
3 0.00995
4 0.00545
5 0.00391
6 0.0035
7 0.00254
8 0.00195
9 0.00172
};
\addplot [semithick, color0]
table {%
0 0.00889
1 0.00407
2 0.00319
3 0.00211
4 0.00219
5 0.00178
6 0.00168
7 0.00179
8 0.00109
9 0.00106
};
\end{axis}

\end{tikzpicture} \\
        \multirow[c]{1}{*}[1.5em]{Dirichlet}    & 
        \tiny
\begin{tikzpicture}

\definecolor{color0}{rgb}{0.698,0.133,0.133}

\begin{axis}[
height=\figheight,
tick align=outside,
tick pos=left,
width=\figwidth,
x grid style={white!69!black},
xmin=-0.45, xmax=9.45,
xtick style={color=black},
xtick={-2,0,2,4,6,8,10},
xticklabels={
  \(\displaystyle -2\),
  \(\displaystyle 0\),
  \(\displaystyle 2\),
  \(\displaystyle 4\),
  \(\displaystyle 6\),
  \(\displaystyle 8\),
  \(\displaystyle 10\)
},
y grid style={white!69!black},
ymin=0, ymax=0.000418,
ytick style={color=black},
ytick={0,0.00025,0.0005},
yticklabels={
  \(\displaystyle 0.00000\),
  \(\displaystyle 0.00025\),
  \(\displaystyle 0.00050\)
}
]
\addplot [semithick, black]
table {%
0 nan
1 0.000418
2 0.000224
3 0.000153
4 0.000112
5 8.85e-05
6 7.28e-05
7 6.12e-05
8 5.26e-05
9 4.59e-05
};
\addplot [semithick, color0]
table {%
0 0.000123
1 9.11e-05
2 5.47e-05
3 4.72e-05
4 3.7e-05
5 3.19e-05
6 2.67e-05
7 2.39e-05
8 1.86e-05
9 1.72e-05
};
\end{axis}

\end{tikzpicture} & \tiny
\begin{tikzpicture}

\definecolor{color0}{rgb}{0.698,0.133,0.133}

\begin{axis}[
height=\figheight,
tick align=outside,
tick pos=left,
width=\figwidth,
x grid style={white!69!black},
xmin=-0.45, xmax=9.45,
xtick style={color=black},
xtick={-2,0,2,4,6,8,10},
xticklabels={
  \(\displaystyle -2\),
  \(\displaystyle 0\),
  \(\displaystyle 2\),
  \(\displaystyle 4\),
  \(\displaystyle 6\),
  \(\displaystyle 8\),
  \(\displaystyle 10\)
},
y grid style={white!69!black},
ymin=0, ymax=0.062,
ytick style={color=black},
ytick={0,0.05,0.1},
yticklabels={\(\displaystyle 0.00\),\(\displaystyle 0.05\),\(\displaystyle 0.10\)}
]
\addplot [semithick, black]
table {%
0 nan
1 0.062
2 0.0144
3 0.00606
4 0.00359
5 0.00211
6 0.00153
7 0.00114
8 0.000785
9 0.000728
};
\addplot [semithick, color0]
table {%
0 0.015
1 0.00744
2 0.00388
3 0.00293
4 0.00177
5 0.00145
6 0.000572
7 0.000765
8 0.000636
9 0.000645
};
\end{axis}

\end{tikzpicture} \\
        \multirow[c]{1}{*}[1.5em]{Wishart}        & 
        \tiny
\begin{tikzpicture}

\definecolor{color0}{rgb}{0.698,0.133,0.133}

\begin{axis}[
height=\figheight,
tick align=outside,
tick pos=left,
width=\figwidth,
x grid style={white!69!black},
xmin=-0.45, xmax=9.45,
xtick style={color=black},
xtick={-2,0,2,4,6,8,10},
xticklabels={
  \(\displaystyle -2\),
  \(\displaystyle 0\),
  \(\displaystyle 2\),
  \(\displaystyle 4\),
  \(\displaystyle 6\),
  \(\displaystyle 8\),
  \(\displaystyle 10\)
},
y grid style={white!69!black},
ymin=0, ymax=17.3,
ytick style={color=black},
ytick={0,10,20},
yticklabels={\(\displaystyle 0\),\(\displaystyle 10\),\(\displaystyle 20\)}
]
\addplot [semithick, black]
table {%
0 nan
1 nan
2 17.3
3 9.17
4 6.92
5 5.88
6 5.29
7 4.99
8 4.85
9 4.78
};
\addplot [semithick, color0]
table {%
0 3.28
1 2.18
2 1.49
3 1.06
4 0.816
5 0.796
6 0.783
7 0.935
8 1.15
9 1.44
};
\addplot [semithick, blue]
table {%
0 2.31
1 1.42
2 0.939
3 0.71
4 0.641
5 0.695
6 0.855
7 1.09
8 1.4
9 1.78
};
\end{axis}

\end{tikzpicture} & \tiny
\begin{tikzpicture}

\definecolor{color0}{rgb}{0.698,0.133,0.133}

\begin{axis}[
height=\figheight,
tick align=outside,
tick pos=left,
width=\figwidth,
x grid style={white!69!black},
xmin=-0.45, xmax=9.45,
xtick style={color=black},
xtick={-2,0,2,4,6,8,10},
xticklabels={
  \(\displaystyle -2\),
  \(\displaystyle 0\),
  \(\displaystyle 2\),
  \(\displaystyle 4\),
  \(\displaystyle 6\),
  \(\displaystyle 8\),
  \(\displaystyle 10\)
},
y grid style={white!69!black},
ymin=0, ymax=0.159,
ytick style={color=black},
ytick={0,0.1,0.2},
yticklabels={\(\displaystyle 0.0\),\(\displaystyle 0.1\),\(\displaystyle 0.2\)}
]
\addplot [semithick, black]
table {%
0 nan
1 nan
2 0.159
3 0.0404
4 0.014
5 0.00748
6 0.00426
7 0.00273
8 0.00175
9 0.0011
};
\addplot [semithick, color0]
table {%
0 0.0308
1 0.0326
2 0.0372
3 0.0303
4 0.0281
5 0.0275
6 0.0256
7 0.0252
8 0.0214
9 0.0211
};
\addplot [semithick, blue]
table {%
0 0.127
1 0.0911
2 0.0813
3 0.0719
4 0.0599
5 0.0568
6 0.0504
7 0.0459
8 0.0461
9 0.0369
};
\end{axis}

\end{tikzpicture} \\
        \multirow[c]{1}{*}[1.5em]{Inv. Wishart} & 
        \tiny
\begin{tikzpicture}

\definecolor{color0}{rgb}{0.698,0.133,0.133}

\begin{axis}[
height=\figheight,
tick align=outside,
tick pos=left,
width=\figwidth,
x grid style={white!69!black},
xmin=-0.45, xmax=9.45,
xtick style={color=black},
xtick={-2,0,2,4,6,8,10},
xticklabels={
  \(\displaystyle -2\),
  \(\displaystyle 0\),
  \(\displaystyle 2\),
  \(\displaystyle 4\),
  \(\displaystyle 6\),
  \(\displaystyle 8\),
  \(\displaystyle 10\)
},
y grid style={white!69!black},
ymin=0, ymax=10.4,
ytick style={color=black},
ytick={0,10,20},
yticklabels={\(\displaystyle 0\),\(\displaystyle 10\),\(\displaystyle 20\)}
]
\addplot [semithick, black]
table {%
0 nan
1 nan
2 nan
3 10.4
4 9.92
5 9.68
6 9.24
7 9.13
8 8.78
9 8.81
};
\addplot [semithick, color0]
table {%
0 5.75
1 4.83
2 4.22
3 3.91
4 3.73
5 3.72
6 3.91
7 4.1
8 4.39
9 4.78
};
\addplot [semithick, blue]
table {%
0 10.4
1 9.75
2 8.88
3 8.06
4 7.58
5 7.13
6 6.87
7 6.78
8 6.73
9 6.8
};
\end{axis}

\end{tikzpicture} & \tiny
\begin{tikzpicture}

\definecolor{color0}{rgb}{0.698,0.133,0.133}

\begin{axis}[
height=\figheight,
tick align=outside,
tick pos=left,
width=\figwidth,
x grid style={white!69!black},
xmin=-0.45, xmax=9.45,
xtick style={color=black},
xtick={-2,0,2,4,6,8,10},
xticklabels={
  \(\displaystyle -2\),
  \(\displaystyle 0\),
  \(\displaystyle 2\),
  \(\displaystyle 4\),
  \(\displaystyle 6\),
  \(\displaystyle 8\),
  \(\displaystyle 10\)
},
y grid style={white!69!black},
ymin=0, ymax=0.414,
ytick style={color=black},
ytick={0,0.25,0.5},
yticklabels={\(\displaystyle 0.00\),\(\displaystyle 0.25\),\(\displaystyle 0.50\)}
]
\addplot [semithick, black]
table {%
0 nan
1 nan
2 nan
3 0.353
4 0.313
5 0.287
6 0.261
7 0.235
8 0.203
9 0.177
};
\addplot [semithick, color0]
table {%
0 0.115
1 0.105
2 0.102
3 0.0909
4 0.0915
5 0.0878
6 0.0852
7 0.0815
8 0.0791
9 0.0686
};
\addplot [semithick, blue]
table {%
0 0.414
1 0.307
2 0.241
3 0.185
4 0.157
5 0.138
6 0.125
7 0.104
8 0.0851
9 0.0742
};
\end{axis}

\end{tikzpicture} \\
        \bottomrule
    \end{tabular}
    } 
    \caption{Comparison of Distances. MMD and KL-divergence for increasing values of $\theta$. \usebox0 shows distances in the standard base, \usebox1 when transformed by the logarithm, logit or inverse-softmax, and \usebox2 by the square root. In general, the approximation in the standard base is worse than in the transformed base, and all approximations improve towards larger $\theta$s. See Appendix \ref{sec:appendix_C} for more detailed explanations and further discussion.
    }
    \label{table:distances}
\end{table}

We find that both the KL-divergence and MMD are minimized in a base provided by LM rather than the standard base. Furthermore, we can see that both metrics decrease for the standard base when the parameters increase indicating that exponential families become more Gaussian for large parameters. Lastly, no transformation is clearly superior to the other when compared across all exponential families. Thus we recommend choosing the transformation depending on the application.
%
%
%
\section{Conclusion}
\label{sec:conclusion}
We proposed and evaluated an approximate inference scheme that aims to be fast while achieving sufficiently good approximation quality. This is done by a change of basis for different exponential family distributions, applying a Laplace approximation, and deriving an explicit closed-form transformation between the parameters of the resulting Gaussian and the original distribution. This process allows the use of closed-form Gaussian (process) inference on non-Gaussian data without the use of any iterative optimization scheme, as long as their data type has an exponential family conjugate prior. We proposed two closed-form transformations for each of the most common exponential families, and demonstrated the simplicity and functionality of Laplace Matching by applying a GP to non-trivial data types, such as binary data, count data, categorical data and positive semi-definite matrices. 
In all cases, we show conceptual and empirical evidence for the approximation quality of the model.  
%


\acks{The authors gratefully acknowledge financial support by the European Research Council through ERC StG Action 757275 / PANAMA; the DFG Cluster of Excellence “Machine Learning - New Perspectives for Science”, EXC 2064/1, project number 390727645; the German Federal Ministry of Education and Research (BMBF) through the Tübingen AI Center (FKZ: 01IS18039A); and funds from the Ministry of Science, Research and Arts of the State of Baden-Württemberg. 
MH is grateful to the International Max Planck Research School for Intelligent Systems (IMPRS-IS) for support.
MH is thankful for all the feedback and discussions during the process of researching and writing the paper.}


\newpage

\appendix
\section{Math Background}
\label{sec:appendix_A}
\subsection{Notation}

Throughout the paper we use the following notation.
\begin{align*}
	x &: \text{variable of the probability distribution in standard basis} \\
	\pi &: \text{variable of the probability distribution in standard basis for the Dirichlet} \\
	y &: \text{variable of the probability distribution in transformed basis} \\
	g(x) &: \text{variable transformation} \\
	g^{-1}(y) = x(y) &: \text{inverse variable transformation} \\
	u &: \text{helper variable for non-trivial transformations} \\
	\phi(x)  &:	\text{sufficient statistics} \\	
	w &: \text{natural parameters} \\
	Z(w) &: \text{partition function/normalization constant} \\
	h(x) &: \text{base measure}
\end{align*}
\subsection{Example Transformation}

\label{subsec:chi2-normal}
It is well-known that a chi-square distribution with $k$ degrees of freedom describes the sum of the squares of $k$ independent, standard normal random variables. To introduce a certain `trick' we show the forward and backward transformation between chi-square and normal distribution when $k=1$. The trick deals with the problem that the square-root is not a bijective transformation and has to be split into multiple ranges. 

Let $X$ be normal with $\mu = 0, \sigma^2 = 1$. Let $Y = X^2$ and therefore $g(x) = x^2$, which is neither monotonic nor injective. Take $I_1 = (-\infty, 0)$ and $I_2 = [0, +\infty)$. Then $g$ is monotonic and injective on $I_1$ and $I_2$ and $I_1 \cup I_2 = \mathbb{R}$. $g(I_1) = (0, \infty)$ and $g(I_2) = [0, \infty)$. Then $g_1^{-1}: [0, \infty) \rightarrow \mathbb{R}$ by $g_1^{-1}(y) = -\sqrt{y}$ and $g_2^{-1}: [0, \infty) \rightarrow \mathbb{R}$ by $g_2^{-1}(y) = \sqrt{y}$. Let furthermore $\mathbf{1}_r(y)$ be one if $y$ is the range $r$ and 0 otherwise. The derivative is given by
$$\left\vert \frac{\partial g_i^{-1}(y)}{\partial y} \right\vert = \left\vert \frac{1}{2 \sqrt{y}} \right\vert = \frac{1}{2 \sqrt{y}}$$

Applying a change of variable (Equation \ref{eq:1D_variable_transform}) we transform a standard normal distribution to a chi-square distribution.
\begin{subequations}
\begin{align}
	f_Y(y) &= f_X(g_1^{-1}(y))	\left\vert\frac{\partial g_1^{-1}(y)}{\partial y} \right\vert \mathbf{1}_r(y) + f_X(g_2^{-1}(y))	\left\vert\frac{\partial g_2^{-1}(y)}{\partial y} \right\vert \mathbf{1}_r(y) \\
	&= \frac{1}{\sqrt{2\pi}} \exp(-\frac{y}{2}) \frac{1}{2\sqrt{y}} + \frac{1}{\sqrt{2\pi}} \exp(-\frac{y}{2}) \frac{1}{2\sqrt{y}} \qquad(y > 0)\\
	&= \frac{1}{\sqrt{2\pi}} \frac{1}{\sqrt{y}}\exp(-\frac{y}{2}) \\
	&= \frac{1}{2^{\frac{1}{2}}} \frac{1}{\Gamma(\frac{1}{2})} y^{-\frac{1}{2}} \exp(-\frac{1}{2}) \\
	&= \chi^2 (1)
\end{align}
\end{subequations}
The `trick' was to split up the variable transformation in two parts to adjust for the fact that the square-root is not bijective on $\mathbb{R}$. We can reverse the same procedure to transform a chi-square to a normal distribution. We keep the variable names from before. Let $X = \sqrt{Y}$ and therefore $h(x) = \sqrt{x}$. Then $h_1^{-1}: \mathbb{R} \rightarrow (-\infty, 0)$ by $h_1^{-1}(x) = -x^2$ and $h_2^{-1}: \mathbb{R} \rightarrow [0, \infty)$ by $h_2^{-1}(x) = x^2$. Then
\begin{equation}
\left\vert\frac{\partial h_i^{-1}(y)}{\partial y} \right\vert = \vert 2y \vert 
\end{equation}
and
\begin{subequations}
\begin{align}
	f_X(x) &= f_y(h_1^{-1}(x)) \left\vert\frac{\partial h_1^{-1}(y)}{\partial y} \right\vert \mathbf{1}_r(y) + f_y(h_2^{-1}(x)) \left\vert\frac{\partial h_2^{-1}(y)}{\partial y} \right\vert \mathbf{1}_r(y) \nonumber \\
	&= \frac{1}{\sqrt{2\pi}} \frac{1}{2\sqrt{x^2}} \exp(-\frac{x^2}{2}) |2x| \mathbf{1}_{(-\infty, 0)}(x) + \frac{1}{\sqrt{2\pi}} \frac{1}{2\sqrt{x^2}} \exp(-\frac{x^2}{2}) |2x| \mathbf{1}_{[0, \infty)}(x) \\
	&= \frac{1}{\sqrt{2\pi}} \exp(-\frac{x^2}{2})  \\
	&= \mathcal{N}(x; \mu=0, \sigma^2=1)
\end{align}
\end{subequations}
which is defined on the entirety of $\mathbb{R}$.

\subsubsection*{A generalization to matrices}

A positive definite matrix $A$ has $n$ distinct eigenvalues and $2^n$ possible square roots. $A$ has a decomposition $A=UDU^{-1}$ where $U$s columns are the eigenvectors of $A$ and $D$ is a diagonal matrix containing the eigenvalues $\lambda_i$ of $A$. Any square root of $A$ is given by $A^{\frac{1}{2}} = UD^{\frac{1}{2}}U^{-1}$. Since there are two possible choices for the square root of each eigenvalue $+\sqrt{\lambda_i}$ and $-\sqrt{\lambda_i}$ there are $2^n$ possible choices for the matrix $D^{\frac{1}{2}}$. 

If the matrix $A$ is symmetric and positive definite, its square root $A^{\frac{1}{2}}$ is also symmetric because of the decomposition $A=UDU^{-1}$.
The square root of $A$ that uses only the positive square roots of the eigenvalues is called the principle square root of $A$.

The Gamma distribution is the 1-dimensional special case of the Wishart distribution and therefore has $2^1=2$ possible options for $\lambda^{\frac{1}{2}}$, namely $-\lambda^{\frac{1}{2}}$ and $+\lambda^{\frac{1}{2}}$. Higher dimensional functions such as Wishart and inverse Wishart have $2^n$ different possibilities which have to be accounted for within the transformation.

\section{Derivations of the Transformations}
\label{sec:appendix_B}
\subsection{Exponential Distribution}
\label{subsec:exponential_dist}

\subsubsection{Standard Exponential Distribution}

The pdf of the exponential distribution is
\begin{subequations}
\begin{align}
	p(x| \lambda) &= \lambda \exp(-\lambda x)
	\label{eq:exponential_pdf} \\
	 &= \exp\left[-\lambda x + \log\lambda\right]
\end{align}
\end{subequations}
with exponential family values $h(x) = 1$, $\phi(x)=x$, $w = -\lambda$ and $Z(\lambda) = -\log\lambda$.

\subsubsection*{Laplace Approximation of the Exponential Distribution}
\begin{align*}
\text{log-pdf: } &\left( \log \lambda - \lambda x\right) \\
\text{1st derivative: }& - \lambda \\
\text{2nd derivative: }& 0
\end{align*}
The Laplace Approximation in the standard base is not defined since the second derivative is not positive. 

\subsubsection{Log-transformed Exponential Distribution}

We choose $X = \log(Y)$ with $g(x) = \log(x)$ and $x(y) = g^{-1}(y) = \exp(y)$. Also, $\left\vert \frac{\partial x(y)}{\partial y}\right\vert = \exp(y)$. It follows that the pdf in log-basis is
\begin{subequations}
\begin{align}
	\mathcal{E}_{Y_{\log}}(y; \lambda) &= \lambda \exp(-\lambda x(y)) \cdot \exp(y) \\ 
	&= \lambda \exp(-\lambda \exp(y) + y) \\
	&=  \exp\left[-\lambda \exp(y) + y + \log\lambda\right]
\end{align}
\end{subequations}
with exponential family values $h(y) =1$, $\phi(x) = (y, \exp(y))$, $w = (1, -\lambda)$ and $Z(\lambda)=\log\lambda$.

\subsubsection*{Laplace Approximation of the log-transformed Exponential Distribution}

\begin{align*}
\text{log-pdf: } & -\lambda \exp(y) + y + \log\lambda \\
\text{1st derivative: }& -\lambda\exp(y) + 1 \\
\text{mode: } & y = \log(1/\lambda) \\
\text{2nd derivative: }& -\lambda\exp(y)\\
\text{insert mode: }& -\lambda\exp(1/\lambda) = -1\\
\text{invert \& times -1: }&\sigma^2 = 1
\end{align*}
Therefore, the Laplace approximation in the transformed basis is given by $\mathcal{N}(y, \log(1/\lambda), 1)$.

\subsubsection*{The Bridge for the log-transformed Exponential Distribution}

\begin{figure}[!htb]
	\centering
	\includegraphics[width=\textwidth]{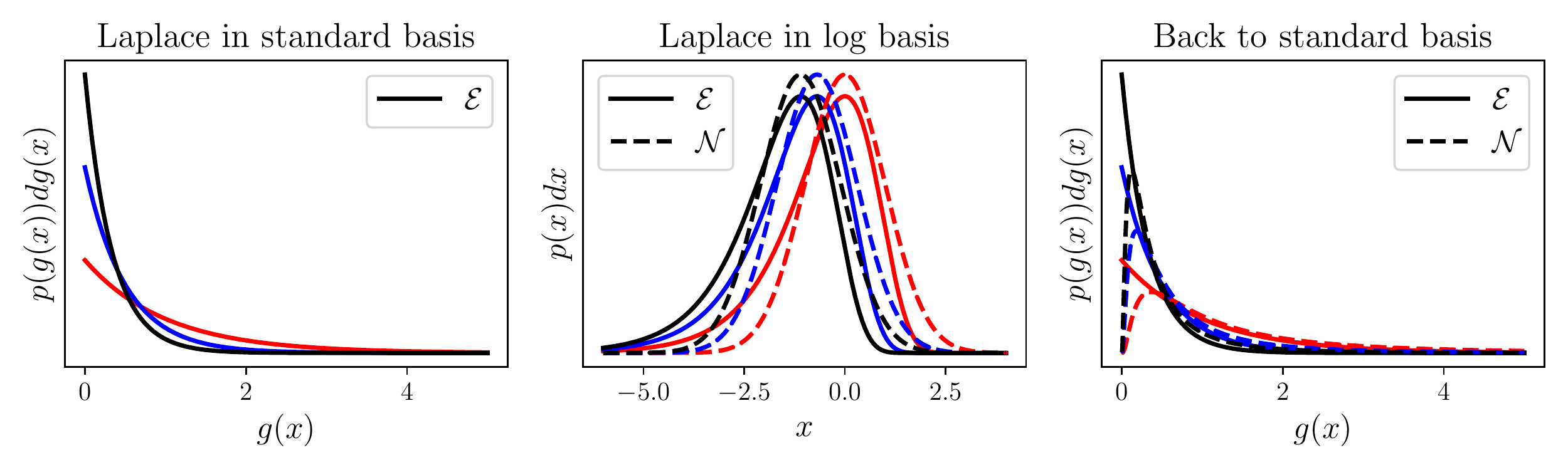}
	\caption{log-Bridge for the Exponential distribution}
	\label{fig:exponential_log_bridge}
\end{figure}

We have already found $\mu$ and $\sigma$. The inverse transformation is easily found through $\mu = \log(1/\lambda) \Leftrightarrow \lambda = 1/\exp(\mu)$. In summary:
\begin{subequations}
\begin{align}
	\mu &= \log(1/\lambda) \\
	\sigma^2 &= 1 \\
	\lambda &= 1/\exp(\mu)
\end{align}
\end{subequations}
A visual interpretation can be found in Figure \ref{fig:exponential_log_bridge}.

\subsubsection{Sqrt-transformed Exponential Distribution}

We transform the Exponential distribution with the sqrt-transformation, i.e. $Y = \sqrt{X}, g(x) = \sqrt{x},x_1(y) =  g_1^{-1}(y) = -y^2, x_2(y) = g_2^{-1}(y) = y^2$ and $\left\vert\frac{\partial x_i(y)}{\partial y} \right\vert = \left\vert\frac{\partial g_i^{-1}(y)}{\partial y} \right\vert = \vert 2y \vert$.
\begin{subequations}
\begin{align}
\mathcal{E}_{Y_\text{sqrt}}(y; \lambda) &= \lambda \exp(-\lambda y^2) \cdot 2y \\
&= 2 \cdot \exp\left[\log(y)-\lambda y^2 + \log\lambda\right]
\end{align}
\end{subequations}
with exponential family values $h(y)=2$, $\phi(y) = (\log(y), y^2))$, $w = (1, -\lambda)$ and $Z(\lambda)=\log\lambda$.

\subsubsection*{Laplace Approximation of the sqrt-transformed Exponential Distribution}
\begin{align*}
\text{log-pdf: } & \log(y)-\lambda y^2 + \log\lambda \\
\text{1st derivative: }& \frac{1}{y} - 2\lambda y \\
\text{mode: } & y = \sqrt{\frac{1}{2\lambda}}\\
\text{2nd derivative: }& -\frac{1}{y^2} - 2\lambda\\
\text{insert mode: }& -\frac{1}{\frac{1}{2\lambda}} - 2\lambda = -4\lambda\\
\text{invert \& times -1: }&\sigma^2 = \frac{1}{4\lambda}
\end{align*}
Therefore the resulting Gaussian is $\mathcal{N}\left(y; \mu=\sqrt{\frac{1}{2\lambda}}, \sigma^2=\frac{1}{4\lambda} \right)$

\subsubsection*{The Bridge for the sqrt-transformed Exponential Distribution}

\begin{figure}[!htb]
	\centering
	\includegraphics[width=\textwidth]{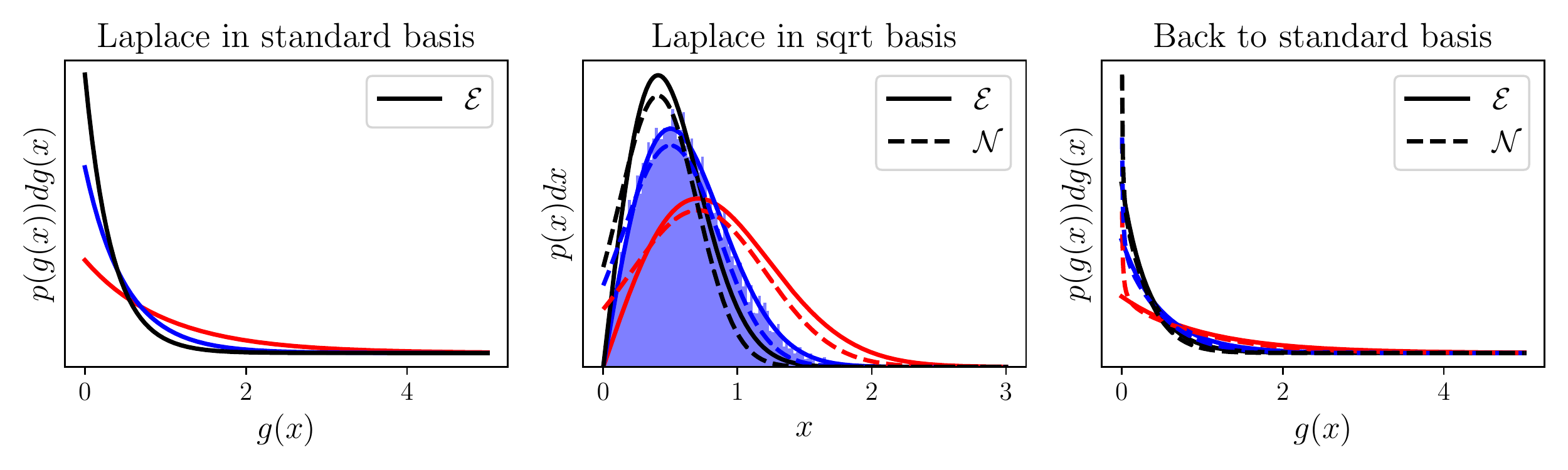}
	\caption{sqrt-bridge for the Exponential distribution. Transformed samples indicate that the closed-form transformation is correct.}
	\label{fig:exponential_sqrt_bridge}
\end{figure}

To get the inverse of the Bridge in the sqrt-base we can invert the mode $\mu = \sqrt{\frac{1}{2\lambda}} \Leftrightarrow \frac{1}{2\mu^2}$. In summary we have
\begin{subequations}
\begin{align}
	\mu &= \sqrt{\frac{1}{2\lambda}} \\
	\sigma^2 &= \frac{1}{4\lambda} \\
	\lambda &= \frac{1}{2\mu^2}
\end{align}
\end{subequations}
A visual interpretation can be found in Figure \ref{fig:exponential_sqrt_bridge}.
\subsection{Gamma Distribution}
\label{subsec:gamma_dist}

\subsubsection{Standard Gamma Distribution}

The pdf of the Gamma distribution in the standard base is
\begin{equation}
	\mathcal{G}_X(x, \alpha, \lambda) = \frac{\lambda^\alpha}{\Gamma(\alpha)} \cdot x^{(\alpha - 1)} \cdot e^{(-\lambda x)}
	\label{eq:gamma_pdf}
\end{equation}
where $\Gamma(\alpha)$ is the Gamma function. This can be written as
\begin{subequations}
\begin{align}
	\mathcal{G}_X(x, \alpha, \lambda) &= \exp \left[(\alpha -1)\log(x) - \lambda x + \alpha \log(\lambda) - \log(\Gamma(\alpha))\right] \\
	&= \frac{1}{x} \exp\left[\alpha\log(x) - \lambda x + \alpha \log(\lambda) - \log(\Gamma(\alpha))\right]
	\label{eq:gamma_exp_family}
\end{align}
\end{subequations}
with exponential family values $h(x) = \frac{1}{x},\, \phi(x)=(\log x, x),\, w=(\alpha, -\lambda)$ and $Z(\alpha, \lambda) = \log(\Gamma(\alpha)) - \alpha  \log(\lambda)$. 

\subsubsection*{Laplace Approximation of the Gamma Distribution}

\begin{align*}
	\text{log-pdf: } &\log\left( \frac{\lambda^\alpha}{\Gamma(\alpha)} \cdot x^{(\alpha - 1)} \cdot e^{(-\lambda x)}\right) \\
	&= \alpha \cdot \log(\lambda) - \log(\Gamma(\alpha)) + (\alpha -1)\log(x) -\lambda x\\
	\text{1st derivative: }& \frac{(\alpha-1)}{x} - \lambda \\
	\text{mode: }&  \frac{(\alpha-1)}{x} - \lambda = 0 \Leftrightarrow x=\frac{\alpha -1}{\lambda}\\
	\text{2nd derivative: }& -\frac{(\alpha-1)}{x^2}\\
	\text{insert mode: }& -\frac{(\alpha-1)}{(\frac{\alpha -1}{\lambda})^2} = -\frac{\lambda^2}{\alpha - 1} \\
	\text{invert and times -1: }&\sigma^2 = \frac{\alpha-1}{\lambda^2}
\end{align*}

The Laplace approximation of the Gamma distribution is therefore $\mathcal{N}\left(x; \frac{\alpha - 1}{\lambda}, \frac{\alpha-1}{\lambda^2}\right)$.

\subsubsection{Log-Transform of the Gamma Distribution}

We transform the Gamma Distribution with the Log-Transformation, i.e. $Y = \log(X), g(x) = \log(x), x(y) = g^{-1}(x) = \exp(x)$. Also, $\left\vert \frac{\partial x(y)}{\partial y}\right\vert = \exp(y)$. The transformed pdf is
\begin{subequations}
\begin{align}
\mathcal{G}_{Y_{\log}}(y, \alpha, \lambda) &= \frac{\lambda^\alpha}{\Gamma(\alpha)} \cdot x(y)^{(\alpha - 1)} \cdot e^{-\lambda x(y)} \cdot \exp(y) \\ 
&=\frac{\lambda^\alpha}{\Gamma(\alpha)} \cdot \exp(y)^{\alpha} \cdot e^{-\lambda \exp(y)} \\
&= \exp\left[\alpha y - \lambda\exp(y) - \Gamma(\alpha) + \alpha\log(\lambda) \right]
\end{align}
\end{subequations}
with exponential family parameters $h(y) = 1$, $\phi(y) = (y, \exp(y)), \eta = (\alpha, -\lambda)$ and $Z(\alpha, \lambda) = \log(\Gamma(\alpha)) - \alpha  \log(\lambda)$. 

\subsubsection*{Laplace Approximation of the log-transformed Gamma Distribution}

\begin{align*}
\text{log-pdf: } &= \alpha \log(\lambda) - \log(\Gamma(\alpha)) + \alpha y - \lambda \exp(y)\\
\text{1st derivative: }&  \alpha - \lambda \exp(y)\\
\text{mode: }& \alpha - \lambda \exp(y) = 0 \Leftrightarrow y = \log\left(\frac{\alpha}{\lambda}\right)\\
\text{2nd derivative: }&  -\lambda \exp(y)\\
\text{insert mode: }& -\lambda \exp(\log\left(\frac{\alpha}{\lambda}\right)) = -\alpha \\
\text{invert and times -1: }&\sigma^2 = \frac{1}{\alpha}
\end{align*}
The resulting Gaussian is $\mathcal{N}(y; \log\left(\frac{\alpha}{\lambda}\right), \frac{1}{\alpha})$.

\subsubsection*{The bridge for the log-transformation}

\begin{figure}[!htb]
	\centering
	\includegraphics[width=\textwidth]{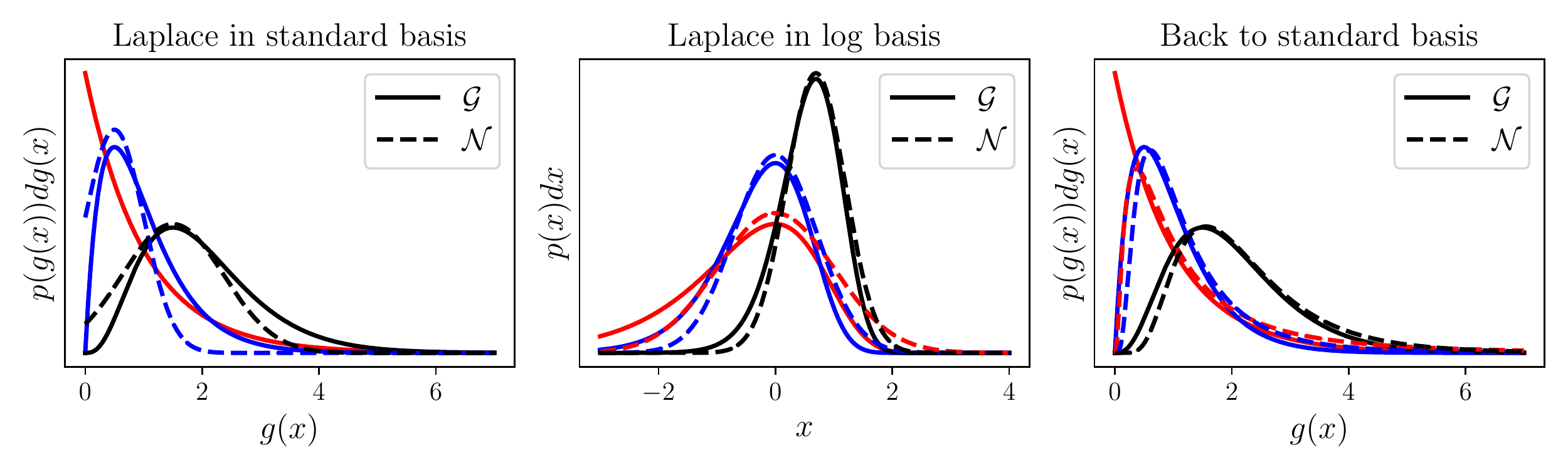}
	\caption{log-bridge for Gamma distribution}
	\label{fig:gamma_log_bridge}
\end{figure}

We already know how to get $\mu$ and $\sigma$ from $\lambda$ and $\alpha$. To invert we calculate $\mu = \log(\alpha/\lambda) \Leftrightarrow \lambda= \alpha/\exp(\mu)$ and insert $\alpha=1/\sigma^2$. In summary we have
\begin{subequations}
\begin{align}
\mu &= \log\left(\frac{\alpha}{\lambda}\right) \\
\sigma^2 &= \frac{1}{\alpha} \\
\lambda &=  \frac{1}{\exp(\mu)\sigma^2}\\
\alpha &= \frac{1}{\sigma^2}
\end{align}
\end{subequations}
A visual interpretation can be found in Figure \ref{fig:gamma_log_bridge}.

\subsubsection{Sqrt-Transform of the Gamma Distribution}

We transform the Gamma Distribution with the sqrt-transformation, i.e. $Y = \sqrt{X}, g(x) = \sqrt{x},x_1(y) =  g_1^{-1}(y) = -y^2, x_2(y) = g_2^{-1}(y) = y^2$ and $\left\vert\frac{\partial x_i(y)}{\partial y} \right\vert = \left\vert\frac{\partial g_i^{-1}(y)}{\partial y} \right\vert = \vert 2y \vert$. We use the same `trick' as in Appendix \ref{subsec:chi2-normal} to split up the transformation in two parts. 
\begin{subequations}
\begin{align}
\mathcal{G}_Y(y) &=  \frac{1}{2} \cdot \mathcal{G}_X(x_1(y)) \left\vert\frac{\partial x_1(y)}{\partial y} \right\vert \mathbf{1}_r(y) +  \frac{1}{2} \cdot\mathcal{G}_X(x_2(y)) \left\vert\frac{\partial x_2(y)}{\partial y} \right\vert \mathbf{1}_r(y) \\
&= \frac{1}{2y^2} \exp[\alpha \log(y^2) - \lambda y^2 - A(\alpha, \lambda)] |2y| \mathbf{1}_{(-\infty, 0)}(y) + \frac{1}{2y^2} \exp[\alpha \log(y^2) - \lambda y^2 - A(\alpha, \lambda)] |2y| \mathbf{1}_{[0, \infty)}(y) \\
&= \frac{1}{\sqrt{y}}\exp[2\alpha \log(y) - \lambda y^2 - A(\alpha, \lambda)] \mathbf{1}_{(-\infty, +\infty)}(y) \\
&= \frac{1}{\sqrt{y}}\exp[2\alpha \log(y) - \lambda y^2 - A(\alpha, \lambda)]
\end{align}
\end{subequations}
which is defined on the entirety of $\mathbb{R}$ and is an exponential family with $h(y) = \frac{1}{y},\, \phi(y)=(\log(y), y^2), \,w=(2\alpha, -\lambda)$ and $Z(\alpha, \lambda) = \log(\Gamma(\alpha)) - \alpha \log(\lambda)$.

\subsubsection*{Laplace Approximation of the sqrt-transformed Gamma Distribution}
\begin{align*}
\text{log-pdf: } &(2\alpha-1) \log(y) - \lambda y^2 + \alpha \log(\lambda) - \log(\Gamma(\alpha)) \\
\text{1st derivative: }&  \frac{2\alpha-1}{y} - 2\lambda y\\
\text{mode: }& \frac{2\alpha-1}{y} - 2\lambda y = 0 \Leftrightarrow y = \sqrt{\frac{\alpha-0.5}{\lambda}}\\
\text{2nd derivative: }&  -\frac{2\alpha-1}{x^2} - 2\lambda\\
\text{insert mode: }& -\frac{2\alpha-1}{\frac{\alpha-0.5}{\lambda}} - 2\lambda = -4\lambda\\
\text{invert and times -1: }& \sigma^2 = \frac{1}{4\lambda}
\end{align*}
The resulting Gaussian is $\mathcal{N}\left(y; \sqrt{\frac{\alpha-0.5}{\lambda}}, \frac{1}{4\lambda} \right)$.

\subsubsection*{The bridge for the sqrt-transformation}

\begin{figure}[!htb]
	\centering
	\includegraphics[width=\textwidth]{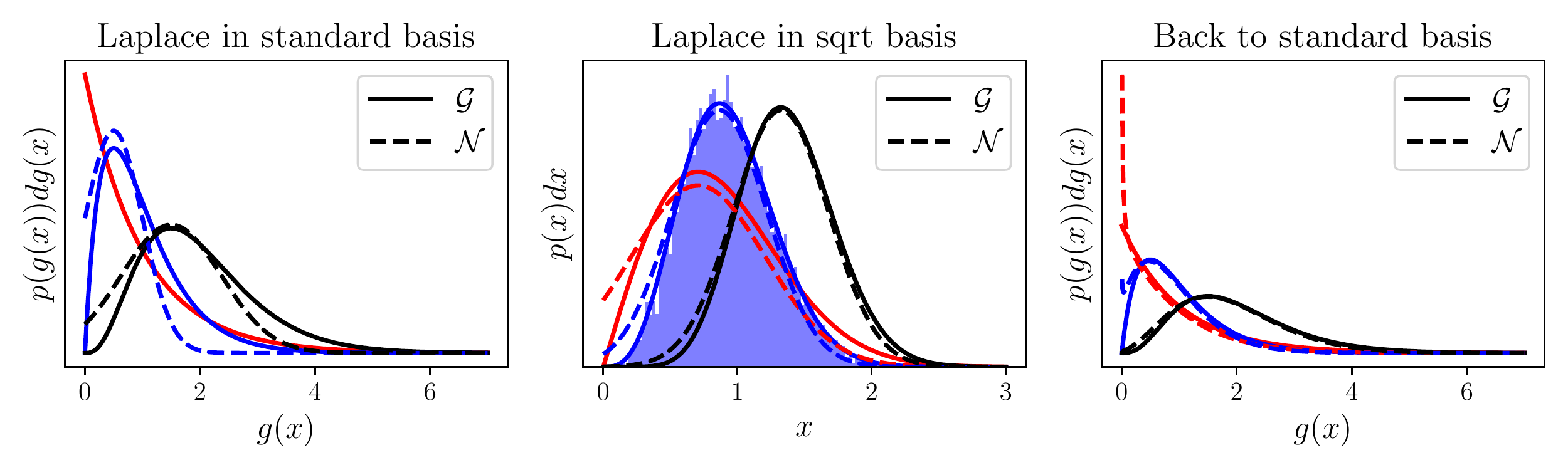}
	\caption{sqrt-bridge for the Gamma distribution. The transformed samples indicate that the closed-form transformation is correct.}
	\label{fig:gamma_sqrt_bridge}
\end{figure}

We already know how to get $\mu$ and $\sigma$ from $\lambda$ and $\alpha$. To invert we calculate $\mu = \sqrt{\frac{\alpha-0.5}{\lambda}} \Leftrightarrow \alpha = \frac{\mu^2}{\lambda}-0.5$ and insert $\lambda=\frac{4}{\sigma^2}$. In summary we have
\begin{subequations}
\begin{align}
\mu &= \sqrt{\frac{\alpha-0.5}{\lambda}} \\
\sigma^2 &= \frac{1}{4\lambda} \\
\lambda &= \frac{4}{\sigma^2} \\
\alpha &= \frac{4\mu^2}{\sigma^2}+0.5 
\end{align}
\end{subequations}
A visual interpretation can be found in Figure \ref{fig:gamma_sqrt_bridge}.
\subsection{Inverse Gamma Distribution}
\label{subsec:inv_gamma_dist}

\subsubsection{Standard Inverse Gamma Distribution}

The pdf of the inverse Gamma is 
\begin{subequations}
\begin{align}
	\mathcal{IG}(x, \alpha, \lambda) &= \frac{\lambda^{\alpha}}{\Gamma(\alpha)} x^{-\alpha-1} \exp(-\frac{\lambda}{x}) \\
	&=\exp \frac{1}{x}\left[-\alpha\log(x) - \lambda/x + \alpha \log(\lambda) -\log\Gamma(\alpha)\right]
	\label{eq:pdf_inverse_gamma}
\end{align}
\end{subequations}
with exponential family values $h(x) = \frac{1}{x}$, $\phi(x)=(\log(x), x), w= (-\alpha, -\lambda)$ and $Z(\alpha,\lambda) = \log\Gamma(\alpha) - \alpha\log\lambda$.

\subsubsection*{Laplace Approximation of the standard inverse gamma distribution}
\begin{align*}
\text{log-pdf: } &(-\alpha-1)\log(y) - \lambda/x + \alpha \log(\lambda) -\log\Gamma(\alpha) \\
\text{1st derivative: }&  \frac{-\alpha-1}{y} + \frac{\lambda}{y^2}\\
\text{mode: }& \frac{-\alpha-1}{y} + \frac{\lambda}{y^2} = 0 \Leftrightarrow y = \frac{\lambda}{a+1}\\
\text{2nd derivative: }&  \frac{\alpha+1}{y^2} - 2\frac{\lambda}{y^3}\\
\text{insert mode: }& \frac{\alpha+1}{\frac{\lambda}{a+1}^2} - 2\frac{\lambda}{\frac{\lambda}{a+1}^3} = -\frac{(\alpha +1)^3}{\lambda^2} \\
\text{invert and times -1: }&\sigma^2 = \frac{\lambda^2}{(\alpha +1)^3}
\end{align*}
Therefore the resulting Gaussian is $q(y) = \mathcal{N}(y; \mu = \frac{\lambda}{a+1}, \sigma^2 =\frac{\lambda^2}{(\alpha +1)^3} )$. 

\subsubsection{Log-Transform of the inverse Gamma distribution}

We transform the Inverse Gamma distribution with the log-transformation, i.e. $Y = \log(X)$. Therefore $g(x) = \log(x)$, and thereby $x(y) = g^{-1}(x) = \exp(y)$. It follows that the new pdf is 
\begin{subequations}
\begin{align}
	\mathcal{IG}_{Y_{\log}}(y, \alpha, \lambda) &= \frac{\lambda^{\alpha}}{\Gamma(\alpha)} \exp(y)^{-\alpha-1} \exp(-\lambda/\exp(y)) \cdot \exp(y) \\
	&=  \frac{\lambda^{\alpha}}{\Gamma(\alpha)} \exp(y)^{-\alpha} \exp(-\lambda/\exp(y)) \\
	&= \exp\left[-\alpha x - \frac{\lambda}{\exp(x)} - \log\Gamma(\alpha) + \alpha\log(\lambda)\right]
	\label{eq:inv_gamma_trans_pdf}
\end{align}
\end{subequations}
with exponential family values $h(y) = 1, \phi(y) = (y, \exp(y)), w=(-\alpha, \lambda)$ and $Z(\alpha, \lambda) = \log\Gamma(\alpha) - \alpha \log \lambda$.

\subsubsection*{Laplace Approximation of the log-transformed Inverse Gamma Distribution}
\begin{align*}
\text{log-pdf: } &-\alpha y - \frac{\lambda}{\exp(y)} + Z(\alpha, \lambda) \\
\text{1st derivative: }&  -\alpha + \frac{\lambda}{\exp(y)}\\
\text{mode: }&  -\alpha + \frac{\lambda}{\exp(y)} = 0 \Leftrightarrow y = \log(\lambda/\alpha)\\
\text{2nd derivative: }&  -\frac{\lambda}{\exp(y)}\\
\text{insert mode: }&  -\frac{\lambda}{\exp(\log(\lambda/\alpha))} = -\alpha\\
\text{invert and times -1: }&\sigma^2 = \frac{1}{\alpha}
\end{align*}
Therefore the resulting Gaussian is $\mathcal{N}(y; \log(\frac{\lambda}{\alpha}), \frac{1}{\alpha})$.

\subsubsection*{The Bridge for the log-transformed Inverse Gamma Distribution}
\begin{figure}[!htb]
	\centering
	\includegraphics[width=\textwidth]{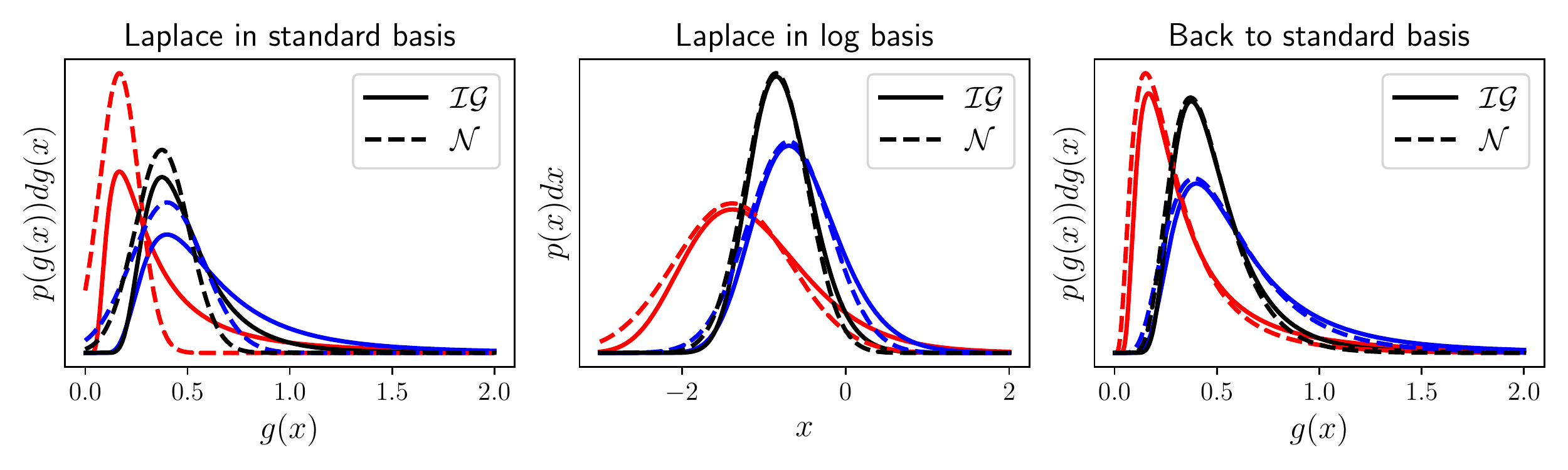}
	\caption{inverse gamma log-bridge}
	\label{fig:inverse_gamma_log_bridge}
\end{figure}

Inverting the equations from above we get in summary
\begin{subequations}
\begin{align}
	\mu &= \log\left(\frac{\lambda}{\alpha}\right) \\
	\sigma^2 &= \frac{1}{\alpha} \\
	\alpha &= \frac{1}{\sigma^2}\\
	\lambda &= \frac{\exp(\mu)}{\sigma^2}
\end{align}
\end{subequations}
For a visual interpretation consider Figure \ref{fig:inverse_gamma_log_bridge}.

\subsubsection{Sqrt-Transform of the inverse Gamma distribution}

We transform the Inverse Gamma Distribution with the sqrt-transformation, i.e. $Y = \sqrt{X}, g(x) = \sqrt{x},x_1(y) =  g_1^{-1}(y) = -y^2, x_2(y) = g_2^{-1}(y) = y^2$ and $\left\vert\frac{\partial x_i(y)}{\partial y} \right\vert = \left\vert\frac{\partial g_i^{-1}(y)}{\partial y} \right\vert = \vert 2y \vert$. We use the same `trick' as in Appendix \ref{subsec:chi2-normal} to split up the transformation in two parts. 
\begin{subequations}
\begin{align}
\mathcal{G}^{-1}_{Y_{\text{sqrt}}}(y) &=  \frac{1}{2} \cdot \mathcal{G}^{-1}_X(x_1(y)) \left\vert\frac{\partial x_1(y)}{\partial y} \right\vert \mathbf{1}_r(y) +  \frac{1}{2} \cdot\mathcal{G}^{-1}_X(x_2(y)) \left\vert\frac{\partial x_2(y)}{\partial y} \right\vert \mathbf{1}_r(y) \\ 
&=\frac{1}{2}\frac{\lambda^{\alpha}}{\Gamma(\alpha)} y^{-2\alpha-1} \exp(-\lambda/y^2)) |2y| \mathbf{1}_{(-\infty, 0)}(y) + \frac{1}{2}\frac{\lambda^{\alpha}}{\Gamma(\alpha)} y^{-2\alpha-1} \exp(-\lambda/y^2)) |2y| \mathbf{1}_{[0, \infty)}(y) \\
&= \frac{\lambda^{\alpha}}{\Gamma(\alpha)} y^{-2\alpha-1} \exp(-\lambda/y^2)) \mathbf{1}_{(-\infty, +\infty)}(y) \\
&= \frac{1}{\sqrt{y}}\exp\left[(-2\alpha-1) \log(y) - \frac{\lambda}{y^2} - \log\Gamma(\alpha) + \alpha\log(\lambda)\right]
\label{eq:inv_gamma_sqrt_pdf}
\end{align}
\end{subequations}
with exponential family values $h(y) = \frac{1}{\sqrt{y}}, \phi(y) = (\log(y), y^2), w=(-2\alpha, -\lambda)$ and $Z(\alpha, \lambda) = \log\Gamma(\alpha) - \alpha \log \lambda$.

\subsubsection*{Laplace Approximation of the sqrt-transformed Inverse Gamma Distribution}
\begin{align*}
\text{log-pdf: } &(-2\alpha-1)\log(y) - \frac{\lambda}{y^2} + Z(a,\lambda) \\
\text{1st derivative: }& -\frac{2\alpha+1}{y} + 2\frac{\lambda}{y^3} \\
\text{mode: }&  y = \sqrt{\frac{\alpha+0.5}{\lambda}}\\
\text{2nd derivative: }&  \frac{2\alpha+1}{y^2} - 6\frac{\lambda}{y^4}\\
\text{insert mode: }&  -4\frac{(\alpha+0.5)^2}{\lambda}\\
\text{invert and times -1: }&\sigma^2 = \frac{\lambda}{4 (\alpha+0.5)^2}
\end{align*}
yielding $q(y) = \mathcal{N}(y; \mu = \sqrt{\frac{\alpha+0.5}{\lambda}}, \sigma^2 = \frac{\lambda}{4 (\alpha+0.5)^2})$. 
\subsubsection*{The Bridge for the sqrt-transformed Inverse Gamma Distribution}
\begin{figure}[!htb]
	\centering
	\includegraphics[width=\textwidth]{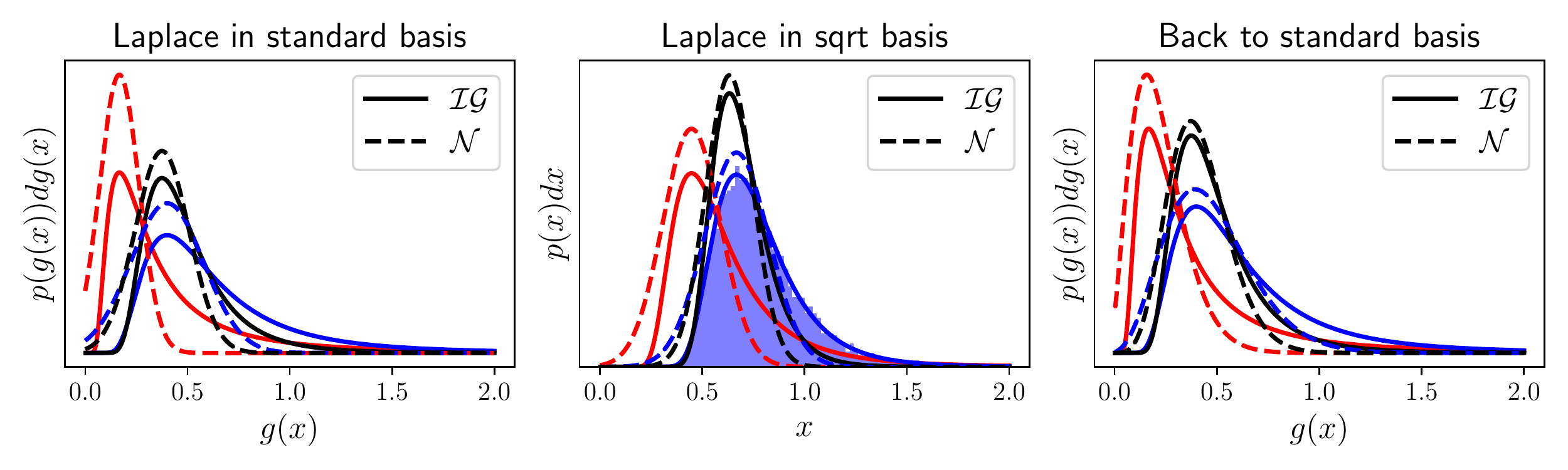}
	\caption{inverse gamma sqrt-bridge. Transformed samples indicate that the closed-form transformation is correct.}
	\label{fig:inverse_gamma_sqrt_bridge}
\end{figure}

We get $\alpha$ and $\lambda$ by inverting the equations for $\mu$ and $\sigma$. In summary we get
\begin{subequations}
\begin{align}
\mu &= \sqrt\frac{\lambda}{\alpha} \\
\sigma^2 &= \frac{\lambda}{4\alpha^2} \\
\alpha &= \frac{\mu^2}{4\sigma^2}-0.5\\
\lambda &= \frac{\mu^4}{4\sigma^2}
\end{align}
\end{subequations}
For a visual interpretation consider Figure \ref{fig:inverse_gamma_sqrt_bridge}.
\subsection{Chi-square Distribution}
\label{subsec:chi2_dist}

\subsubsection{Standard Chi-square distribution}

The pdf of the Chi-square distribution in the standard basis is
\begin{subequations}
\begin{align}\label{eq:chi2_pdf}
	\chi^2_X(x, k) &= \frac{1}{2^{k/2}\Gamma(k/2)}  x^{k/2 -1} \exp(-x/2) \\
	&= \frac{1}{x}\exp \left[(k/2)\log(x) - x/2 - \log(2^{k/2}\Gamma(k/2))\right]
\end{align} 
\end{subequations}
with exponential family values $h(x)=\frac{1}{x}$, $\phi(x) = (\log(x), x), w = k/2$ and $Z(k) = \log(2^{k/2}\Gamma(k/2))$.

\subsubsection*{Laplace approximation of the standard Chi-square distribution}

\begin{align*}
\text{log-pdf: } &(k/2-1)\log(x) - x/2 - \log(2^{k/2}\Gamma(k/2)) \\
\text{1st derivative: }&  \frac{k/2-1}{x} - \frac{1}{2} \\
\text{mode: }&  \frac{k/2-1}{x} - \frac{1}{2} = 0 \Leftrightarrow x = k-2\\
\text{2nd derivative: }&  -\frac{k/2-1}{x^2}\\
\text{insert mode: }& -\frac{k/2-1}{(k-2)^2} = -\frac{(k-2)}{2(k-2)^2}\\
\text{invert and times -1: }&\sigma^2 = 2(k-2)
\end{align*}
The resulting Gaussian is therefore $\mathcal{N}(x; k-2, 2(k-2))$ for $k>2$.

\subsubsection{Log-Transformed Chi-square distribution}

we transform the distribution with $g(x) = \log(x)$, i.e. $x(y) = g^{-1}(x) = \exp(y)$. The new pdf becomes
\begin{subequations}
\begin{align}
	\chi^2_{Y_{\log}}(y,k) &= \frac{1}{2^{k/2}\Gamma(k/2)}  \exp(y)^{k/2 -1} \exp(-\exp(y)/2) \cdot \exp(y) \\
	&= \frac{1}{2^{k/2}\Gamma(k/2)}  \exp(y)^{k/2} \exp(-\exp(y)/2) \\
	&= \exp\left[\frac{k}{2}y - \frac{\exp(y)}{2} - \log(2^{k/2}\Gamma(k/2))\right]
\end{align}
\end{subequations}
with $h(y) = 1, \phi(y) =(y, \exp(y)), \eta=(k/2)$ and $Z(k) =  \log(2^{k/2}\Gamma(k/2))$.

\subsubsection*{Laplace approximation of the log-transformed Chi-square distribution}

\begin{align*}
\text{log-pdf: } &\frac{k}{2}y - \frac{\exp(y)}{2} - \log(2^{k/2}\Gamma(k/2)) \\
\text{1st derivative: }&  \frac{k}{2} - \frac{\exp(y)}{2} \\
\text{mode: }& k/2 - \frac{\exp(y)}{2} = 0 \Leftrightarrow y = \log(k)\\
\text{2nd derivative: }&  -\frac{\exp(y)}{2}\\
\text{insert mode: }& -\frac{\exp(y)}{2} = -k/2\\
\text{invert and times -1: }&\sigma^2 = 2/k
\end{align*}
which yields the Laplace Approximation $q(y) = \mathcal{N}(y; \mu= \log(k), \sigma^2 = 2/k)$.

\subsubsection*{The Bridge for log-transform}
\begin{figure}[!htb]
	\centering
	\includegraphics[width=\textwidth]{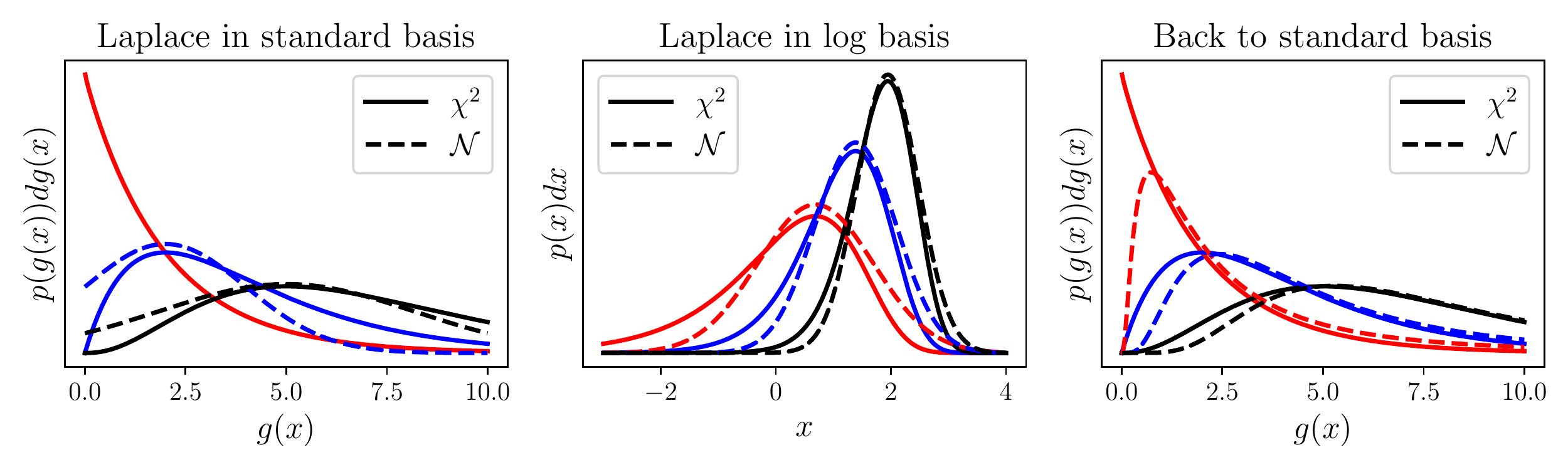}
	\caption{Log-bridge for the Chi-square distribution.}
	\label{fig:chi2_log_bridge}
\end{figure}

In summary we get
\begin{subequations}
\begin{align}
	\mu &= \log(k) \\
	\sigma^2 &= 2/k \\
	k &= \exp(\mu) \\
	\text{or}\quad k &= 2/\sigma^2
\end{align}
\end{subequations}
For a visual interpretation consider Figure \ref{fig:chi2_log_bridge}.

\subsubsection{Sqrt-Transformed Chi-square distribution}

We transform the chi-square distribution with the sqrt-transformation, i.e. $Y = \sqrt{X}, g(x) = \sqrt{x},x_1(y) =  g_1^{-1}(y) = -y^2, x_2(y) = g_2^{-1}(y) = y^2$ and $\left\vert\frac{\partial x_i(y)}{\partial y} \right\vert = \left\vert\frac{\partial g_i^{-1}(y)}{\partial y} \right\vert = \vert 2y \vert$. We use the same `trick' as in Appendix \ref{subsec:chi2-normal} to split up the transformation in two parts. 
\begin{subequations}
\begin{align}
    \chi^2_{Y_{\text{sqrt}}}(y) &=  \frac{1}{2} \cdot \chi^2_X(x_1(y)) \left\vert\frac{\partial x_1(y)}{\partial y} \right\vert \mathbf{1}_r(y) +  \frac{1}{2} \cdot\chi^2_X(x_2(y)) \left\vert\frac{\partial x_2(y)}{\partial y} \right\vert \mathbf{1}_r(y) \\                                        &=\frac{1}{2}\frac{1}{2^{k/2}\Gamma(k/2)}  y^{k-1} \exp(-\frac{y^2}{2}) |2y| \mathbf{1}_{(-\infty, 0)}(y) + \frac{1}{2}\frac{1}{2^{k/2}\Gamma(k/2)}  y^{k-1} \exp(-\frac{y^2}{2}) |2y| \mathbf{1}_{[0, \infty)}(y) \\
    &= \frac{1}{2^{k/2}\Gamma(k/2)}  y^{k-1} \exp(-\frac{y^2}{2}) \mathbf{1}_{(-\infty, +\infty)}(y) \\
    &= \exp \left[(k-1)\log(y) - \frac{y^2}{2} - \log(2^{k/2}\Gamma(k/2))\right]
\end{align}
\end{subequations}
with exponential family values $h(y) = 1, \phi(y)=(\log(y), y^2), w=(k, 1/2)$ and $Z(k) =  \log(2^{k/2}\Gamma(k/2))$. 

\subsubsection*{Laplace approximation of the sqrt-transformed Chi-square distribution}

\begin{align*}
\text{log-pdf: } &((k-1)\log(x) - \frac{x^2}{2} - \log(2^{k/2}\Gamma(k/2)) \\
\text{1st derivative: }&  \frac{k}{x} - x \\
\text{mode: }& \frac{k-1}{x} -x = 0 \Leftrightarrow x = \sqrt{k-1}\\
\text{2nd derivative: }&  -\frac{k-1}{x^2} - 1\\
\text{insert mode: }& -\frac{k-1}{k-1}-1 = -2\\
\text{invert and times -1: }&\sigma^2 = 1/2
\end{align*}
yielding $q(y) = \mathcal{N}(y; \mu = \sqrt{k}, \sigma^2 = 1/2)$.

\subsubsection*{The Bridge for sqrt-transform}
\begin{figure}[!htb]
	\centering
	\includegraphics[width=\textwidth]{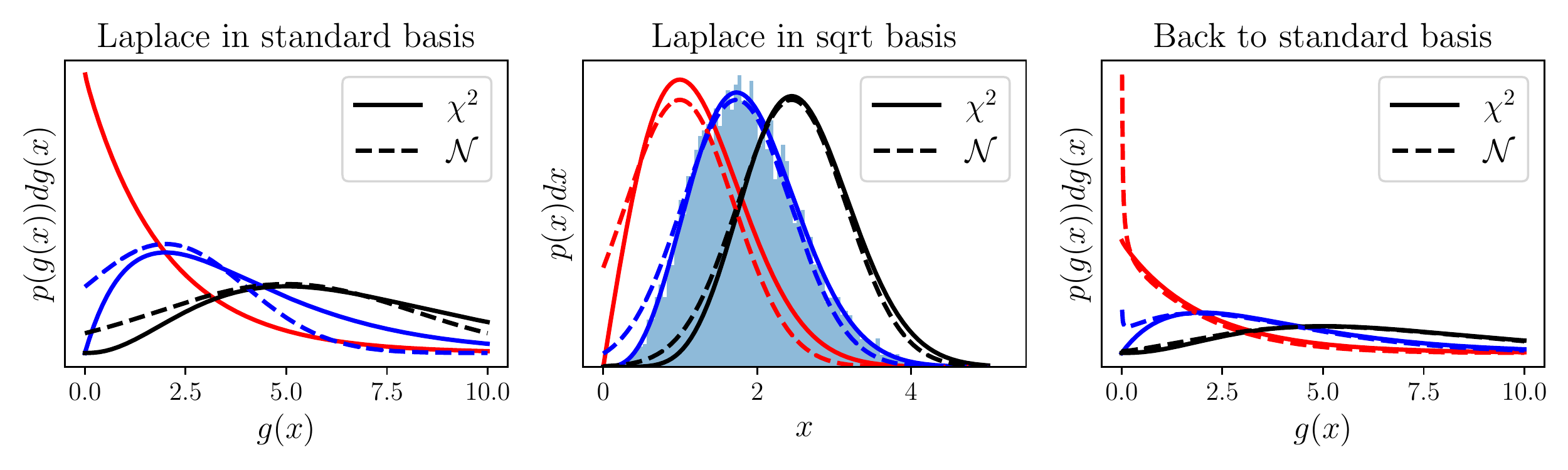}
	\caption{Sqrt-bridge for the Chi-square distribution. Transformed samples indicated that the closed-form transformation is correct.}
	\label{fig:chi2_sqrt_bridge}
\end{figure}

In summary we have
\begin{subequations}
\begin{align}
\mu &= \sqrt{k-1} \\
\sigma^2 &= 1/2 \\
k &= \mu^2 + 1
\end{align}
\end{subequations}
For a visual interpretation consider Figure \ref{fig:chi2_sqrt_bridge}.
\subsection{Beta Distribution}
\label{subsec:beta_dist}

\subsubsection{Standard Beta Distribution}

The pdf of the Beta distribution in the standard basis is
\begin{subequations}
\begin{align}
	\mathcal{B}(x, \alpha, \beta) &= \frac{x^{(\alpha - 1)} \cdot (1-x)^{(\beta-1)}}{B(\alpha, \beta)} \\
	 &=  \exp\left[(\alpha-1) \log(x) + (\beta-1)\log(1-x) - \log(B(\alpha,\beta)))\right]\\
	&= \frac{1}{x(1-x)}\exp\left[\alpha\log(x) + \beta\log(1-x) - \log(B(\alpha,\beta)))\right]
\end{align}
\end{subequations}
with exponential family values $h(x) = \frac{1}{x(1-x)}, \phi(x)=(\log(x), \log(1-x), w = (\alpha, \beta)$ and $Z(\alpha, \beta) = \log(B(\alpha,\beta)))$ where $B(\alpha, \beta) = \frac{\Gamma(\alpha)\Gamma(\beta)}{\Gamma(\alpha + \beta)}$ and $\Gamma(x)$ is the Gamma function.

\subsubsection*{Laplace approximation of the standard Beta distribution}

\begin{align*}
\text{log-pdf: } &\log\left( \frac{x^{(\alpha - 1)} \cdot (1-x)^{(\beta-1)}}{B(\alpha, \beta)} \right) \\
&= (\alpha-1) \log(x) + (\beta-1)\log(1-x) - \log(B(\alpha,\beta)))\\
\text{1st derivative: }& \frac{(\alpha-1)}{x} - \frac{(\beta-1)}{1-x}  \\
\text{mode: }& \frac{(\alpha-1)}{x} - \frac{(\beta-1)}{1-x}  = 0 \Leftrightarrow x = \frac{\alpha-1}{\alpha + \beta - 2} \\
\text{2nd derivative: }& \frac{\alpha -1}{x^2} + \frac{\beta - 1}{(1 - x)^2} \\
\text{insert mode: }& \frac{\alpha -1}{\frac{\alpha-1}{\alpha + \beta - 2}^2} + \frac{\beta - 1}{(1 - \frac{\alpha-1}{\alpha + \beta - 2})^2} = \frac{(\alpha + \beta - 2)^3}{(\alpha-1)(\beta-1)}\\
\text{invert: }& \frac{(\alpha -1)(\beta-1)}{(\alpha + \beta - 2)^3}
\end{align*}
The Beta distribution in standard basis is therefore approximated by $N(\mu = \frac{\alpha-1}{\alpha + \beta - 2}, \sigma^2 = \frac{(\alpha -1)(\beta-1)}{(\alpha + \beta - 2)^3})$.

\subsubsection{Logit-Transform of the Beta distribution}

We transform the Beta distribution using $g(x) = \log(\frac{x}{1-x})$. Therefore $x(y) = g^{-1}(y) = \sigma(y) = \frac{1}{1+ \exp(-y)}$. This yields the following pdf
\begin{subequations}
\begin{align}
	\mathcal{B}_{Y_\text{logit}}(y, \alpha, \beta) &= \frac{1}{\sigma(y)(1-\sigma(y))}\exp\left[\alpha\log(\sigma(y))) + \beta\log(1-\sigma(y)) - \log(B(\alpha,\beta)))\right] \cdot (\sigma(y)(1-\sigma(y)) \\
	&= \exp\left[\alpha\log(\sigma(y)) + \beta\log(1-\sigma(y)) - \log(B(\alpha,\beta)))\right]
	\label{eq:beta_logit_trans_pdf}
\end{align}
\end{subequations}
with exponential family values $h(y) = 1, \phi(y)=(\log(\sigma(y)), \log(1-\sigma(y)), w = (\alpha, \beta)$ and $Z(\alpha, \beta) = \log(B(\alpha, \beta))$.

\subsubsection*{Laplace approximation of the logit transformed Beta distribution}

\begin{align*}
\text{log-pdf: } &\log\left( \frac{\sigma(y)^{\alpha} \cdot (1 - \sigma(y)^{\beta})}{B(\alpha, \beta)} \right) \\
&= \alpha \log(\sigma(y)) + \beta \log(1 - \sigma(x)) - \log(B(\alpha, \beta))\\
\text{1st derivative: }&  \alpha (1 - \sigma(y)) - \beta \sigma(y)\\
\text{mode: }& \alpha (1 - \sigma(y)) - \beta \sigma(y) = 0 \Leftrightarrow y = \log(\frac{\alpha}{\beta}) \\
\text{2nd derivative: }& (\alpha + \beta)\sigma(y)(1 - \sigma(y))  \\
\text{insert mode: }& (\alpha + \beta)\sigma(-\log(\frac{\beta}{\alpha}))(1 - \sigma(-\log(\frac{\beta}{\alpha}))) = \frac{\alpha\beta}{\alpha + \beta}  \\
\text{invert: }& \frac{\alpha + \beta}{\alpha \beta}
\end{align*}
The Laplace approximation of the Beta in the logit base is therefore given by $\mathcal{N}(y; \mu=\log(\frac{\alpha}{\beta}), \sigma^2 = \frac{\alpha + \beta}{\alpha \beta})$.

\subsubsection*{The Bridge for the logit transformation}

\begin{figure}[!htb]
	\centering
	\includegraphics[width=\textwidth]{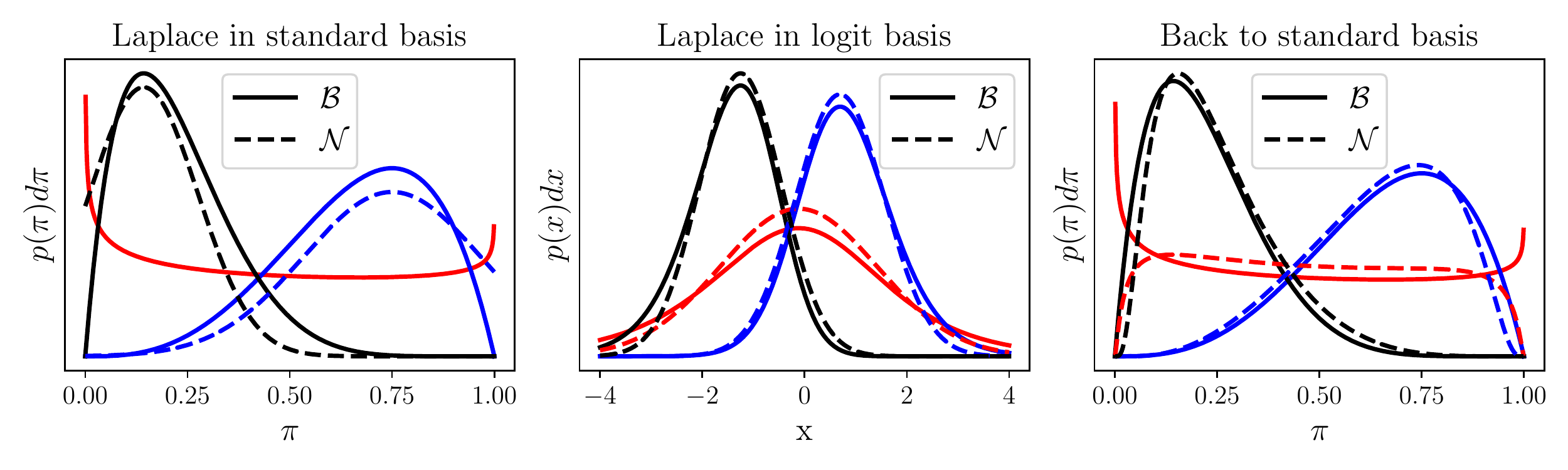}
	\caption{beta logit bridge}
	\label{fig:beta_logit_bridge}
\end{figure}

In summary we have
\begin{subequations}
\begin{align}
	\mu &=\log\left(\frac{\alpha}{\beta}\right) \\
	\sigma^2 &= \frac{\alpha + \beta}{\alpha \beta} \\
	\alpha &= \frac{\exp(\mu) + 1}{\sigma^2} \\
	\beta &= \frac{\exp(-\mu) + 1}{\sigma^2} 
\end{align}
\end{subequations}
For a visual interpretation consider Figure \ref{fig:beta_logit_bridge}.

\subsection{Dirichlet Distribution}
\label{subsec:Dirichlet_dist}

\subsubsection{Standard Dirichlet distribution}

The pdf for the Dirichlet distribution in the standard basis (i.e. probability space) is

\begin{subequations}
\begin{align}
\mathrm{Dir}(\vpi | \valpha) &= \frac{\Gamma \left( \sum_{k=1}^K \alpha_k \right)}{\prod_{k=1}^K \Gamma(\alpha_k)} \prod_{k=1}^K \pi_k^{\alpha_k-1} \\
&= \frac{1}{B(\alpha)} \prod_{k=1}^K \pi_k^{\alpha_k-1} \\
&= \exp\left[\sum_k(\alpha_k-1)\log(\pi_k) - \log(B(\alpha))\right] \\
&= \frac{1}{\prod_k \pi_k} \exp\left[\sum_k\alpha_k\log(\pi_k) - \log(B(\alpha))\right] \\
\end{align}
\end{subequations}
with sufficient statistics $\phi(x_i) = \log(x_i)$, natural parameters $w_i=\alpha_i$, base measure $h(x) = \prod_k x_k$, and partition function $Z(w) = \log(B(\alpha))$.

\subsubsection*{Laplace approximation of the standard Dirichlet distribution}

\begin{align*}
\text{log-pdf: } & f = \sum_k(\alpha_k-1)\log(x_k) - \log(B(\alpha)) \\
\text{1st derivative: }&  \frac{\partial f}{\partial x_i} =  \frac{\alpha_i-1}{x_i}\\
\text{mode: }& x_i = \frac{(\alpha_i - 1)}{\sum_k \alpha_k - K} \\
\text{2nd derivative: }&  \frac{\partial^2 f}{\partial x_i \partial x_j} = - \delta_{ij} \frac{(\alpha_i - 1)}{x_i^2} \\
\text{insert mode: }& - \delta_{ij}\frac{(\sum_k \alpha_k - K)^2}{(\alpha_i - 1)} \\
\text{invert and times -1: }&\Sigma_{ij} = \delta_{ij} \frac{\alpha_i - 1}{(\sum_k \alpha_k -K)^2}
\end{align*}
Which yields a diagonal Covariance matrix for the Laplace approximation.

\subsubsection{Softmax-Transform of the Dirichlet distribution}

We aim to transform the basis of this distribution from base $\vy$ via the softmax transform to be in the new base $\pi$:
\begin{equation}\label{eq:softmax}
\pi_k(\vy) := \frac{\exp(y_k)}{\sum_{l=1}^K \exp(y_l)} \, ,
\end{equation}
David J. MacKay used this transformation in \citep{MacKay1998} already. We provide another, potentially simpler, derivation for the Laplace approximation of the Dirichlet in the softmax basis. 

The softmax transform has no analytic inverse $\pi_k^{-1}(y)$. However, the inverse-softmax is not necessary for our computation since we assume $\pi(y)$ to be the inverse transformation already (i.e. $g^{-1}(y) = \pi(y)$). Our transformation is from a variable in $\mathbb{R}^d$ (which has $d$ degrees of freedom) to a variable that is in $\mathbb{P}^d$ (which has $d-1$ degrees of freedom). To account for the difference in size of the two spaces we create a helper variable for the transformation as described in the following.

We want to transform $D$ variables $y_i$ from $\mathbb{R}^d$ to $\tau_i = \exp(y_i)$. For $\tau_i$ to be equal to $\pi_i$ we need to ensure that it sums to 1, $u = \sum_i \tau_i = 1$. With the helper-variable $u$ our variable transform $g(\pi, u)$ becomes
\begin{align}
	p_{y,u}(\pi(y), u) &= p_{\pi, u}(\pi(y), u) |\det g(\pi(y), u)|
\end{align}
with 
\begin{subequations}
\begin{align}
	|\det g(\pi(y), u)| &= \left|\det\begin{pmatrix}
		\frac{\partial \tau_1}{\partial y_1} & \cdots & \frac{\partial \tau_k}{\partial y_1} & \frac{\partial u}{\partial y_1} \\
		\vdots & \ddots & \vdots & \vdots \\
		\frac{\partial \tau_1}{\partial y_k} & \cdots & \frac{\partial \tau_k}{\partial y_k} & \frac{\partial u}{\partial y_k} \\
		\frac{\partial \tau_1}{\partial u} & \cdots & \frac{\partial \tau_k}{\partial u} & \frac{\partial u}{\partial u}
	\end{pmatrix} \right| \\
	&= \left|\det\begin{pmatrix}
	\tau_1 = \exp(y_1) & \cdots & 0 & \tau_1 \\
	\vdots & \ddots & \vdots & \vdots \\
	0 & \cdots & \tau_k & \tau_k \\
	0 & \cdots & 0 & 1
	\end{pmatrix} \right| \\
	&= \prod_i^{K} \tau_i 
\end{align}
\end{subequations}
To get $p_y(\pi(y))$ we have to integrate out $u$. 
\begin{subequations}
\begin{align}
	p_y(\pi(y)) &=\int_{-\infty}^{\infty} p_{\pi, u}(\pi(y), u) |\det g(\pi, u)| du \\
	&= \int_{-\infty}^{\infty} p_{\pi}(\pi(y))  \prod_i^{K} \tau_i  du \\
	&= p_{\pi}(\pi(y)) \int_{0}^{\infty}  \prod_i^{K} \tau_i \delta(u-1) du \\
	&= p_{\pi}(\pi(y)) \int_{0}^{\infty}  \prod_i^{K} \tau_i \frac{u}{u} \delta(u-1) du \\
	&= p_{\pi}(\pi(y)) \int_{0}^{\infty}  \underbrace{\prod_i^{K} \pi_i u}_{f(u)} \delta(u-1) du \\
	&= p_{\pi}(\pi(y)) \cdot \prod_i^{K} \pi_i(y) \\
	&= \frac{1}{\prod_k \pi_k(y)} \exp\left[\sum_k\alpha_k\log(\pi_k(y)) - \log(B(\alpha))\right] \prod_k^{K} \pi_k(y) \\
	&= \exp\left[\sum_k\alpha_k\log(\pi(y_k)) - \log(B(\alpha))\right]
\end{align}
\end{subequations}
where we used the fact that $u > 0$ since it is a sum of exponentials and $\frac{\tau_i(y)}{u} = \pi_i(y)$. We multiplied with $\delta(u-1)$ since this transformation is only valid if $\sum_i \tau_i = u = 1$ because otherwise it is not a probability space. Additionally, we use
\begin{equation}
	\int_{-\infty}^{\infty} f(x)\delta(x-t)dx = f(t)
\end{equation}
which is known as the shifting property or sampling property of the Dirac delta function $\delta$. Using all of the above we get the pdf of the Dirichlet distribution in the new basis $\vy$: 
\begin{subequations}
\begin{align}\label{eq:dirichlet_softmax}
\mathrm{Dir}_{\vy}(\vpi(\vy) | \valpha) &:= \frac{\Gamma \left( \sum_{k=1}^K \alpha_k \right)}{\prod_{k=1}^K \Gamma(\alpha_k)} \prod_{k=1}^K \pi_k(\vy)^{\alpha_k}  \\
&= \exp\left[\sum_k\alpha_k\log(\pi(y_k)) - \log(B(\alpha))\right]
\end{align}
\end{subequations}
with sufficient statistics $\phi(y_i) = \log(\pi_i(y))$, natural parameters $w_i = \alpha_i$, base measure $h(y) = 1$ and normalizing constant $Z = \log(B(\alpha))$.

\subsubsection*{Laplace approximation of the softmax-transformed Dirichlet distribution}

The inversion of the Laplace approximation in the softmax basis has been provided in \citep{Hennig2010}. The following is merely a summary. 
The mean and mode of the inverse-softmax Dirichlet are identical, i.e. $\vpi(\vy) = \frac{\mathbf{\alpha}}{\sum_i \alpha_i}$. For a visual confirmation of this fact, consider the figure of the Beta distribution in Subsection \ref{subsec:beta_dist}. Additionally, the elements of $\vy$ must sum to zero. These two constraints combined yield only one possible solution for $\vmu$.
\begin{equation}
\mu_k = \log \alpha_k  - \frac{1}{K} \sum_{l=1}^{K} \log \alpha_l
\label{eq:mu_k}
\end{equation}
Calculating the covariance matrix $\vSigma$ is more complicated but detailed in the following. The logarithm of the Dirichlet is, up to additive constants
\begin{equation}
\log p_y(y|\alpha) = \sum_k \alpha_k \pi_k 
\end{equation}
Using $\pi_k$ as the softmax of $\vy$ as shown in Equation \ref{eq:softmax} we can find the elements of the Hessian $\vL$
\begin{equation}
L_{kl} = \hat{\alpha}(\delta_{kl}\hat{\pi_k} - \hat{\pi_k} \hat{\pi_l})
\end{equation}
where $\hat{\valpha} := \sum_k \alpha_k$ and $\hat{\vpi} = \frac{\alpha_k}{\hat{\alpha}}$ for the value
of $\vpi$ at the mode. Analytically inverting $\vL$ is done via a lengthy derivation using the fact that we can write $\vL = \mA + \mX\mB\mX^\top$ and inverting it with the Schur-complement. This process results in the inverse of the Hessian
\begin{equation}
L_{kl}^{-1} = \delta_{kl} \frac{1}{\alpha_k} - \frac{1}{K} \left[\frac{1}{\alpha_k} + \frac{1}{\alpha_l} - \frac{1}{K}\left(\sum_u^K \frac{1}{\alpha_u}\right) \right]
\end{equation}
We are mostly interested in the diagonal elements, since we desire a sparse encoding for computational reasons and we otherwise needed to map a $K \times K$ covariance matrix to a $K\times 1$ Dirichlet parameter vector which would be a very overdetermined mapping. Note that $K$ is a scalar not a matrix. The diagonal elements of $\vSigma = \vL^{-1}$ can be calculated as
\begin{equation}
\label{eq:Hessian_diag}
\Sigma_{kk} = \frac{1}{\alpha_k} \left(1 - \frac{2}{K}\right)  + \frac{1}{K^2} \sum_{l}^{k} \frac{1}{\alpha_l}.
\end{equation}
To invert this mapping we transform Equation \ref{eq:mu_k} to 
\begin{equation}
\label{eq:reform_mu_k}
\alpha_k = e^{\mu_k} \prod_l^{K} \alpha_l^{1/K}
\end{equation}
by applying the logarithm and re-ordering some parts. Inserting this into Equation \ref{eq:Hessian_diag} and re-arranging yields
\begin{equation}
\prod_l^K \alpha_l^{1/K} = \frac{1}{\vSigma_{kk}} \left[e^{-\mu}\left(1 - \frac{2}{K}\right)  + \frac{1}{K^2} \sum_u^K e^{-\mu_u} \right]
\end{equation}
which can be re-inserted into Equation \ref{eq:reform_mu_k} to give
\begin{equation}
\label{eq:mapping_alpha}
\alpha_k = \frac{1}{\Sigma_kk} \left(1 - \frac{2}{K} + \frac{e^{-\mu_k}}{K^2} \sum_l^K e^{-\mu_k} \right)
\end{equation}
which is the final mapping. With Equations \ref{eq:mu_k} and \ref{eq:Hessian_diag} we are able to map from Dirichlet to Gaussian and with Equation \ref{eq:mapping_alpha} we are able to map the inverse direction. 

\subsubsection*{The Bridge for the inverse-softmax transform}

In summary we get the following forward and backward transformations between $\vy \in \mathbb{R}^d$ and $\pi \in \mathbb{P}^d$.
\begin{subequations}
\begin{align}
	\mu_k &= \log \alpha_k  - \frac{1}{K} \sum_{l=1}^{K} \log \alpha_l \, , \label{eq:mubridge}\\
	\Sigma_{k\ell} &= \delta_{k\ell}\frac{1}{\alpha_k} - \frac{1}{K}\left[\frac{1}{\alpha_k} + \frac{1}{\alpha_\ell} - \frac{1}{K}\sum_{u=1} ^K \frac{1}{\alpha_u} \right].
	\label{eq:Sigmabridge} 
\end{align}
\end{subequations}
The corresponding derivations require care because the Gaussian parameter space is evidently larger than that of the Dirichlet and not fully identified by the transformation.
The pseudo-inverse of this map---as summarized above---was provided by \citet{KernelTopicModels2012}. It maps the Gaussian parameters to those of the Dirichlet as
\begin{equation} \label{eq:alpha_transform}
	\alpha_k = \frac{1}{\Sigma_{kk}}\left(1 - \frac{2}{K} + \frac{e^{\mu_k}}{K^2}\sum_{l=1}^K e^{-\mu_l} \right) \,
\end{equation}
\begin{figure}[!htb]
	\centering
	\includegraphics[width=\textwidth]{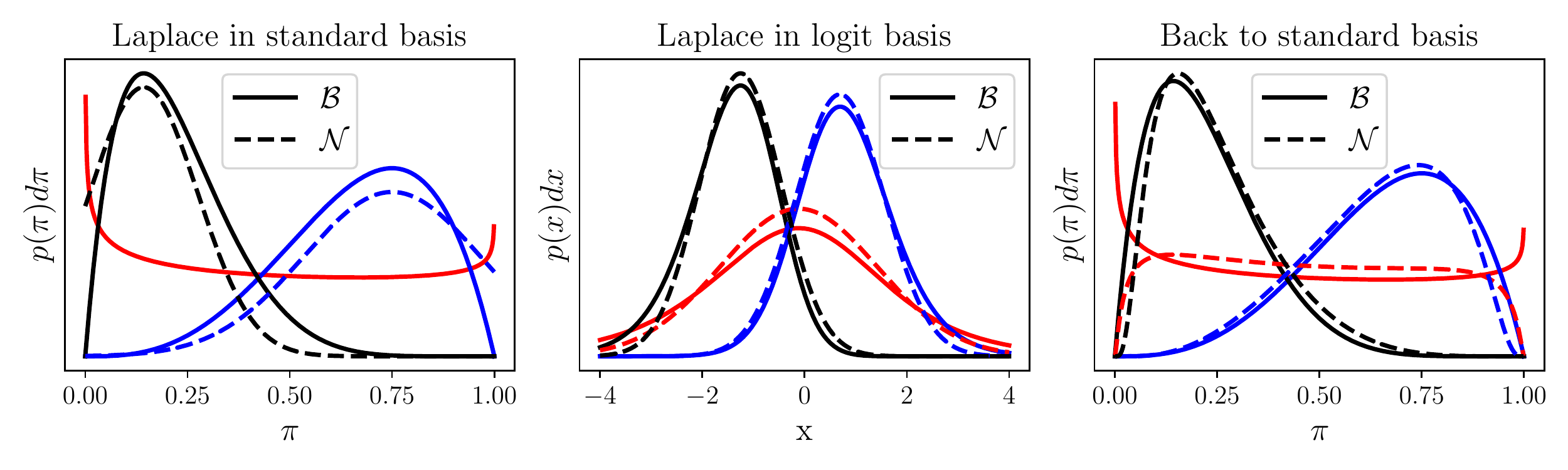}
	\caption{inverse-softmax Dirichlet bridge. We show the Beta distribution as it is the 1-dimensional Version of the Dirichlet.}
	\label{fig:dirichlet_softmax_bridge}
\end{figure}
\subsection{Wishart Distribution}
\label{subsec:wishart_dist}


\subsubsection{Background: Kronecker-product}

Kronecker-product: $A \otimes B \in \mathbb{R}^{(m_1m_2) \times (n_1n_2)}$ is defined by $(A \otimes B)_{(i - 1)m_2+j,(k - 1)n_2+l} = a_{ik}b_{jl} = (A \otimes B)_{(ij)(kl)}$.

Box-product: $A \boxtimes B \in \mathbb{R}^{(m_1m_2) \times (n_1n_2)}$ is defined by $(A \boxtimes B)_{(i - 1)m_2+j,(k - 1)n_1+l} = a_{il}b_{jk} = (A \boxtimes B)_{(ij)(kl)}$.

Symmetric Kronecker Product: $A \ostimes B = A \otimes B + A \boxtimes B + B \otimes A + B \boxtimes A$


\subsubsection{Standard Wishart distribution}

the pdf of the Wishart is
\begin{subequations}
\begin{align}
\mathcal{W}(X; n,p,V) &= \frac{1}{2^{np/2} \left|{\mathbf V}\right|^{n/2} \Gamma_p\left(\frac {n}{2}\right ) }{\left|\mathbf{X}\right|}^{(n-p-1)/2} e^{-(1/2)\operatorname{tr}({\mathbf V}^{-1}\mathbf{X})} \\
&= \exp \left[(n-p-1)/2 \log(|X|) -(1/2)\operatorname{tr}({\mathbf V}^{-1}\mathbf{X}) - \log\left(2^{np/2} \left|{\mathbf V}\right|^{n/2} \Gamma_p\left(\frac {n}{2}\right )\right) \right]\\
&= \frac{1}{X^{\frac{1}{2}}}\exp \left[(n-p)/2 \log(|X|) -(1/2)\operatorname{tr}({\mathbf V}^{-1}\mathbf{X}) - \log\left(2^{np/2} \left|{\mathbf V}\right|^{n/2} \Gamma_p\left(\frac {n}{2}\right )\right) \right]
\end{align}
\end{subequations}
with exponential family values $h(X) = 1/X^{\frac{1}{2}}$, $\phi(X)=(\log(X), X), w=((n-p)/2, V^{-1})$ and $Z(n,p,V)=\log\left(2^{np/2} \left|{\mathbf V}\right|^{n/2} \Gamma_p\left(\frac {n}{2}\right )\right)$.

\subsubsection*{Laplace Approximation of the standard Wishart distribution}

Using $\frac{\partial \det(X)}{\partial X} = \det(X)(X^{-1})^\top$ and $\frac{\partial}{\partial X} Tr(AX) = A^\top$ we can calculate the mode by setting the first derivative of the log-pdf to zero
\begin{align*}
\frac{\partial \log \mathcal{W}(X; n,p,V)}{\partial X} &= \frac{(n-p-1)\det(X)(X^{-\top})}{2\det(X)} - \frac{V^{-\top}}{2} \\
\Rightarrow 0 &= \frac{(n-p-1)X^{-1}}{2} - \frac{V^{-1}}{2} \\
\Leftrightarrow  \frac{(n-p-1)X^{-1}}{2} &= \frac{V^{-1}}{2} \\
\Leftrightarrow X &= (n-p-1)V
\end{align*}
Using the fact that the matrix is symmetric and $\frac{\partial X^{-1}}{\partial X} = -X^{-1} \otimes X^{-1}$ we get 
\begin{align*}
\frac{\partial^2 \log f(X; n,p,V)}{\partial^2 X} &= -\frac{(n-p-1)}{2} X^{-1} \otimes X^{-1}
\end{align*}
Using $(\alpha A)^{-1} = \alpha^{-1}A^{-1}$, the linearity of the Kronecker product to pull out scalars and $X^{-1} \otimes X^{-1} = (X \otimes X)^{-1}$ to insert the mode and invert we get:
\begin{align*}
-\frac{(n-p-1)}{2} X^{-1} \otimes X^{-1} &= -\frac{(n-p-1)}{2} \frac{1}{(n-p-1)} V^{-1} \otimes \frac{1}{(n-p-1)} V^{-1} \\
&= -\frac{1}{2(n-p-1)}(V \otimes V)^{-1} \\
\Rightarrow \Sigma &= 2(n-p-1)(V \otimes V)
\end{align*}
In summary, the Laplace approximation of a Wishart distribution in the standard basis is $\mathcal{N}\left(X; (n-p-1)V, 2(n-p-1)(V \otimes V)\right)$, where the representation of the symmetric positive definite matrices has been changed from $\mathbb{R}^{n\times n}$ to $\mathbb{R}^{n^2}$.

\subsubsection*{Laplace Approximation of the sqrtm-transformed Wishart distribution}

For the derivations we will introduce a symmetry constraint since the matrix-sqrt of a symmetric matrix is also symmetric. The constraint is $\frac{1}{2\beta}||Y-Y^\top||^2_{F}$ with $\beta \rightarrow \infty$ such that the constraint becomes a dirac delta in the limit.

Using $\frac{\partial \det(Y)}{\partial Y} = \det(Y)(Y^{-1})^\top$ and $\frac{\partial}{\partial Y} \operatorname{tr}(AY^\top Y) = YA^\top + YA$ we can calculate the mode by setting the first derivative of the log-pdf to zero
\begin{subequations}
\begin{align}
\frac{\partial \log \mathcal{W}_{\text{sqrtm}}(Y; n,p,V)}{\partial Y} &=
\frac{\partial}{\partial Y} (n-p) \log(|\mathbf{Y}|) - (1/2)\operatorname{tr}({\mathbf V}^{-1}\mathbf{Y^\top Y}) - C - \frac{1}{2\beta}||Y-Y^\top||^2_{F} \\
&= \frac{(n-p)\det(Y)(Y^{-\top})}{\det(Y)} - \frac{(YV^{-\top} + YV^{-1})}{2} - \frac{1}{\beta}(Y - Y^\top)\\
&= \frac{(n-p)\det(Y)(Y^{-\top})}{\det(Y)} - YV^{-1} - \frac{1}{\beta}(Y - Y^\top)\\
\Rightarrow 0 &= (n-p)Y^{-\top} - YV^{-1} \\
\Rightarrow (n-p)Y^{-1} &= YV^{-1} \\
\Rightarrow Y &= \operatorname{sqrtm}\left((n-p)V)\right)
\end{align}
\end{subequations}
Computing the second derivative by using $\frac{\partial}{\partial Y}Y^{-1} = -Y^{-\top} \otimes Y^{-1}$, $\frac{\partial (YA)_{kl}}{\partial Y_{ij}} = \delta_{ki} A_{jl} \Leftrightarrow \frac{\partial AY}{\partial Y}  = I \otimes A$. To get the covariance matrix we multiply with $-1$ and invert the matrix. 
\begin{subequations}
\begin{align}
&\frac{\partial^2 \log \mathcal{W}_{\text{sqrtm}}(Y; n,p,V)}{\partial^2 Y} = \frac{\partial}{\partial Y} \left[(n-p)Y^{-\top} - YV^{-1} - \frac{1}{\beta}(Y - Y^\top)\right]\\
&= -(n-p) (Y^{-\top} \otimes Y^{-1}) - I \otimes V^{-1} - \frac{1}{\beta}(I \otimes I - I\boxtimes I)\\
&\overset{\text{mode}}{\Rightarrow} -(n-p) \left[\sqrt{\frac{1}{(n-p)}}V^{-\frac{1}{2}} \otimes \sqrt{\frac{1}{(n-p)}}V^{-\frac{1}{2}}\right] -  I \otimes V^{-1} - \frac{1}{\beta}(I \otimes I - I\boxtimes I)\\
&= -\left(V^{-\frac{1}{2}} \otimes V^{-\frac{1}{2}} + I_p \otimes V^{-1} + \frac{1}{\beta}(I \otimes I - I\boxtimes I)\right) \\
\Leftrightarrow \frac{\partial^2 \log \mathcal{W}_{\text{sqrtm}}(Y; n,p,V)}{\partial Y_{ik} \partial Y_{jl}} &= -\left( V^{-\frac{1}{2}}_{ik} V^{-\frac{1}{2}}_{jl} + \delta_{ik} V^{-1}_{jl} + \frac{1}{\beta}(\delta_{ik} \delta_{jl} - \delta_{il} \delta_{jk})\right)
\end{align}
\end{subequations}
In general, any matrix can be split into a symmetric and a skew-symmetric matrix, $A = \frac{1}{2}(A + A^\top) + \frac{1}{2}(A - A^\top)$. For both parts we can define projection operators $\Gamma$ and $\Delta$ with
\begin{subequations}
\begin{align}
    \Gamma A &= \frac{1}{2}(A+A^\top), \Gamma_{(ij),(kl)} = \frac{1}{2}(\delta_{ik}\delta_{jl} + \delta_{il}\delta_{kj}) \\
    \Delta A &= \frac{1}{2}(A-A^\top), \Delta_{(ij),(kl)} = \frac{1}{2}(\delta_{ik}\delta_{jl} - \delta_{il}\delta_{kj}) \\   
\end{align}
\end{subequations}
For these projection operators it holds that $\Gamma \Gamma = \Gamma$, $\Delta \Delta = \Delta$, $\Delta \Gamma = \Gamma \Delta = \mathbf{0}$, and $\Delta + \Gamma = \Gamma + \Delta = \mathbf{I}$. Furthermore, it holds that
\begin{subequations}
\begin{align}
    A \otimes B &= (\Gamma + \Delta)(A \otimes B)(\Gamma + \Delta)^\top \\
    &= \Gamma(A \otimes B)\Gamma^\top + \Gamma(A \otimes B)\Delta^\top + \Delta(A \otimes B)\Gamma^\top + \Delta(A \otimes B)\Delta^\top
\end{align}
\end{subequations}

In the context of the Wishart this means
\begin{align}
    &\Gamma\left(V^{-\frac{1}{2}} \otimes V^{-\frac{1}{2}} + I_p \otimes V^{-1} + \frac{1}{\beta}\Delta\right)\Gamma^\top + \Gamma\left(V^{-\frac{1}{2}} \otimes V^{-\frac{1}{2}} + I_p \otimes V^{-1} + \frac{1}{\beta}\Delta\right)\Delta^\top\\ &+ \Delta\left(V^{-\frac{1}{2}} \otimes V^{-\frac{1}{2}} + I_p \otimes V^{-1} + \frac{1}{\beta}\Delta\right)\Gamma^\top + \Delta\left(V^{-\frac{1}{2}} \otimes V^{-\frac{1}{2}} + I_p \otimes V^{-1} + \frac{1}{\beta}\Delta\right)\Delta^\top \\
    &= \left(V^{-\frac{1}{2}} \ostimes V^{-\frac{1}{2}} + I_p \ostimes V^{-1}\right) + 0 + 0 + 0 
\end{align}
because all the terms involving $\Delta$ yield 0 probability mass on symmetric matrices due to the symmetry constraint. 

In practice, e.g. for the computation of the KL-divergence, the symmetric Kronecker product is suboptimal for computation because it is only invertible in $\mathbb{R}_S^d$ but not in $\mathbb{R}^d$. Thus we chose to compute the covariance matrix with 
\begin{align}
    \Sigma &= \left(V^{-\frac{1}{2}} \otimes V^{-\frac{1}{2}} + I_p \otimes V^{-1}\right)^{-1}
\end{align}
and adapt the pdf of the normal distribution to the space of symmetric matrices
\begin{align}
    \label{eq:gaussian_sym_adapted}
    \mathcal{N}(X, \mu, \Sigma) &= \frac{\frac{1}{2} d (d+1)}{d^2} (2 \pi)^{-\frac{2}{d}} \det(U^\top \Sigma U)^{-\frac{1}{2}} \exp{-1/2 (X - \mu)^\top \Sigma^{-1} (X - \mu)}
\end{align}
where $\frac{\frac{1}{2} d (d+1)}{d^2}$ is the ratio of degrees of freedom between a symmetric and asymmetric matrix, $U$ are all eigenvectors of the symmetry projection $\Gamma$ and $d$ is the number of dimensions.

We can simplify the Hessian further to make it computationally easier to invert it. 
\begin{subequations}
\begin{align}
    ... &= -\Gamma\left(V^{-\frac{1}{2}} \otimes V^{-\frac{1}{2}} + I_p \otimes V^{-1}\right)\Gamma^\top\\
    &= -\left(V^{-\frac{1}{2}} \ostimes V^{-\frac{1}{2}} + I_p \ostimes V^{-1}\right) \\
    &\overset{\cdot -1}{=} \left(I_{p^2} + V^{\frac{1}{2}} \ostimes V^{-\frac{1}{2}}\right)\left(V^{-\frac{1}{2}} \ostimes V^{-\frac{1}{2}}\right)\\
    \Rightarrow \Sigma &= \left(V^{\frac{1}{2}}\ostimes V^{\frac{1}{2}}\right) \left(I_{p^2} + V^{\frac{1}{2}} \ostimes V^{-\frac{1}{2}}\right)^{-1}
\end{align}
\end{subequations}
which could be inverted more easily using equation 5 of \cite{NIPS2011_EfficientMatrix}.
However, if you aren't interested in the covariance matrix itself, but e.g. only want to compute the Gaussian, it makes sense to compute everything with the inverse covariance matrix to save the computational effort of an inversion. 

\subsubsection*{The Bridge for sqrtm-tranform}

We use $\mu =  ((n-p)V)^{\frac{1}{2}} \Leftrightarrow \mu^2 = (n-p)V \Leftrightarrow V = \frac{\mu^2}{(n-p)}$. Remember that $\mu$ is reshaped to be the same size as $V$ even though we usually think of it in vector-form. 

\begin{subequations}
\begin{align}
	\mu &=  ((n-p)V)^{\frac{1}{2}} \\
	\Sigma &= \left(V^{\frac{1}{2}}\ostimes V^{\frac{1}{2}}\right) \left(I_{p^2} + V^{\frac{1}{2}} \ostimes V^{-\frac{1}{2}}\right)^{-1} \\
	V &= ** \\
	n &= **
\end{align}
\end{subequations}

The resulting $\Sigma$ cannot be easily solved for $V$ and thus there are three ways to choose a matching from $\mu, \Sigma$ to $n, V$. a) We can assume that $\Sigma$ has to have the same structure as shown above, i.e. a product of Kronecker products. Then we can compute $V$ and insert it in the equation for $\mu$ to get $n$. b) We can treat $V$ or $n$ as a free parameter and compute our solution solely from the equation of $\mu$. c) We could just use the logm-transform which is has good inversions for both $n$ and $V$. 
\begin{figure}[!htb]
	\centering
	\includegraphics[width=\textwidth]{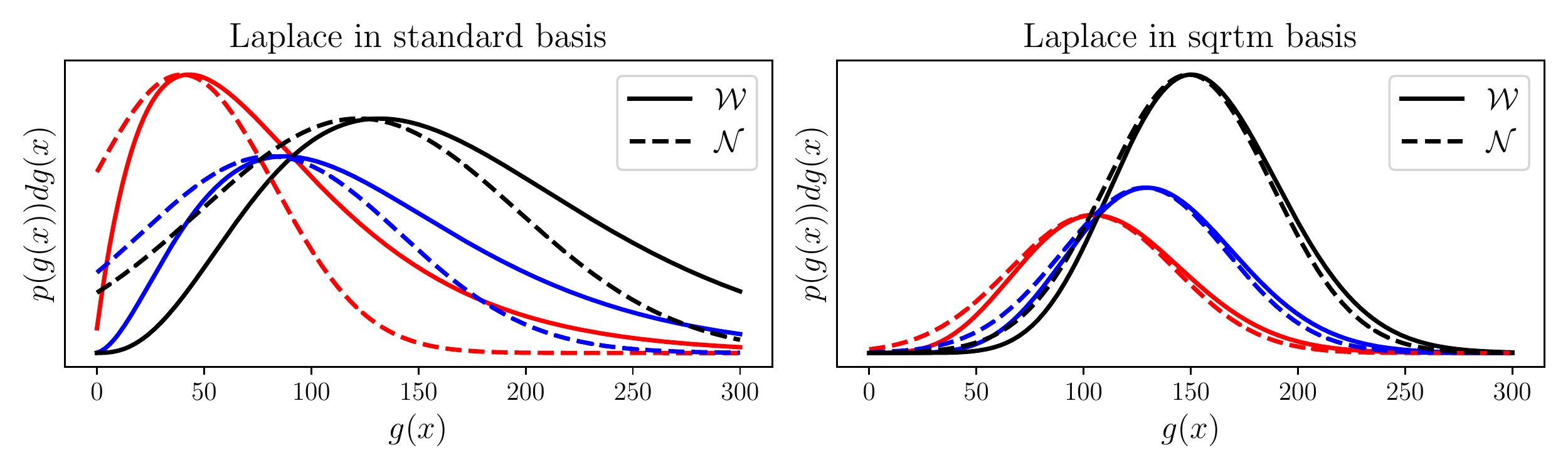}
	\caption{sqrtm-bridge for the Wishart distribution.}
	\label{fig:wishart_sqrtm_bridge}
\end{figure}

\subsubsection{Logm-Transformed Wishart distribution}

We transform the distribution with $g(X) = \text{logm}(X)$, i.e. $X(Y) = g^{-1}(X) = \text{expm}(Y)$, where $\text{expm}(Y)$ is the matrix exponential and $\text{logm}(Y)$ is the matrix logarithm of $Y$. The new pdf becomes
\begin{subequations}
\begin{align}
\mathcal{W}_{\text{logm}}(Y; n, p, V) &= \frac{1}{2^{np/2} \left|{\mathbf V}\right|^{n/2} \Gamma_p\left(\frac {n}{2}\right ) }{\left|\mathbf{\operatorname{expm}Y}\right|}^{(n-p-1)/2} e^{-(1/2)\operatorname{tr}({\mathbf V}^{-1}\mathbf{\operatorname{expm}Y})} \cdot |\operatorname{expm}Y| \\ 
&= \frac{1}{2^{np/2} \left|{\mathbf V}\right|^{n/2} \Gamma_p\left(\frac {n}{2}\right ) }{\left|\mathbf{\operatorname{expm}Y}\right|}^{(n-p+1)/2} e^{-(1/2)\operatorname{tr}({\mathbf V}^{-1}\mathbf{\operatorname{expm}Y})} \\ 
&= \exp \left[C + \frac{(n-p+1)}{2} \log(\left|\mathbf{\operatorname{expm}Y}\right|)  - \frac{1}{2}\operatorname{tr}({\mathbf V}^{-1}\mathbf{\operatorname{expm}Y}) \right]
\end{align}
\end{subequations}
with exponential family values $h(Y) = Y^{\frac{1}{2}}, \phi(Y) = (Y, \operatorname{expm}Y), w = (n-p), V^{-1}$ and $Z(n,p,V)=\log\left(2^{np/2} \left|{\mathbf V}\right|^{n/2} \Gamma_p\left(\frac {n}{2}\right )\right)$.

\subsubsection*{Laplace Approximation of the logm-transformed Wishart distribution}

The matrix logarithm of a symmetric matrix yields a symmetric matrix again. Thus the symmetry-constraint of the matrix-sqrt transformation also applies here. For brevity we have decided not to include it in the derivation and point it out at selected parts of the text.

To compute the first derivative we use the following
\begin{subequations}
\begin{align}
	\frac{\partial \log(\det(\operatorname{expm}(Y)))}{\partial Y} 
	&= \frac{\partial \log(\det(\operatorname{expm}(Y)))}{\partial \det(\operatorname{expm}(Y))} \cdot \frac{\partial \det(\operatorname{expm}(Y))}{\partial \operatorname{expm}(Y)} \cdot \frac{\partial \operatorname{expm}(Y)}{\partial Y} \\
	&= \frac{1}{\det(\operatorname{expm}(Y))} \cdot \det(\operatorname{expm}(Y)) \operatorname{expm}(Y)^{-\top} \cdot \operatorname{expm}(Y) \\
	&= I_p
\end{align}
\end{subequations}
where $I_p$ is the identity matrix of size $p$ and we use the fact that the matrix logarithm of a symmetric matrix is symmetric, implying $\operatorname{expm}(Y)^{-\top} = \operatorname{expm}(Y)^{-1}$. With this we get the first derivative
\begin{subequations}
\begin{align}
	\frac{\partial \log W_{log}}{\partial Y} &= \frac{\partial}{\partial Y} \left[C + \frac{(n-p+1)}{2} \log(\left|\mathbf{\operatorname{expm}Y}\right|)  - \frac{1}{2}\operatorname{tr}({\mathbf V}^{-1}\mathbf{\operatorname{expm}Y}) \right] \\
	&=  \frac{(n-p+1)}{2} I_p - \frac{1}{2}V^{-1}\mathbf{\operatorname{expm}Y} 
\end{align}
\end{subequations}
By setting this to zero we get a mode of
\begin{subequations}
\begin{align}
	0 &=  \frac{(n-p+1)}{2} I_p - \frac{1}{2}V^{-1}\mathbf{\operatorname{expm}Y} \\
	\Leftrightarrow (n-p+1)I_p &= V^{-1}\mathbf{\operatorname{expm}Y} \\
	\Leftrightarrow Y &= \operatorname{logm}((n-p+1)V)
\end{align}
\end{subequations}
For the second derivative we use the fact that
\begin{subequations}
\begin{align}
	\frac{\partial (B\operatorname{expm}(Y))_{kl}}{\partial Y_{ij}} &= \delta_{jl}(B\operatorname{exmp}(Y))_{ki} \\
	\Leftrightarrow \frac{\partial B\operatorname{expm}(Y)}{\partial Y} &= \left(B\operatorname{expm}(Y) \otimes I_p\right)
\end{align}
\end{subequations}
yielding
\begin{subequations}
\begin{align}
	\frac{\partial^2 \log \mathcal{W}_{\text{logm}}}{\partial^2 Y} &= \frac{\partial \log \mathcal{W}_{\text{logm}}}{\partial Y} \left[ \frac{(n-p+1)}{2} I_p - \frac{1}{2}V^{-1}\mathbf{\operatorname{expm}Y} \right] \\
	&= -\frac{1}{2}(V^{-1}\mathbf{\operatorname{expm}Y} \otimes I_p) \\
	&\overset{\text{mode}}{\Rightarrow} -\frac{1}{2}((n-p+1)V^{-1}V \otimes I_p) \\
	&= -\frac{(n-p+1)}{2} (I_p \otimes I_p)\\
	\Leftrightarrow \Sigma &= \frac{2}{n-p+1} I_{p^2}
\end{align}
\end{subequations}
where $I_{p^2}$ is an Identity matrix of size $p^2$. With the symmetry constraint, we would get $(I_p \ostimes I_p)$ instead of $(I_p \otimes I_p)$.

\subsubsection*{The Bridge for logm-tranform}

$\mu$ and $\Sigma$ are already given by the Laplace approximation. Inverting the mode yields an estimate for $V$. 
\begin{align}
	\mu &= \operatorname{logm}((n-p+1)V) \Leftrightarrow \operatorname{expm}(\mu) = (n-p+1)V \Leftrightarrow V = \frac{\operatorname{expm}(\mu)}{(n-p+1)}
\end{align}
where $\mu$ and $V$ are reshaped to a matrix of size $p\times p$. The parameter $n$ can be derived from the equation of $\Sigma$ by
\begin{align}
    \Sigma &= \frac{2}{n-p+1} I_{p^2} \\
    \Leftrightarrow \operatorname{tr}(\Sigma) &= \frac{2p}{n-p+1}p^2 \\
    \Leftrightarrow n &= \frac{2p^2}{\operatorname{tr}(\Sigma)} + p - 1
\end{align}
In summary this yields
\begin{subequations}
\begin{align}
	\mu &= \operatorname{logm}((n-p+1)V) \\
	\Sigma &= \frac{2}{n-p+1} (I_p \otimes I_p)^{-1} \\
	V &= \frac{\operatorname{expm}(\mu)}{(n-p+1)} \\
	n &= \frac{2p^2}{\operatorname{tr}(\Sigma)} + p - 1
\end{align}
\end{subequations}
Similar to the sqrtm-transformation, we can adapt the normal distribution to the space of symmetric matrices only and compute with the unconstrained version (see previous subsection).
\begin{figure}[!htb]
	\centering
	\includegraphics[width=\textwidth]{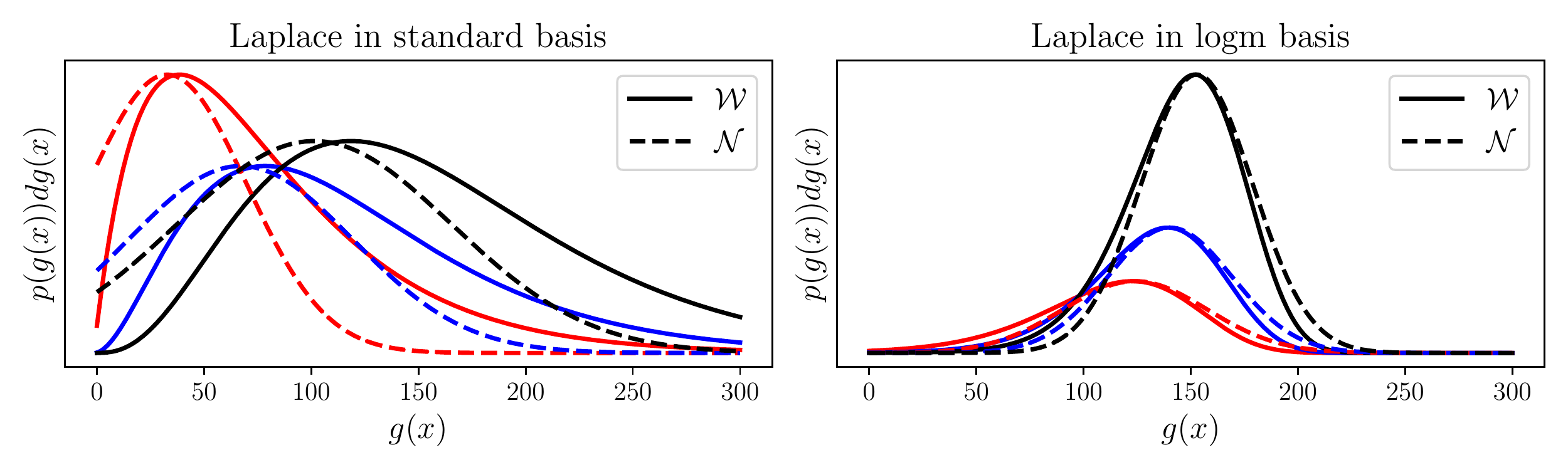}
	\caption{logm-bridge for the Wishart distribution}
	\label{fig:wishart_logm_bridge}
\end{figure}

\subsection{Inverse Wishart Distribution}
\label{subsec:inv_wishart_dist}

\subsubsection{Standard Inverse Wishart distribution}

The pdf of the inverse Wishart is
\begin{subequations}
\begin{align}
\mathcal{IW}_{\mathbf x}({\mathbf x}; {\mathbf \Psi}, \nu) &= \frac{\left|{\mathbf\Psi}\right|^{\nu/2}}{2^{\nu p/2}\Gamma_p(\frac \nu 2)} \left|\mathbf{x}\right|^{-(\nu+p+1)/2} e^{-\frac{1}{2}\operatorname{tr}(\mathbf\Psi\mathbf{x}^{-1})}
\label{eq:inverse_wishart_pdf} \\
&= \exp \left[-(\nu + p + 1)/2 \log(|x|) - \frac{1}{2} \text{tr}(\Psi x^{-1}) + \log(\frac{\left|{\mathbf\Psi}\right|^{\nu/2}}{2^{\nu p/2}\Gamma_p(\frac \nu 2)}) \right]
\end{align}
\end{subequations}
with exponential family values $h(X) = 1/X^\frac{1}{2}, \phi(X)=(\log(X), X^{-1}), w=(-(\nu+p)/2, \Psi)$ and $Z(n,p,V)=- \log(\frac{\left|{\mathbf\Psi}\right|^{\nu/2}}{2^{\nu p/2}\Gamma_p(\frac \nu 2)})$.

\subsubsection*{Laplace Approximation of the standard inverse Wishart distribution}

Using $\frac{\partial \det(X)}{\partial X} = \det(X)(X^{-1})^\top$ and $\frac{\partial}{\partial X} \operatorname{tr}(AX)^{-1} = (X^{-1}AX^{-1})^\top$ we can calculate the mode by setting the first derivative of the log-pdf to zero:
\begin{subequations}
\begin{align}
	\frac{\partial \log \mathcal{IW}_{\mathbf X}({\mathbf X}; {\mathbf \Psi}, \nu)}{\partial X} &= \frac{-(\nu + p + 1) \det(X) X^{-\top}}{2\det(X)} + \frac{(X^{-1} \Psi X^{-1})^\top}{2} \\
	&=\frac{-(\nu + p + 1) X^{-\top}}{2} + \frac{(X^{-1} \Psi X^{-1})^\top}{2} \\
	\Rightarrow 0 &=\frac{-(\nu + p + 1) X^{-\top}}{2} + \frac{(X^{-1} \Psi X^{-1})^\top}{2} \\
	\Leftrightarrow (\nu + p + 1) X^{-1} &= X^{-1} \Psi X^{-1} \\
	\Leftrightarrow (\nu + p + 1) &= X^{-1} \Psi\\
	\Leftrightarrow X &= \frac{1}{\nu + p + 1} \Psi
\end{align}
\end{subequations}
Using 
\begin{subequations}
\begin{align}
	\frac{\partial(XBX)_{kl}}{\partial X_{ij}} &= \delta_{ki}(BX)_{lj} + \delta_{lj}(XB)_{ki} \\
	\frac{\partial X^{-1}}{\partial X} &= -(X^{-1} \otimes X^{-1}) \\
	\frac{\partial (X^{-1}B X^{-1})}{\partial X} &= \frac{\partial (X^{-1}B X^{-1})}{\partial X^{-1}} \frac{\partial X^{-1}}{\partial X} = -[\delta_{ki}(BX^{-1})_{lj} + \delta_{lj}(X^{-1}B)_{ki}] (X^{-1} \otimes X^{-1})\\
	(AB)^{-1} &= B^{-1} A^{-1}
\end{align}
\end{subequations}
we can get the covariance matrix by inverting the Hessian and multiplying with -1. 
\begin{subequations}
\begin{align}
		&\frac{\partial^2 \log f_{\mathbf X}({\mathbf X}; {\mathbf \Psi}, \nu)}{\partial^2 X} = \frac{(\nu + p + 1)}{2}(X^{-1} \otimes X^{-1})^\top - \frac{[\delta_{ki}(\Psi X^{-1})_{lj} + \delta_{lj}(X^{-1}\Psi)_{ki}]}{2} (X^{-1} \otimes X^{-1})^\top \\
		&= \left\{\frac{(\nu + p + 1)}{2} I_{n^2}- \frac{[\delta_{ki}(\Psi X^{-1})_{lj} + \delta_{lj}(X^{-1}\Psi)_{ki}]}{2}\right\} (X^{-1} \otimes X^{-1})^\top \\
		&\overset{\text{mode}}{=} \left\{\frac{(\nu + p + 1)}{2} I_{n^2}- \frac{[\delta_{ki}((\nu + p + 1)\Psi \Psi^{-1})_{lj} + \delta_{lj}((\nu + p + 1)\Psi^{-1}\Psi)_{ki}]}{2}\right\} (\nu + p + 1)^2(\Psi \otimes \Psi)^{-\top} \\
		&= \left\{\frac{(\nu + p + 1)}{2} I_{n^2}- \underbrace{\frac{[\delta_{ki}((\nu + p + 1)I_n)_{lj} + \delta_{lj}((\nu + p + 1)I_n)_{ki}]}{2}}_{=(\nu + p + 1)I_{n^2}}\right\} (\nu + p + 1)^2(\Psi \otimes \Psi)^{-\top} \\
		&= -\underbrace{\frac{1}{2}(\nu + p + 1) I_{n^2}}_{A} \underbrace{(\nu + p + 1)^2(\Psi \otimes \Psi)^{-\top}}_{B} \\
		&\overset{\text{invert}}{\Rightarrow} -\frac{1}{(\nu + p + 1)^2}(\Psi \otimes \Psi)^{\top} \frac{2}{(\nu + p + 1)}I_{n^2} \\
		&\overset{\cdot -1}{\Rightarrow} \Sigma = \frac{2}{(\nu + p + 1)^3}(\Psi \otimes \Psi)^\top
\end{align}
\end{subequations}
where $I_{n}$ and $I_n^2$ are the identity matrix of size $n$ and $n^2$ respectively. We can also ignore the transpose since we are dealing with symmetric positive definite matrices when it comes to the inverse Wishart distribution. This yields a multivariate Gaussian of the form $\mathcal{N}\left(X; \mu = \frac{1}{\nu + p + 1} \Psi, \Sigma = \frac{2}{(\nu + p + 1)^3}(\Psi \otimes \Psi)\right)$. 

\subsubsection{Logm-Transformed inverse Wishart distribution}

Similar to the Wishart derivation, we compute all derivations with a symmetry-constraint. For brevity we omit the constraint but point it out where it makes a difference. 

We transform the distribution with $g(X) = \text{logm}(X)$, i.e. $X(Y) = g^{-1}(X) = \text{expm}(Y)$, where $\operatorname{expm}(Y)$ is the matrix exponential. The new pdf becomes 
\begin{subequations}
\begin{align}
	\mathcal{IW}_{\operatorname{logm}}({\mathbf Y}; {\mathbf \Psi}, \nu) &= \frac{\left|{\mathbf\Psi}\right|^{\nu/2}}{2^{\nu p/2}\Gamma_p(\frac \nu 2)} \left|\operatorname{expm}\mathbf{Y}\right|^{-(\nu+p+1)/2} e^{-\frac{1}{2}\operatorname{tr}(\mathbf\Psi(\operatorname{expm}\mathbf{Y})^{-1})} \cdot |\operatorname{expm}(\mathbf{Y})|\\
	&=  \frac{\left|{\mathbf\Psi}\right|^{\nu/2}}{2^{\nu p/2}\Gamma_p(\frac \nu 2)} \left|\operatorname{expm}\mathbf{Y}\right|^{-(\nu+p-1)/2} e^{-\frac{1}{2}\operatorname{tr}(\mathbf\Psi(\operatorname{expm}\mathbf{Y})^{-1})} \\
	&= \exp\left[C - (\nu+p-1)/2\log  \left|\operatorname{expm}\mathbf{Y}\right| - -\frac{1}{2}\operatorname{tr}(\mathbf\Psi\operatorname{expm}(\mathbf{-Y}))\right]
\end{align}
\end{subequations}
with exponential family values $h(Y) = Y^\frac{1}{2}, \phi(Y) = (\log(\det(\operatorname{expm}(Y))), \operatorname{expm}(Y)), w = ((\nu + p)/2, \Psi)$ and $Z(n,p,V)=- \log\left(\frac{\left|{\mathbf\Psi}\right|^{\nu/2}}{2^{\nu p/2}\Gamma_p(\frac \nu 2)}\right)$

\subsubsection*{Laplace Approximation of the logm-transformed inverse Wishart distribution}

For the first derivative we use the same concepts as for the Wishart in Subsection \ref{subsec:wishart_dist}. 
\begin{subequations}
\begin{align}
	\frac{\partial \log \mathcal{IW}_{logm}}{\partial Y} &= \frac{\partial}{\partial Y} -(\nu+p-1)/2\log  \left|\operatorname{expm}\mathbf{Y}\right| - -\frac{1}{2}\operatorname{tr}(\mathbf\Psi\operatorname{expm}(\mathbf{-Y}))\\
	&= -\frac{(\nu+p-1)}{2}I_p + \frac{1}{2}\mathbf\Psi\operatorname{expm}(\mathbf{-Y})
\end{align}
\end{subequations}
which yields the mode by setting it to zero and solving for $Y$.
\begin{subequations}
\begin{align}
	0 &= -\frac{(\nu+p-1)}{2}I_p     + \frac{1}{2}\mathbf\Psi\operatorname{expm}(\mathbf{-Y}) \\
	\Leftrightarrow (\nu+p-1) I_p &=  \mathbf\Psi(\operatorname{expm}(\mathbf{Y}))^{-1} \\
	\Leftrightarrow Y &= \operatorname{logm}\left(\frac{\mathbf\Psi}{\nu+p-1}\right)
\end{align}
\end{subequations}
For the second derivative we also use the same concepts as for the Wishart and get
\begin{subequations}
\begin{align}
	\frac{\partial^2 \log \mathcal{IW}_{logm}}{\partial^2 Y} &= \frac{\partial }{\partial Y} \left[-\frac{(\nu+p-1)}{2} + \frac{1}{2}\mathbf\Psi\operatorname{expm}(\mathbf{-Y}) \right] \\
	&= -\frac{1}{2}\left[\mathbf\Psi\operatorname{expm}(\mathbf{Y})^{-1} \otimes I_p\right] \\
	&\overset{\text{mode}}{\Rightarrow} -\frac{1}{2}\left[\mathbf\Psi\frac{\mathbf\Psi^{-1}}{(\nu+p-1)} \otimes I_p\right] \\
	&= -\frac{1}{2(\nu+p-1)} I_{p\times p} \\
	\Rightarrow \Sigma &= 2(\nu+p-1) I_{p\times p}
\end{align}
\end{subequations}
With the symmetry-constraint $I_{p\times p}$ becomes $(I \ostimes I)^{-1}$. 
This yields a multivariate Normal distribution $\mathcal{N}\left(Y; \mu = \operatorname{logm}\left(\frac{\mathbf\Psi}{n+p-1}\right), \Sigma = 2(\nu+p-1) I_{p\times p}\right)$ which is constraint to symmetric matrices and its normalization is thus adapted as described in Equation \ref{eq:gaussian_sym_adapted}. 

\subsubsection*{The Bridge for the logm-transformed inverse Wishart distribution}

We get $\mu$ and $\Psi$ from the Laplace approximation and determine $V$ by inverting $\mu$
\begin{subequations}
\begin{align}
	\mu = \operatorname{logm}\left(\frac{\mathbf\Psi}{n+p-1}\right) \Leftrightarrow \operatorname{expm}(\mu) = \frac{\mathbf\Psi}{n+p-1} \Leftrightarrow \mathbf\Psi = \operatorname{expm}(\mu)(n+p-1)
\end{align}
\end{subequations}
Additionally we can get $\nu$ from the equation for $\Sigma$ by
\begin{align}
    \Sigma &= 2(\nu+p-1) I_{p\times p} \\
    \Leftrightarrow \operatorname{tr}(\Sigma) &= 2(\nu + p -1)p^2 \\
    \Leftrightarrow \nu &= \frac{\operatorname{tr}(\Sigma)}{2p^2} - p + 1
\end{align}
In summary we have
\begin{subequations}
\begin{align}
	\mu &= \operatorname{logm}\left(\frac{\mathbf\Psi}{n+p-1}\right) \\
	\Sigma &= 2(\nu+p-1) I_{p\times p} \\
	\mathbf{\Psi} &=  (\nu+p-1)\operatorname{expm}(\mu) \\
	\nu &= \frac{\operatorname{tr}(\Sigma)}{2p^2} - p + 1
\end{align}
\end{subequations}
where $\mathbf{\Psi}$ and $\mu$ are reshaped to a $p \times p$ matrix. For this derivation we assume to use only symmetric matrices. If this constraint is lifted, we have to replace $I_{p\times p}$ with $I_p \ostimes I_p$. 
\begin{figure}[!htb]
	\centering
	\includegraphics[width=\textwidth]{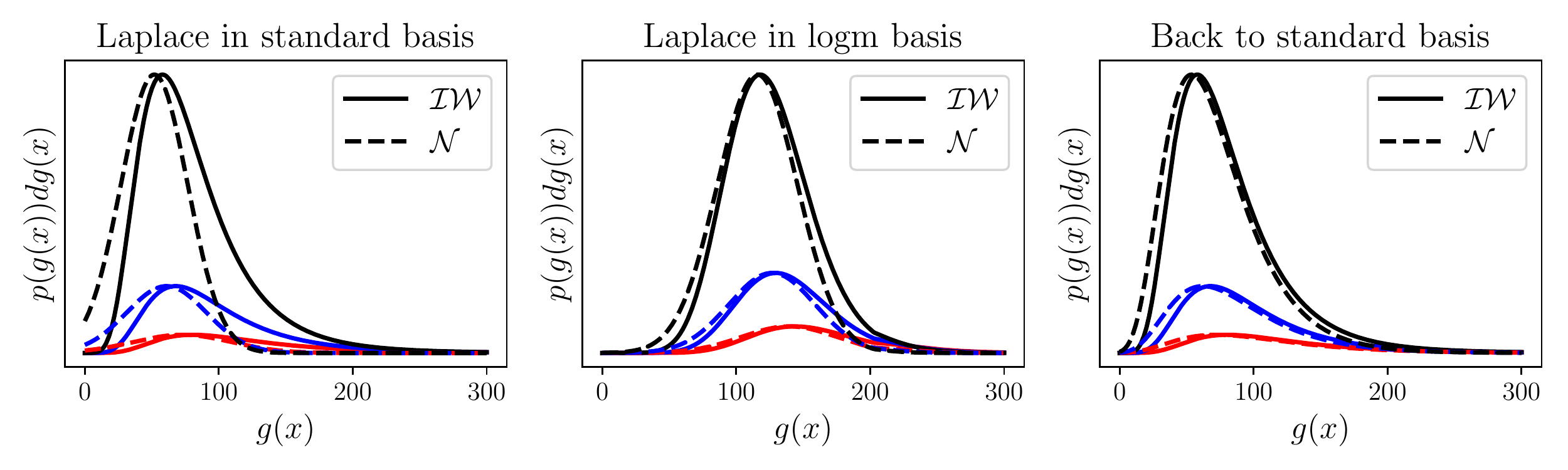}
	\caption{logm-bridge for the inverse Wishart distribution.}
	\label{fig:inverse_wishart_logm_bridge}
\end{figure}

\subsubsection{Sqrtm-Transformed inverse Wishart distribution}

Similar to the derivation of the sqrtm-transformation for the Wishart (see Subsection \ref{subsec:wishart_dist}, the inverse Wishart also has a symmetry constraint. For brevity, we omit it for the derivation and point out where it makes a difference. 

We transform the distribution with $g(X) = \text{sqrtm}(X) = X^{\frac{1}{2}}$, i.e. $X(Y) = g^{-1}(X) = Y^2$, where $\text{sqrtm}(Y)$ is the square root of the matrix. We choose the principle square root of the positive definite matrix $X$, i.e. $X = Y^\top Y = YY$. The new pdf becomes 
\begin{subequations}
\begin{align}
	\mathcal{IW}_{\operatorname{sqrtm}}({\mathbf Y}; {\mathbf \Psi}, \nu) &= \frac{\left|{\mathbf\Psi}\right|^{\nu/2}}{2^{\nu p/2}\Gamma_p(\frac \nu 2)} \left|\mathbf{Y^2}\right|^{-(\nu+p+1)/2} e^{-\frac{1}{2}\operatorname{tr}(\mathbf\Psi\mathbf{Y^\top Y}^{-1})} |2Y| \\
	&= \frac{\left|{\mathbf\Psi}\right|^{\nu/2}}{2^{\nu p/2}\Gamma_p(\frac \nu 2)} \left|\mathbf{Y}\right|^{-(\nu+p+1)} e^{-\frac{1}{2}\operatorname{tr}(\mathbf\Psi\mathbf{Y^\top Y}^{-1})} 2^p|Y| \\
	&= \frac{\left|{\mathbf\Psi}\right|^{\nu/2}}{2^{\nu p/2}\Gamma_p(\frac \nu 2)} \left|\mathbf{Y}\right|^{-(\nu+p)} e^{-\frac{1}{2}\operatorname{tr}(\mathbf\Psi\mathbf{Y}^{-2})} 2^p \\ 
	&= \exp\left[-(\nu + p) \log(|Y|) - \frac{1}{2}\text{tr}(\Psi (Y^\top Y)^{-1}) + \log(C)\right]
\end{align}
\end{subequations}
with $h(Y) = 1, \phi(Y) = (\log(|Y|), Y^{-2}), w = (-(\nu + p), \Psi)$ and $Z(w) = \log\left( \frac{\left|{\mathbf\Psi}\right|^{\nu/2}}{2^{\nu p/2}\Gamma_p(\frac \nu 2)}\right)$. 

\subsubsection*{Laplace Approximation of the sqrtm-transformed inverse Wishart distribution}

Using 
\begin{subequations}
\begin{align}
    \frac{\partial \det(Y)}{\partial Y} = \det(Y)(Y^{-1})^\top
	\frac{\partial \operatorname{tr}(\Psi (Y^\top Y)^{-1})}{\partial Y} &= 2\Psi Y^{-3}
\end{align}
\end{subequations}
we can calculate the mode by setting the derivative of the log-pdf to zero:
\begin{subequations}
\begin{align}
	\frac{\partial \log \mathcal{IW}_{\text{sqrtm}}(Y, \Psi, \nu)}{\partial Y} &= \frac{-(\nu + p )\det(Y) Y^{-\top}}{\det(Y)} + 2\Psi Y^{-3} \\
	&= -(\nu + p)Y^{-\top} + 2\Psi Y^{-3} \\
	\Rightarrow 0 &= -(\nu + p)Y^{-\top} + 2\Psi Y^{-3} \\
	\Leftrightarrow  (\nu + p)Y^{-\top} &= 2\Psi Y^{-3} \\
	\Leftrightarrow Y &= \operatorname{sqrtm}\left(\frac{1}{(\nu+p)} \Psi\right)
\end{align}
\end{subequations}
Using
\begin{subequations}
\begin{align}
	\frac{\partial X^{-1}}{\partial X} &= -X^{-1} \otimes X^{-1} \\
	\frac{\partial YX^{-3}}{\partial X} &= \frac{\partial \Psi X^{-3}}{\partial X^{-3}} \frac{\partial X^{-3}}{\partial X^{-1}} \frac{\partial X^{-1}}{\partial X} \\
	&= -\left(\Psi \otimes I\right) \left(I \otimes X^{-2} + X^{-1} \otimes X^{-1} + X^{-2} \otimes I\right)\left(X^{-1} \otimes X^{-1}\right)
\end{align}
\end{subequations}
We can calculate the Hessian and by multiplying with $-1$ and inverting it we get the covariance matrix.
\begin{subequations}
\begin{align}
	&\frac{\partial^2 \log \mathcal{IW}_{\text{sqrtm}}(Y, \Psi, \nu)}{\partial^2 Y} = \frac{\partial}{\partial Y} -(\nu + p)Y^{-\top} + \Psi Y^3 \\ 
	&= (\nu + p)(Y^{-1} \otimes Y^{-1}) - \left(\Psi \otimes I\right) \left(I \otimes Y^{-2} + Y^{-1} \otimes Y^{-1} + Y^{-2} \otimes I\right)\left(Y^{-1} \otimes Y^{-1}\right) \\
	&\overset{\text{mode}}{=} (\nu + p)(\sqrt{(\nu + p)}\Psi^{-\frac{1}{2}} \otimes \sqrt{(\nu + p)}\Psi^{-\frac{1}{2}}) \\
	&- \left(\Psi \otimes I\right) \left(I \otimes (\nu + p)\Psi^{-1} + \sqrt{(\nu + p)}\Psi^{-\frac{1}{2}} \otimes \sqrt{(\nu + p)}\Psi^{-\frac{1}{2}} + (\nu + p)\Psi^{-1} \otimes I\right)\\
	&\cdot \left(\sqrt{(\nu + p)}\Psi^{-\frac{1}{2}} \otimes \sqrt{(\nu + p)}\Psi^{-\frac{1}{2}}\right) \\
	&= (\nu + p)^2 \left(\Psi^{-\frac{1}{2}} \otimes \Psi^{-\frac{1}{2}}\right) - (\nu + p)^2 \left( \Psi \otimes \Psi^{-1} + \Psi^{\frac{1}{2}} \otimes \Psi^{-\frac{1}{2}} + I_{p^2} \right) \left(\Psi^{-\frac{1}{2}} \otimes \Psi^{-\frac{1}{2}}\right) \\
	&= -(\nu + p)^2  \left(\Psi \otimes \Psi^{-1} + \Psi^{\frac{1}{2}} \otimes \Psi^{-\frac{1}{2}} \right)\left(\Psi^{-\frac{1}{2}} \otimes \Psi^{-\frac{1}{2}} \right) \\
	&= -(\nu + p)^2  \left(\Psi^{\frac{1}{2}} \otimes \Psi^{-\frac{1}{2}} + I_{p^2}\right) \left(I_p \otimes \Psi^{-1}\right) \\
	\Leftrightarrow \Sigma &= \frac{1}{(\nu + p)^2} \left(I_p \otimes \Psi\right)\left(\Psi^{\frac{1}{2}} \otimes \Psi^{\frac{1}{2}} + I_p\right)^{-1}
\end{align}
\end{subequations}
which could be inverted more easily using equation 5 of \citet{NIPS2011_EfficientMatrix}. For further notes on efficient computation see Equation \ref{eq:gaussian_sym_adapted} and surrounding text. With the symmetry constraints, all $\otimes$ become $\ostimes$. 

\subsubsection*{The Bridge for the sqrtm-transformed inverse Wishart distribution}

The resulting $\Sigma$ cannot be easily solved for $\Psi$ and thus there are three ways to choose a matching from $\mu, \Sigma$ to $\nu, \Psi$. a) We can assume that $\Sigma$ has to have the same structure as shown above, i.e. a product of Kronecker products. Then we can compute $\Psi$ and insert it in the equation for $\mu$ to get $\nu$. b) We can treat $\Psi$ or $\nu$ as a free parameter and compute our solution solely from the equation of $\mu$. c) We could just use the logm-transform which is has good inversions for both $\nu$ and $\Psi$. 

In summary we have
\begin{align}
    \mu &= \operatorname{sqrtm}\left(\frac{1}{(\nu+p)} \Psi\right) \\
    \Sigma &= \frac{1}{(\nu + p)^2} \left(I_p \ostimes \Psi\right)\left(\Psi^{\frac{1}{2}}  \ostimes \Psi^{\frac{1}{2}} + I_p\right)^{-1} \\
    \Psi &= ** \\
    n &= **
\end{align}
where $**$ is described above.
\begin{figure}[!htb]
	\centering
	\includegraphics[width=\textwidth]{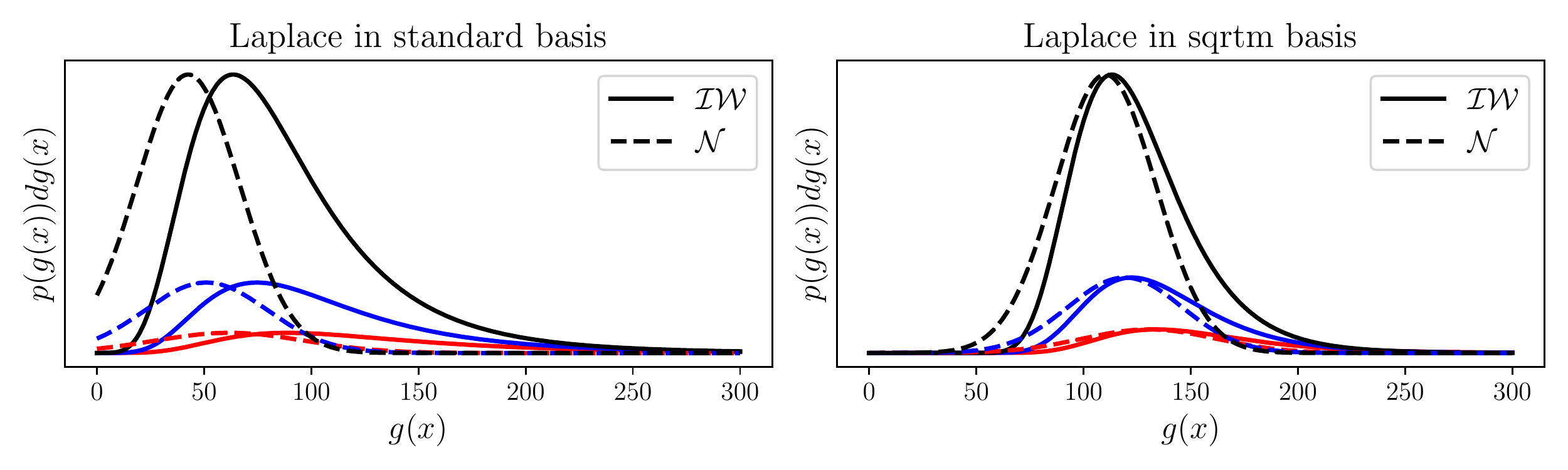}
	\caption{sqrtm-bridge for the inverse Wishart distribution.}
	\label{fig:inverse_wishart_sqrtm_bridge}
\end{figure}

\section{Experiments: Additional Explanations and Figures}
\label{sec:appendix_C}
\subsection{Distances}

\subsubsection{Parameters for KL and MMD}

As already discussed in the main text, the 10 (pairs of) parameters used to compute the KL-divergence and MMD are chosen such that they start from a small value which doesn't yield a valid Laplace approximation in the standard base and end with a larger value that is approximately Gaussian in the standard base since this is a feature many exponential families share. In Table \ref{table:disctance_parameters} you can find the exact parameters used for the figures in Table \ref{table:distances}.
\begin{table}[htb]
    \centering
    \caption{Parameters for the computation of KL-divergence and MMD.}
    \resizebox{\textwidth}{!}{
    \begin{tabular}{lllcccc}
        \toprule
        \textbf{Distribution}    & Parameter 1 & Parameter 2\\
        \midrule
        Exponential & $\lambda=[1,2,...,10]$ & - \\
        Gamma & $\alpha=[0.5, 1.5, ..., 9.5]$ & $\lambda=[0.5,1,...,5.5]$ \\
        Inv. Gamma & $\alpha=[1,...,10]$ & $\lambda=[0.5,1,...,5.5]$ \\
        Chi-squared & $k=[1,...10]$ & - \\
        Beta & $\alpha=[0.7, 1.2,...,5.2]$ & $\beta=[0.8, 1.05, ..., 3.05]$ \\
        Dirichlet & $\alpha=[0.8,0.8,0.8] * [1.5, 1, 0.75] * i$ & for $i$ in 1,...,10 \\
        Wishart & Discussed below & \\
        Inv. Wishart & Discussed below & \\
        \bottomrule
    \end{tabular}
    } 
    \label{table:disctance_parameters}
\end{table}

\subsubsection{Dirichlet KL computation} 

\begin{figure}[!tb]
    \centering
    \includegraphics[width=0.49\textwidth]{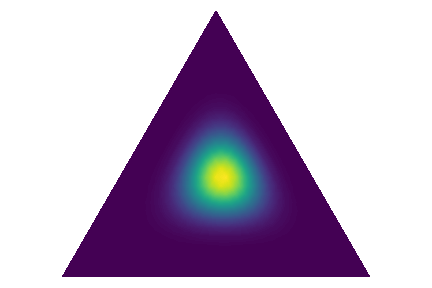} \hfill
    \includegraphics[width=0.49\textwidth]{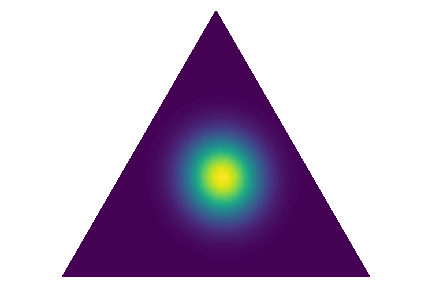} 

    \caption{Comparing original Dirichlet (left) with the fake softmax Gaussian descibed in the text (right).}
    \label{fig:Dirichlet_vs_fake_sm}
\end{figure}

Computing the KL-divergence for the Dirichlet is tricky because the Gaussian and the Dirichlet are defined on different domains. Computing the KL-divergence between the Gaussian and the Dirichlet in the inverse-softmax basis is also complicated because there is no inverse-softmax transformation that could transform samples as the softmax is not a bijective function. To solve this problem we transform the Gaussian into the probability domain. Since the simplex has one degree of freedom less than $\mathbb{R}^K$ we have to update $\mu$ and $\Sigma$ with a rank-1 constraint as already discussed in \cite{Hennig2010}.
\begin{align}
    \Bar{\mu} &= \mu - \frac{\Sigma \mathbf{1} \mathbf{1}^\top \mu}{\mathbf{1}^\top \Sigma \mathbf{1}} \\
    \Bar{\Sigma} &= \Sigma - \frac{\Sigma \mathbf{1} \mathbf{1}^\top \Sigma}{\mathbf{1}^\top \Sigma \mathbf{1}}
\end{align}
Then we project the variable $x$ into a $K-1$ dimensional subspace with $U=Ax$ where $A_{ij}$ for $i=1,...,K-1$ yielding
\begin{align}
    p(u) = \mathcal{N}(u | A\Bar{\mu}, A\Bar{\Sigma} A^\top) |\det \frac{\partial u}{\partial y}|
\end{align}
with $x(u) = [u, -\sum_i u_i]^\top$. Since we chose $y = \sigma(x) = \frac{\exp(x)}{\sum_j \exp(x_j)}$ we get $|\det \frac{\partial u}{\partial y}| = \frac{1}{y_i}\delta_{ij} - \frac{1}{K} \frac{1}{y_i}$ for $i=1,...,K-1$. By using
\begin{align}
    |Z + UWV| &= |Z||W||W^{-1} + V^\top Z^{-1} U|
\end{align}
we get
\begin{align}
    |\det \frac{\partial u}{\partial y}| &= (\prod_j^{K-1} \frac{1}{y_j}) \frac{1}{K} (K - \underbrace{\frac{1}{y}\text{diag}(y)}_{K-1}) \\
    &= \frac{1}{K} \prod_i^{K-1} \frac{1}{y_i}
\end{align}
We then draw samples $y$ from a Dirichlet distribution transform them with a fake inverse-softmax $x=\log(y) - \frac{1}{K}\sum_i\log(y_i)$ and apply the transformed Gaussian to get $\sum_j \log(\mathcal{D}(y_j)/\mathcal{N}(x_j)$ to estimate the KL-divergence. The implementation can be found in the accompanying code. 
We find that the Dirichlet and our constructed Gaussian look very similar (see Figure \ref{fig:Dirichlet_vs_fake_sm}). 

\subsubsection{(inverse-)Wishart KL-divergence}

The Wishart distribution has two parameters $n$ and $V$. The parameters of the inverse Wishart are $\nu$ and $\Psi$. The parameters for the distances always have the form of $n_0 + n_0 \cdot c \cdot i$ and $V_0 + V_0 \cdot d \cdot i$ for $i=0,1,...,9$ where $c$ and $d$ are constants. Then there are three interesting combinations of parameters to investigate: fixed $\nu$ with increasing $\Psi$, increasing $\nu$ with fixed $\Psi$, and both increasing. In the main paper we present only the first combination, here we present all of them. For the following we choose $n_0=2.5, c=0.5, V_0 = [[0.75, 0.5],[0.5,1]], d=0.25$.
Results for the inverse Wishart for all three scenarios can be found in Table \ref{table:distances_appendix}. Setup and trends are similar for the Wishart distribution.

\begin{table}[tb]
    \setlength{\figwidth}{0.5\textwidth}
    \setlength{\figheight}{0.12\textheight}
    \centering
    
    \sbox0{\hwplotStandard}\sbox1{\hwplotLog}\sbox2{\hwplotSqrt}%
    \caption{Comparison of Distances for Wishart distributions.
    }
    \resizebox{\textwidth}{!}{
    \begin{tabular}{l cc}
        \toprule
        \textbf{Version} &  KL-Divergence $\downarrow$ & MMD $\downarrow$ \\
        \midrule
        \multirow[c]{1}{*}[1.5em]{static $\nu$, inc. $\Psi$}  &
        \tiny 
\begin{tikzpicture}

\definecolor{color0}{rgb}{0.698,0.133,0.133}

\begin{axis}[
height=\figheight,
tick align=outside,
tick pos=left,
width=\figwidth,
x grid style={white!69!black},
xmin=-0.45, xmax=9.45,
xtick style={color=black},
xtick={-2,0,2,4,6,8,10},
xticklabels={
  \(\displaystyle -2\),
  \(\displaystyle 0\),
  \(\displaystyle 2\),
  \(\displaystyle 4\),
  \(\displaystyle 6\),
  \(\displaystyle 8\),
  \(\displaystyle 10\)
},
y grid style={white!69!black},
ymin=0, ymax=11.2,
ytick style={color=black},
ytick={0,10,20},
yticklabels={\(\displaystyle 0\),\(\displaystyle 10\),\(\displaystyle 20\)}
]
\addplot [semithick, black]
table {%
0 11.2
1 11
2 10.7
3 10.3
4 10.2
5 10.1
6 9.92
7 9.65
8 9.67
9 9.52
};
\addplot [semithick, color0]
table {%
0 4.42
1 4.24
2 4.02
3 3.9
4 3.78
5 3.65
6 3.53
7 3.45
8 3.35
9 3.28
};
\addplot [semithick, blue]
table {%
0 8.66
1 8.58
2 8.28
3 8.24
4 8.07
5 7.93
6 7.7
7 7.58
8 7.61
9 7.37
};
\end{axis}

\end{tikzpicture} & \tiny
\begin{tikzpicture}

\definecolor{color0}{rgb}{0.698,0.133,0.133}

\begin{axis}[
height=\figheight,
tick align=outside,
tick pos=left,
width=\figwidth,
x grid style={white!69!black},
xmin=-0.45, xmax=9.45,
xtick style={color=black},
xtick={-2,0,2,4,6,8,10},
xticklabels={
  \(\displaystyle -2\),
  \(\displaystyle 0\),
  \(\displaystyle 2\),
  \(\displaystyle 4\),
  \(\displaystyle 6\),
  \(\displaystyle 8\),
  \(\displaystyle 10\)
},
y grid style={white!69!black},
ymin=0, ymax=0.452,
ytick style={color=black},
ytick={0,0.25,0.5},
yticklabels={\(\displaystyle 0.00\),\(\displaystyle 0.25\),\(\displaystyle 0.50\)}
]
\addplot [semithick, black]
table {%
0 0.258
1 0.298
2 0.32
3 0.359
4 0.39
5 0.404
6 0.434
7 0.441
8 0.441
9 0.452
};
\addplot [semithick, color0]
table {%
0 0.102
1 0.102
2 0.0931
3 0.0926
4 0.0948
5 0.102
6 0.103
7 0.0998
8 0.0953
9 0.102
};
\addplot [semithick, blue]
table {%
0 0.155
1 0.174
2 0.176
3 0.189
4 0.205
5 0.223
6 0.221
7 0.241
8 0.239
9 0.248
};
\end{axis}

\end{tikzpicture} \\
        \multirow[c]{1}{*}[1.5em]{inc $\nu$, static $\Psi$}  &
        \tiny 
\begin{tikzpicture}

\definecolor{color0}{rgb}{0.698,0.133,0.133}

\begin{axis}[
height=\figheight,
tick align=outside,
tick pos=left,
width=\figwidth,
x grid style={white!69!black},
xmin=-0.45, xmax=9.45,
xtick style={color=black},
xtick={-2,0,2,4,6,8,10},
xticklabels={
  \(\displaystyle -2\),
  \(\displaystyle 0\),
  \(\displaystyle 2\),
  \(\displaystyle 4\),
  \(\displaystyle 6\),
  \(\displaystyle 8\),
  \(\displaystyle 10\)
},
y grid style={white!69!black},
ymin=0, ymax=11.3,
ytick style={color=black},
ytick={0,10,20},
yticklabels={\(\displaystyle 0\),\(\displaystyle 10\),\(\displaystyle 20\)}
]
\addplot [semithick, black]
table {%
0 nan
1 nan
2 nan
3 11.3
4 10.8
5 10.5
6 9.94
7 9.72
8 9.2
9 9.07
};
\addplot [semithick, color0]
table {%
0 5.75
1 5.05
2 4.63
3 4.47
4 4.43
5 4.53
6 4.82
7 5.11
8 5.49
9 5.96
};
\addplot [semithick, blue]
table {%
0 10.4
1 10
2 9.36
3 8.64
4 8.24
5 7.84
6 7.61
7 7.55
8 7.53
9 7.6
};
\end{axis}

\end{tikzpicture} & \tiny
\begin{tikzpicture}

\definecolor{color0}{rgb}{0.698,0.133,0.133}

\begin{axis}[
height=\figheight,
tick align=outside,
tick pos=left,
width=\figwidth,
x grid style={white!69!black},
xmin=-0.45, xmax=9.45,
xtick style={color=black},
xtick={-2,0,2,4,6,8,10},
xticklabels={
  \(\displaystyle -2\),
  \(\displaystyle 0\),
  \(\displaystyle 2\),
  \(\displaystyle 4\),
  \(\displaystyle 6\),
  \(\displaystyle 8\),
  \(\displaystyle 10\)
},
y grid style={white!69!black},
ymin=0, ymax=0.414,
ytick style={color=black},
ytick={0,0.25,0.5},
yticklabels={\(\displaystyle 0.00\),\(\displaystyle 0.25\),\(\displaystyle 0.50\)}
]
\addplot [semithick, black]
table {%
0 nan
1 nan
2 nan
3 0.234
4 0.176
5 0.136
6 0.109
7 0.077
8 0.0546
9 0.0418
};
\addplot [semithick, color0]
table {%
0 0.115
1 0.105
2 0.102
3 0.0909
4 0.0915
5 0.0878
6 0.0852
7 0.0815
8 0.0791
9 0.0686
};
\addplot [semithick, blue]
table {%
0 0.414
1 0.286
2 0.205
3 0.141
4 0.109
5 0.0841
6 0.0698
7 0.0519
8 0.0383
9 0.0303
};
\end{axis}

\end{tikzpicture} \\
        \multirow[c]{1}{*}[1.5em]{inc. $\nu$, inc. $\Psi$}  &
        \tiny 
\begin{tikzpicture}

\definecolor{color0}{rgb}{0.698,0.133,0.133}

\begin{axis}[
height=\figheight,
tick align=outside,
tick pos=left,
width=\figwidth,
x grid style={white!69!black},
xmin=-0.45, xmax=9.45,
xtick style={color=black},
xtick={-2,0,2,4,6,8,10},
xticklabels={
  \(\displaystyle -2\),
  \(\displaystyle 0\),
  \(\displaystyle 2\),
  \(\displaystyle 4\),
  \(\displaystyle 6\),
  \(\displaystyle 8\),
  \(\displaystyle 10\)
},
y grid style={white!69!black},
ymin=0, ymax=10.4,
ytick style={color=black},
ytick={0,10,20},
yticklabels={\(\displaystyle 0\),\(\displaystyle 10\),\(\displaystyle 20\)}
]
\addplot [semithick, black]
table {%
0 nan
1 nan
2 nan
3 10.4
4 9.92
5 9.68
6 9.24
7 9.13
8 8.78
9 8.81
};
\addplot [semithick, color0]
table {%
0 5.75
1 4.83
2 4.22
3 3.91
4 3.73
5 3.72
6 3.91
7 4.1
8 4.39
9 4.78
};
\addplot [semithick, blue]
table {%
0 10.4
1 9.75
2 8.88
3 8.06
4 7.58
5 7.13
6 6.87
7 6.78
8 6.73
9 6.8
};
\end{axis}

\end{tikzpicture} & \tiny
\begin{tikzpicture}

\definecolor{color0}{rgb}{0.698,0.133,0.133}

\begin{axis}[
height=\figheight,
tick align=outside,
tick pos=left,
width=\figwidth,
x grid style={white!69!black},
xmin=-0.45, xmax=9.45,
xtick style={color=black},
xtick={-2,0,2,4,6,8,10},
xticklabels={
  \(\displaystyle -2\),
  \(\displaystyle 0\),
  \(\displaystyle 2\),
  \(\displaystyle 4\),
  \(\displaystyle 6\),
  \(\displaystyle 8\),
  \(\displaystyle 10\)
},
y grid style={white!69!black},
ymin=0, ymax=0.414,
ytick style={color=black},
ytick={0,0.25,0.5},
yticklabels={\(\displaystyle 0.00\),\(\displaystyle 0.25\),\(\displaystyle 0.50\)}
]
\addplot [semithick, black]
table {%
0 nan
1 nan
2 nan
3 0.353
4 0.313
5 0.287
6 0.261
7 0.235
8 0.203
9 0.177
};
\addplot [semithick, color0]
table {%
0 0.115
1 0.105
2 0.102
3 0.0909
4 0.0915
5 0.0878
6 0.0852
7 0.0815
8 0.0791
9 0.0686
};
\addplot [semithick, blue]
table {%
0 0.414
1 0.307
2 0.241
3 0.185
4 0.157
5 0.138
6 0.125
7 0.104
8 0.0851
9 0.0742
};
\end{axis}

\end{tikzpicture} 
        \\
        \bottomrule
    \end{tabular}
    } 
    \label{table:distances_appendix}
\end{table}

\subsection{German Elections}

\begin{figure}[!tb]
    \centering
    \includegraphics[width=0.32\textwidth]{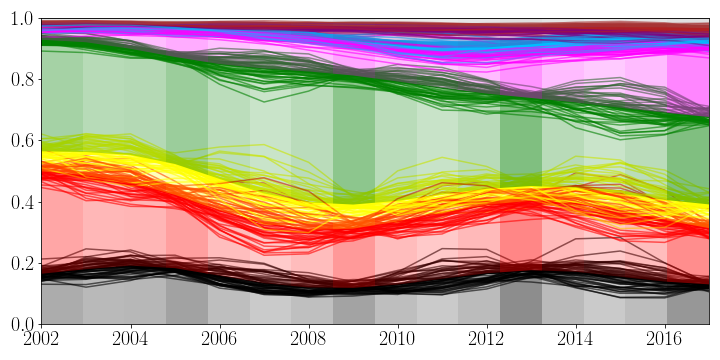} \hfill
    \includegraphics[width=0.32\textwidth]{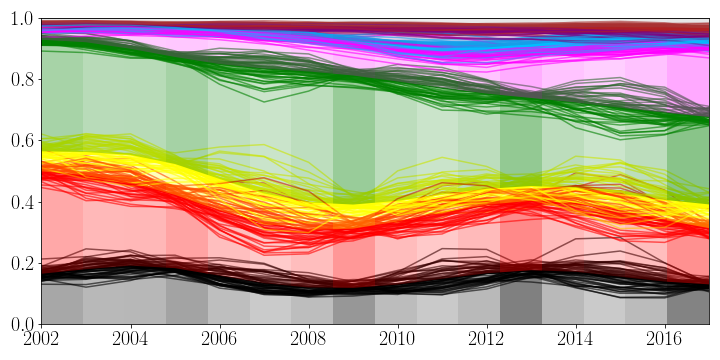} \hfill
    \includegraphics[width=0.32\textwidth]{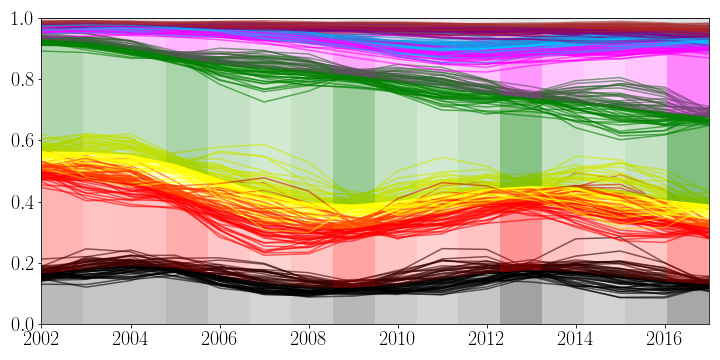}

    \caption{Tübingen elections from 2002 to 2017 for three different neighborhoods of Tübingen.}
    \label{fig:Tübingen_elections_GP}
\end{figure}

In the main text we compute the marginal distributions and the approximation qualities on a smaller dataset which uses much less datapoints, i.e. votes. In Figure \ref{fig:Tübingen_elections_GP} we present a similar visualization for the small local elections as we showed in the main text for the German elections.



\vskip 0.2in
\bibliography{LaplaceMatching}

\end{document}